\documentclass[12pt]{report}   
\usepackage{graphicx}  
\usepackage[letterpaper, left=1in, right=1in, top=1in, bottom=1in]{geometry}
\usepackage{setspace}  
\usepackage{times}  
\usepackage[explicit]{titlesec}  
\usepackage[titles]{tocloft}  
\usepackage[T1]{fontenc} 
\usepackage[utf8]{inputenc} 
\usepackage[backend=biber, natbib=false, sorting=none, bibstyle=ieee, citestyle=numeric-comp, sortcites=true]{biblatex}  
\usepackage{appendix}  
\usepackage{rotating}  
\usepackage[normalem]{ulem}  
\usepackage{textcomp} 
\usepackage{indentfirst} 
\usepackage{booktabs,array,arydshln} 
\usepackage{amsmath} 
\usepackage{newtxtext} 
\usepackage{hyperref}
\hypersetup{colorlinks,linkcolor={black},citecolor={orange},urlcolor={gray}}

\usepackage{xcolor}
\usepackage{algorithm} 
\usepackage{algpseudocode} 
\usepackage{subcaption}
\usepackage{mdframed}
\usepackage{multicol, caption}
\usepackage[font=small,labelfont=bf]{caption}
\usepackage{booktabs}

\usepackage{currfile}

\usepackage{amsmath,amssymb,amsfonts}%
\usepackage{amsthm}%
\usepackage{eqnarray}
\usepackage{bbm}
\usepackage{soul}
\usepackage{csquotes}
\usepackage{lipsum}
\usepackage{epigraph}

\newcommand{\ra}[1]{\renewcommand{\arraystretch}{#1}}
\newcommand{\bm}[1]{\mathbf{#1}}

\newcommand{\llabel}[1]{\label{\thechapter:#1}}
\newcommand{\lref}[1]{\ref{\thechapter:#1}}
\newcommand{\leqref}[1]{\eqref{\thechapter:#1}}

\AtEveryBibitem{\clearfield{issn}}
\AtEveryBibitem{\clearlist{issn}}

\AtEveryBibitem{\clearfield{language}}
\AtEveryBibitem{\clearlist{language}}

\AtEveryBibitem{\clearfield{doi}}
\AtEveryBibitem{\clearlist{doi}}

\AtEveryBibitem{\clearfield{url}}
\AtEveryBibitem{\clearlist{url}}

\AtEveryBibitem{%
  \ifentrytype{online}
    {}
    {\clearfield{urlyear}\clearfield{urlmonth}\clearfield{urlday}}}

\begin{document}
\doublespacing  


\begin{titlepage}
\begin{center}

\begin{singlespacing}
\vspace*{6\baselineskip}
Towards Human-level Dexterity via Robot Learning \\
\vspace{3\baselineskip}
Gagan Muralidhar Khandate \\
\vspace{18\baselineskip}
Submitted in partial fulfillment of the\\
requirements for the degree of\\
Doctor of Philosophy\\
under the Executive Committee\\
of the Graduate School of Arts and Sciences\\
\vspace{3\baselineskip}
COLUMBIA UNIVERSITY\\
\vspace{3\baselineskip}
\the\year
\vfill

\end{singlespacing}

\end{center}
\end{titlepage}


\begin{titlepage}
\begin{singlespacing}
\begin{center}

\vspace*{35\baselineskip}

\textcopyright  \,  \the\year\\
\vspace{\baselineskip}	
Gagan Muralidhar Khandate \\
\vspace{\baselineskip}	
All Rights Reserved
\end{center}
\vfill

\end{singlespacing}
\end{titlepage}

\pagenumbering{gobble}

\begin{titlepage}
\begin{center}

\vspace*{5\baselineskip}
\textbf{\large Abstract}

Towards Human-level Dexterity via Robot Learning

Gagan Muralidhar Khandate
\end{center}
\hspace{5mm}

Dexterous intelligence—the ability to perform complex interactions with multi-fingered hands—is a pinnacle of human physical intelligence and emergent higher-order cognitive skills.  However, contrary to Moravec's paradox, dexterous intelligence in humans appears simple only superficially. Many million years were spent co-evolving the human brain and hands including rich tactile sensing. Achieving human-level dexterity with robotic hands has long been a fundamental goal in robotics and represents a critical milestone toward general embodied intelligence. In this pursuit, computational sensorimotor learning has made significant progress, enabling feats such as arbitrary in-hand object reorientation. However, we observe that achieving higher levels of dexterity requires overcoming very fundamental limitations of computational sensorimotor learning.  

I develop robot learning methods for highly dexterous multi-fingered manipulation by directly addressing these limitations at their root cause. Chiefly, through key studies, this disseration progressively builds an effective framework for reinforcement learning of dexterous multi-fingered manipulation skills. These methods adopt structured exploration, effectively overcoming the limitations of random exploration in reinforcement learning. The insights gained culminate in a highly effective reinforcement learning that incorporates sampling-based planning for direct exploration. Additionally, this thesis explores a new paradigm of using visuo-tactile human demonstrations for dexterity, introducing corresponding imitation learning techniques.

\vspace*{\fill}
\end{titlepage}

\pagenumbering{roman}
\setcounter{page}{1} 
\renewcommand{\cftchapdotsep}{\cftdotsep}  
\renewcommand{\cftchapfont}{\normalfont}  
\renewcommand{\cftchappagefont}{}  
\renewcommand{\cftchappresnum}{Chapter }
\renewcommand{\cftchapaftersnum}{:}
\renewcommand{\cftchapnumwidth}{5em}
\renewcommand{\cftchapafterpnum}{\vskip\baselineskip} 
\renewcommand{\cftsecafterpnum}{\vskip\baselineskip}  
\renewcommand{\cftsubsecafterpnum}{\vskip\baselineskip} 
\renewcommand{\cftsubsubsecafterpnum}{\vskip\baselineskip} 

\titleformat{\chapter}[display]
{\normalfont\bfseries\filcenter}{\chaptertitlename\ \thechapter}{0pt}{\large{#1}}

\renewcommand\contentsname{Table of Contents}

\begin{singlespace}
\tableofcontents
\setlength{\cftparskip}{\baselineskip}
\listoffigures
\listoftables
\end{singlespace}

\clearpage

\phantomsection
\addcontentsline{toc}{chapter}{Acknowledgments}

\clearpage
\begin{center}

\vspace*{5\baselineskip}
\textbf{\large Acknowledgements}
\end{center}

This thesis would not have been possible without the support, collaboration, and guidance of numerous individuals and communities. I would like to express my sincere gratitude to everyone who contributed to this journey.

First and foremost, I extend my deepest thanks to my advisor, Matei Ciocarlie. Matei’s guidance has been invaluable, not only in technical matters but also in shaping my understanding of the research process, instilling in me a love for discovery, and demonstrating the importance of kindness and integrity in academia. Once an advisor, always an advisor—Matei’s mentorship will continue to guide me long after this work is done.

I am fortunate to have collaborated with an incredible group of individuals throughout this work. I thank Eric Chang, Xingsheng Wei, Kevin Mahoney, Cameron Mehlman, Kai Jiang, Sarah Park, Boxuan (Dave) Wang, Weizhe Ni, Joaquin Palacious, Emily Hannigan, Bing Song, and Pratyus Pati for their insights, hard work, and dedication. Special thanks to Siqi Shang for the countless hours spent refining policies for sim-to-real transfer, Tristan Saidi for extending the results to additional tasks such as in-hand reorientation, Seth Dennis for designing and building a robust robotic hand, Ava Chen for her many hardware insights across various projects, and Zhanpeng He for all the long research and life discussions that provided invaluable perspective. I am also deeply grateful to Pedro Piacenza \& Eric Chang for developing the incredible tactile sensors without which this work would not have been possible, and to Maximilian Haas-Heger, from whom I learned invaluable software engineering discipline that has greatly shaped my approach to development.

I am also grateful to my lab members for their camaraderie and support. Ava Chen, Zhanpeng He, Jingxi Ju, and Sharfin Islam, Eric Chang, Katelyn Lee have provided a collaborative and inspiring environment that has greatly enriched my research experience.

I would also like to express my gratitude to the remarkable individuals I met during my internships. I thank Yifan Hou, Neel Doshi, Michael Mistry, Ankit Bhatia, Yi Tian, Sasha Salter, and  for the many engaging discussions and valuable insights they shared, which broadened my perspective and deepened my understanding of the field. I am especially grateful to Aaron Parness, Rick Warren for the opportunity to work in such inspiring environments, which played an important role in shaping my research journey.

I owe a special thanks to my mentors, Neel Doshi, David Watkins who have helped me navigate various stages of my academic journey.

To my dissertation committee members, Shuran Song, Carl Vondrick, Jitendra Malik, and Gaurav Sukhatme, I am immensely grateful for their guidance, constructive feedback, and encouragement in bringing this work to completion.

I would also like to acknowledge the professors who have provided valuable instruction and insights in various courses: Daniel Russo, Chong Li, Shuran Song, Carl Vondrick and of-course Matei. Their classes have laid the foundation for much of what I know today.

In the Computer Science Department, I thank Carol, DaShante, and the entire team for their administrative support and for making my time in the program a smooth and enjoyable experience. I would also like to acknowledge Amoy and all the staff from Mechanical Engineering for the assistance provided throughout with research.

To my friends, Gaurav Jain, Purva Tendulkar, Harsh Aggrawal, Basile Von-hoorick, Christos Tsanikidis, Jeremy Johnston, Mehrnoosh, Mahshid Ghasemi, Madhumitha Shridharan, Arjun Mani, Nancy Ouyang, thank you for being there through the highs and lows. Your friendship has been a source of strength and joy. Special thanks to Gaurav, Purva, Harsh \& Basile who were source of constant support.

None of this would have been possible without the unwavering support of my family. I am deeply grateful to my mom and dad, Vasanti and Muralidhar, for giving me the freedom to choose my own path and for the countless sacrifices they made along the way. I also thank my sister, Tanu, for being a constant source of love and mischief. A special thanks to Pramod Gayakwad, who guided me through my early education and set me on this journey.

Finally, I express my heartfelt appreciation to everyone who has supported me on this journey. This thesis is as much a product of your guidance and encouragement as it is of my own efforts. Thank you all for being part of this journey.

\clearpage


\phantomsection
\addcontentsline{toc}{chapter}{Dedication}

\begin{center}

\vspace*{5\baselineskip}
\textbf{\large Dedication}
\end{center}

To Mom and Dad, I will always remain indebted for your love and unwavering support.




\clearpage
\pagenumbering{arabic}
\setcounter{page}{1} 


\titleformat{\chapter}[display]
{\normalfont\bfseries\filcenter}{}{0pt}{\large\chaptertitlename\ \large\thechapter : \large\bfseries\filcenter{#1}}  
\titlespacing*{\chapter}
  {0pt}{0pt}{30pt}	
  
\titleformat{\section}{\normalfont\bfseries}{\thesection}{1em}{#1}

\titleformat{\subsection}{\normalfont}{\thesubsection}{0em}{\hspace{1em}#1}

\titleformat{\subsubsection}{\normalfont\itshape}{\thesubsection}{1em}{#1}

\chapter{Introduction}

\begin{displayquote}
“The hand is the cutting edge of the mind. Civilisation is not a collection of finished artefacts, it is the elaboration of processes. In the end, the march of man is the refinement of the hand in action.”
― \textit{Jacob Bronowski, The Ascent Of Man}
\end{displayquote}

Dexterity is the pinnacle of human physical intelligence and is essential for daily life \cite{Ritter2015-fc}. It enables precise and purposeful interactions with the physical world, a capability crucial for robotic systems. By achieving dexterous manipulation, robots can perform complex tasks such as tool use and adapt to unstructured environments, allowing them to handle a wide range of objects.

Furthermore, achieving dexterous manipulation not only enhances the practical abilities of robots but also lays the groundwork for advancing high-level reasoning in physically interactive systems, whose full potential is realized through multi-fingered robotic hands. Understanding how to reach human-level dexterity in robotic hands provides valuable insights into the nature of intelligence itself. While robotics has long been a driving force for innovation in intelligent systems, dexterity remains one of its most challenging problems, making it a fertile area for further advancements. Solving this problem will require reevaluating current assumptions and exploring new approaches.

Now that we have established the importance of achieving human-level dexterity, it is essential to recognize just how challenging this goal is. Human dexterity, particularly the ability to integrate tactile sensing to achieve high levels of precision, may seem straightforward at first glance, but it is, in fact, highly complex. Humans routinely use in-hand manipulation for tasks like reorienting a tool from an initial grasp into a useful position, securing a better hold on an object, or exploring the shape of an unfamiliar item. These actions require intricate coordination and precise control, skills that evolution has refined over millennia.

Our hands are the product of a long evolutionary process, and learning to use them with precision takes significant time \cite{Baker2024-kv}. Dexterous manipulation is one of the last motor skills humans develop \cite{Humphry1995-xj, Needham2023-rz, Steinhart2021-br, Viholanen2006-hb}, underscoring its complexity. The central nervous system dedicates substantial resources to both sensing and controlling the hand, reflecting the importance of this capability \cite{van-Beers2002-jw, Cabrera2023-kg}. The sophistication of the human hand is a testament to the intricate balance between our physiology and the neural mechanisms that enable advanced manipulation skills.

Dexterity presents a significant challenge for robotic hands due to the complexity of contact-rich interactions and various physical attributes. Factors such as the texture of an object's surface and the dynamics of the fingers can greatly impact performance. Finally, controlling a high-degree-of-freedom (DOF) hand is inherently challenging resulting in a high degree of complexity.

Another major challenge in achieving human-like dexterity is the difficulty of tactile sensing \cite{Lambeta2020-ir, Piacenza2020-dv, Yuan2017-rq}. Integrating effective tactile sensing into robotic hands is a complex task due to several hardware-related issues. Tactile sensing in robotics involves capturing and interpreting fine-grained feedback from interactions, which requires sophisticated sensors and systems.

The human hand exemplifies the pinnacle of tactile sensing \cite{Kim2020-rx}. Our fingertips are equipped with highly sensitive receptors that provide detailed information about texture, pressure, and temperature. This capability is crucial for a wide range of dexterous skills, such as precise manipulation, delicate gripping, and intricate tasks \cite{Hogan2020-zs}. The ability to detect subtle changes and apply appropriate pressure is essential for effective interaction with various objects and environments. This fine-tuned tactile feedback allows humans to adapt to varying conditions and perform complex tasks with precision .

Achieving similar tactile sensing capabilities in robotic hands is critical for replicating human dexterity. This involves designing advanced sensors and integrating them with the robotic system to capture and process tactile information accurately.

To achieve dexterity in robotic hands, significant work is required in several areas, including hand design, tactile sensing, and learning algorithms. Currently, data-driven sensorimotor learning techniques have shown the most promising results in developing dexterous robotic hands. These methods leverage large scale simulation to train policies that can effectively control robotic hands and adapt to a wide range of tasks, bringing us closer to achieving human levels of dexterity in robotic hands.

\section{Computational sensorimotor learning for dexterity}

Inspired by biological learning systems, computational sensorimotor learning uses experience i.e. data driven learning. Computational Sensorimotor Learning is going sensing to actions via a neural network end-to-end in fashion.  This has also been used to learn dexterous skills with varying approaches \cite{Weinberg2024-iz, Pehoski1997-ko, }. A broad spectrum of computational sensorimotor learning methods, from reinforcement learning starting in a simulated environment to imitation learning with real-world demonstrations, have been attempted in literature.

Reinforcement learning has been responsible for most of the impressive demonstrations of dexterous manipulation skills  as seen in recent works \cite{OpenAI2019-ng, Allshire2021-qp, Chen2022-rc, Qi2022-wy, Khandate2022-qt, Khandate2023-gy, Qi2023-iz, Yin2023-qg,  Pitz2023-xi, Pitz2023-kj, Sievers2022-lb, Yang2024-yc} which demonstrate in-hand reorientation to a specific pose or about continuously about an axis of rotation.

Although various imitation learning frameworks have been proposed for multi-fingered dexterous skills,  the dexterity is limited in comparison to the impressive results achieved with reinforcement learning. 

One challenge is the limited use of tactile sensing in reinforcement learning methods \cite{Yousef2011-jb, Chen2018-bs, Kim2024-qq}. Simulating rich tactile sensing is inherently difficult, making it challenging to incorporate into RL frameworks. Despite its critical role in achieving human-level dexterity, tactile sensing is underutilized in current approaches to dexterous manipulation, with only a few works leveraging it effectively \cite{Van_Hoof2017-xy, Yousef2011-jb, Qi2022-wy, Xu2022-jm, Qi2023-iz, Suresh2023-nv, Lepora2024-gi, Lu2024-mh}.

While current methods have enabled the development of impressive dexterous skills, progress toward human-level dexterity remains hindered \cite{Wang2024-po}. The core of this issue relates to data, and to fully understand this, we must contrast RL with imitation learning (IL) methods. We will explore this comparison next, followed by a discussion of the key obstacle preventing the achievement of human-level dexterity.

\section{Reinforcement vs imitation Learning}

Taking a step back from dexterous manipulation, let us contrast the state-of-the-art reinforcement learning with imitation learning. Both RL and IL, have enabled challenging sensorimotor skills. The effectiveness of the approach depends on the specific skill, with opposite classes of learning methods proving effective in different scenarios. 

When robot demonstrations are readily accessible, such as in two-fingered bi-manual manipulation, imitation learning methods with expressive multi-modal policies take the spotlight \cite{Chi2023-mz, Zhao2023-qq, Sharma2023-zh, Zhao2023-jk, Chi2023-vp, Bharadhwaj2023-hn}. In contrast, in situations where obtaining demonstrations is challenging, reinforcement learning emerges as an effective approach,  as seen in complex tasks such as multi-fingered grasping\cite{Zhou2022-ee}, in-hand manipulation  \cite{OpenAI2019-ng, Allshire2021-qp, Chen2022-rc, Qi2022-wy, Khandate2022-qt, Khandate2023-gy, Qi2023-iz, Yin2023-qg,  Pitz2023-xi, Pitz2023-kj, Sievers2022-lb} or locomotion \cite{Cheng2023-nv, Lee2020-lh, Hwangbo2019-lo, Zhuang2023-bi}. 

But both RL and IL have their own limitations. On the one hand, reinforcement learning methods face challenges in sim-to-real transfer and extensive generalization, as well as high simulation costs in pursuing these objectives. Due to the difficulty of exploration, long horizon dexterous skills or skills with unstable dynamics are also hard to achieve with reinforcement learning. On the other hand, imitation learning policies struggle with generalization beyond demonstrations. While pre-training on extensive robot demonstration datasets can yield policies capable of diverse manipulation skills \cite{Jang2022-bp, Brohan2022-jp, Bousmalis2023-wy, Zitkovich2023-rz}, acquiring the necessary demonstrations for this pre-training proves expensive and resource-intensive. 

These unique limitations of RL and IL are significantly exacerbated when applied to multi-fingered dexterity, primarily due to a data scarcity issue.



\begin{figure}
    \centering
    \includegraphics[width=0.9\textwidth]{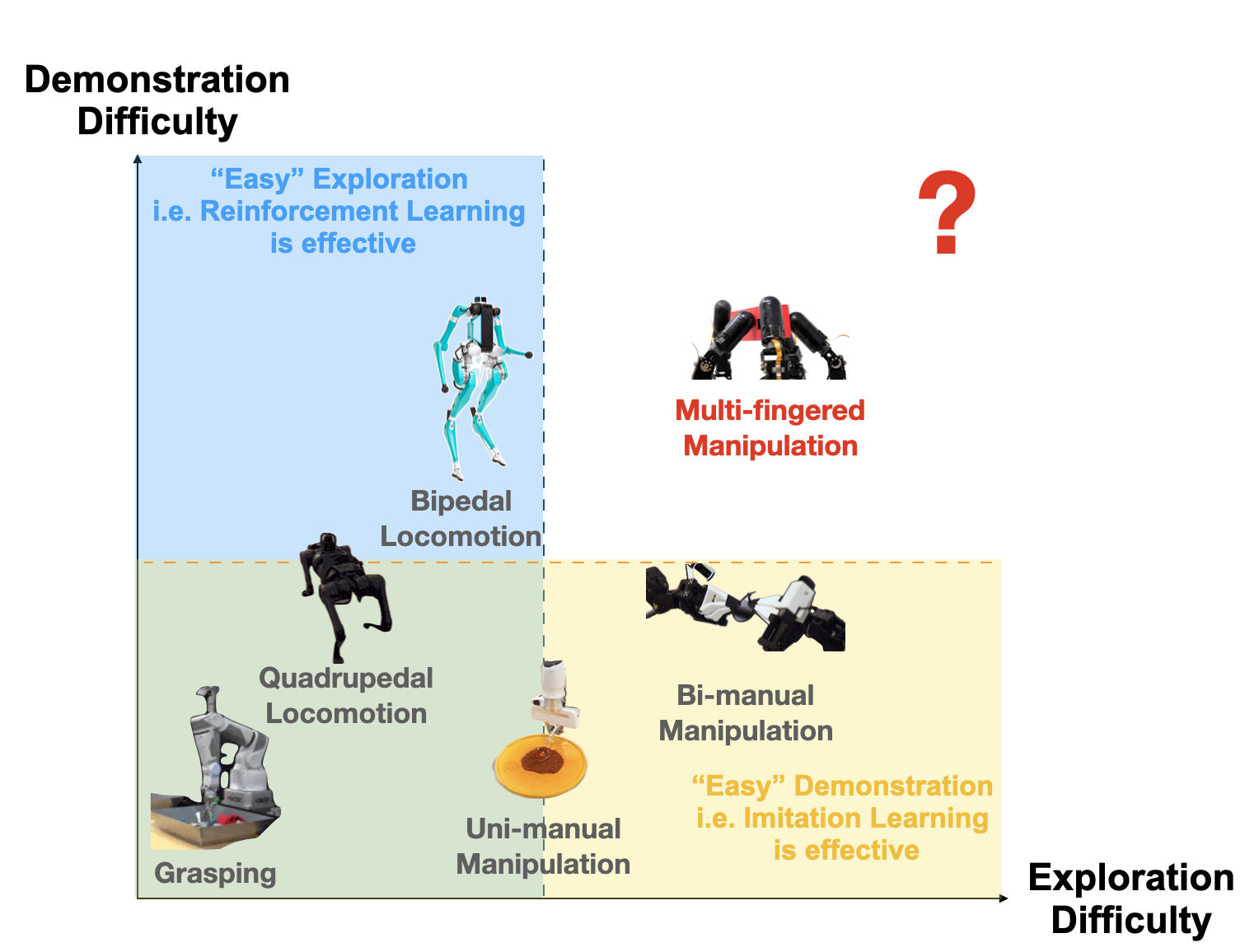}
    \caption{Several representative robot learning skills are presented, categorized by the difficulty of collecting demonstrations and performing exploration. Learning multi-fingered dexterous manipulation is particularly challenging due to the extreme difficulty in both collecting demonstrations and achieving effective exploration, making it a complex problem for both imitation and reinforcement learning.}
    \llabel{fig:rl_vs_il}
\end{figure}

\section{Data scarcity limits dexterity}

We will now substantiate the claim that multi-fingered dexterity faces a significant data problem.

A multi-fingered hand manipulating an object in-hand, particularly when using only fingertips, represents an inherently unstable system where even small perturbations can easily cause the object to be dropped.

Attempting to address this with RL presents significant challenges. Random Gaussian actions used for exploration are insufficient, as the instability discussed above makes exploration particularly difficult. As a result, RL methods struggle to scale effectively, often requiring billions of steps \cite{OpenAI2019-ng} yet still falling short due to the inherent complexity of the task. Thus, RL fails primarily due to a scarcity of exploratory data, exacerbated by the difficult dynamics.

On the other end of the spectrum, we might consider collecting demonstrations through teleoperation. However, the complex dynamics of multi-fingered manipulation make it difficult to provide demonstrations for anything beyond simple tasks, which mostly involve basic open-and-close motions of the hand. While tactile sensing is crucial for tasks requiring high dexterity, collecting demonstrations through teleoperation is not a viable approach. The intricate dynamics, which demand precise tactile feedback, need to be accurately rendered back to the human operator—a challenge that current systems cannot meet. Consequently, IL methods falter here as well, due to the scarcity of demonstrations or data.

With these insights, we can explore solutions to address the data problem in dexterous manipulation. This dissertation presents methods for overcoming data scarcity in both reinforcement learning and imitation learning, aiming to achieve higher levels of dexterity than are currently possible. 

Let us first consider reinforcement learning. As discussed in the previous section, reinforcement learning for dexterous manipulation is severely constrained by data scarcity due to the inherent challenges of exploration. To address this fundamental limitation, we investigate methods specifically designed to overcome these challenges.

\section{Dexterity through structured exploration}

If we can overcome the exploration challenges inherent in reinforcement learning for complex dexterous manipulation, it may become feasible to transfer these skills to a real robotic hand using techniques such as domain randomization \cite{Tobin2017-zh, Tiboni2022-iy}, rapid motor adaptation \cite{Kumar2021-zv}, and fine-tuning with real-world rollouts \cite{Torne2024-id}. Motivated by this potential, we focus on addressing the exploration challenge directly.

Exploration difficulty arises when an agent cannot efficiently reach or explore critical regions of the state-space. Fortunately, in many instances of multi-fingered manipulation, these target regions are often identifiable; for instance, desired grasp configurations can be computed in advance. In our initial study, we leveraged these state-space priors to guide exploration, facilitating reinforcement learning for challenging dexterous skills.

In a second study, we investigated the use of action priors to further streamline exploration. By implementing controllers for specific sub-skills, we constrained exploration to a lower-dimensional action sub-space. This approach alleviates the difficulties associated with high-dimensional action spaces in dexterous manipulation, ultimately enhancing the agent’s capacity to explore the state-space effectively.

Finally, building on these insights, we developed a comprehensive framework for structured exploration by integrating sampling-based planning with reinforcement learning. Sampling-based planning methods supply essential state and action priors, bootstrapping exploration in critical regions of the workspace for dexterous manipulation. In essence, sampling-based planning addresses the data scarcity inherent in reinforcement learning by guiding the agent efficiently toward high-value areas of the state-space. Our results demonstrate that structured exploration through sampling-based planning substantially improves reinforcement learning, enabling the acquisition of highly dexterous skills.

While reinforcement learning with structured exploration can achieve higher levels of dexterity, scaling it to diverse skills and advancing to even greater levels of dexterity present challenges, especially as the sim-to-real gap widens. Setting up simulations for each new task is cumbersome, and the engineering difficulty for effective sim-to-real transfer can continue to increase as we pursue higher levels of dexterity.

We can turn to imitation learning, though, as previously discussed, it also faces challenges with data scarcity. However, imitation learning may still offer a viable path forward, which we will now explore conceptually.

\section{Dexterity by imitating human dexterity}

Imitation learning with human demonstrations addresses these challenges effectively. First, it eliminates the need for exploration, as the demonstration data directly guides the learning process. Second, the data is collected on the real robot, avoiding any sim-to-real gap and ensuring the learned behaviors are immediately applicable.

For imitation learning to succeed, however, solving the data problem means finding a way to collect useful demonstrations. This presents a significant challenge, as achieving on-robot demonstrations creates a chicken-and-egg dilemma: it's impossible to collect meaningful demonstrations without already having a policy in place to guide those demonstrations.

To overcome this, we turn to human embodiment as a solution. By equipping the human hand with sensing capabilities similar to those of the robotic hand, we can observe and record humans performing various challenging manipulation tasks. We can replace the sim-to-real difficulty with a potentially more manageable human-to-robot gap. This approach may allow us to gather valuable demonstration for training dexterous skills, bridging the gap between human dexterity and robotic capability.

\section{Contributions}
This work focuses on overcome key challenges in achieving human-level dexterity in robotic systems, particularly focusing on in-hand object manipulation with tactile sensing. By addressing critical data limitations in multi-fingered dexterity, this research establishes new approaches for structured exploration in reinforcement learning and introduces innovative techniques for obtaining valuable demonstrations for imitation learning. The primary contributions are as follows:

\begin{itemize}
    \item \textbf{A first-of-its-kind framework for structured exploration in reinforcement learning that integrates sampling-based planning with reinforcement learning algorithms:} This novel approach employs a modified Rapidly-exploring Random Tree planning to efficiently traverse the state space of dexterous manipulation tasks. The extracted paths planning from yield informative reset distributions that guide the reinforcement learning agent toward enhanced performance. This framework facilitates the learning of complex finger-gaiting skills and achieves significant gains in sample efficiency over traditional reinforcement learning methods.

    \item \textbf{Demonstration of dexterous in-hand manipulation skills acquired solely from intrinsic tactile and proprioceptive sensing, without reliance on external sensors or support surfaces:}  This enables dexterous manipulation in environments where external sensing is limited or unreliable, such as in the dark. Using intrinsic sensing data alone, this work demonstrates successful learning of finger-gaiting and finger-pivoting policies, advancing the field of autonomous, real-world robotic manipulation.

    \item \textbf{The first state-of-the-art techniques for precision in-hand manipulation of complex, varied object shapes using intrinsic sensing data:} Enabled by structured exploration, this work breaks new ground by addressing the unique challenges posed by achieving in-hand finger-gaiting of challenging object geometries such as large and convex shapes. Leveraging intrinsic sensing data, these techniques enable adaptive manipulation of a range of shapes and novel objects.


    \item \textbf{A novel paradigm for obtaining human demonstrations in dexterous manipulation by leveraging visuo-tactile data:} To address the practical limitations of collecting robot demonstrations for multi-fingered manipulation, this approach introduces a first-of-its-kind method that leverages visuo-tactile data captured from humans. Through the Visuo-Tactile Transformer (ViTacT) architecture, multi-modal sensory observations are encoded from human demonstrations, towards enabling cross-embodiment transfer of dexterous skills to robotic hands and setting a new standard for imitation learning in dexterous manipulation.

\end{itemize}

\section{Summary}

The following chapters detail the methods developed based on the key ideas discussed earlier. Chapter 1 and Chapter 2 present two studies that explore the use of priors derived from human or domain knowledge, which motivated subsequent work on reinforcement learning with structured exploration. These chapters also demonstrate finger-gaiting dexterity using only tactile sensing within a simulated environment.

Chapter 3 introduces a framework for structured exploration, which achieves highly dexterous finger-gaiting skills. Chapter 4 extends this framework to include both structured exploration and exploitation, leading to a more comprehensive approach. Preliminary results from this extended framework are discussed.

Finally, Chapter 5 presents a new paradigm for human demonstrations, as briefly introduced above. This chapter explores cross-embodiment imitation learning for transferring dexterous skills from human to robotic hands, including preliminary results of this approach.

\chapter{Related Work}

Dexterous manipulation with multi-fingered hands has long been a focus of robotics research. Consequently, classical analytical and model-based techniques were predominantly employed for dexterous manipulation before adopting data-driven methods. We discuss these model-based methods (Section \lref{sec:model-based}) followed by a discussion of the full spectrum of sensorimotor learning techniques for dexterous manipulation, encompassing approaches that learn dexterous manipulation from scratch (Section ), methods based on learning from demonstration and hybrid techniques (Section ). We also briefly discuss works that leverage smart or task-optimized hardware to achieve dexterous skills. Next, we discuss prior work relevant to structured exploration, one of the two pursuant ideas in this dissertation. The difficulty of exploration in reinforcement learning is a well-known problem. Thus, we discuss prior techniques for improving exploration with reinforcement learning, followed by relevant work on structured exploration (Section ). Finally, after briefly reviewing key ideas and methods for learning from human videos, we discuss prior work on learning dexterous manipulation from human demonstrations (Section ), a line of work also pursued in this dissertation.

\section{Model-based methods for dexterous in-hand manipulation}
\llabel{sec:model-based}

Multi-fingered dexterity or dexterous manipulation \cite{Okamura2000-cn} involves various within-hand manipulation or in-hand skills \textcite{Ma2011-fo} such as regrouping, in-grasp manipulation, finger-gaiting, finger-pivoting, rolling, and sliding. A key challenge in multi-fingered is the transient nature of contacts, i.e., the fingers of the hand have to make and break contact; this discontinuity in physics makes it challenging to design a single analytical controller to perform dexterous in-hand manipulation. 

Hence, early methods decompose dexterous in-hand manipulation into grasp transitions and present analytical controllers. The grasp transitions are either fixed or discovered via search and are also sometimes modeled as an MDP between different grasp types.  While some works focus on finger-gaiting \cite{Leveroni1996-iy, Han1998-xj, Platt2004-nr, Saut2007-su}, others focused on finger-pivoting \cite{Omata1996-hp}. \textcite{Leveroni1996-iy} search for grasp gaits with maps of grasps involving fingertip combinations. \textcite{Han1998-xj} integrated the relevant theories of contact kinematics, non-holonomic motion planning, coordinated object manipulation, grasp stability, and finger gaits towards a framework for dexterous manipulation. \textcite{Platt2004-nr} further demonstrated controllers based on wrench-closure between different grasps and used MDP \cite{Puterman1990-on} to define transitions between these controllers. These efforts generally used simplifying assumptions,  such as 2D manipulation, accurate models, and smooth object geometries, which make them amenable for analysis but limit their versatility.

The following wave of methods used online model-based trajectory optimization. Most of these efforts used online trajectory optimization with model-predictive control. Online non-linear trajectory optimization techniques with differential dynamic programming coupled with soft-contact models \cite{Todorov2012-xe} enabled complex behavior synthesis \cite{Tassa2012-hx} including dexterous robotic manipulation \textcite{Mordatch2012-wh}. However, these methods did not scale for finger-gaiting, where the transient nature of contacts fatally exacerbates their limitations: transient contacts introduce large non-linearities in the model, which also depend on hard-to-model contact properties. More recently, methods proposed online kinematic trajectory optimization \cite{Fan2017-vz, Sundaralingam2018-zw, Charlesworth2020-oa, Jin2024-we}, circumventing the above difficulties associated with full kinodynamic trajectory optimization and demonstrate finger-gaiting in simulation.  Alternatively, a few works also considered sampling-based planning (RRT) for finger-gaiting in-hand manipulation \cite{Yashima2003-lw, Xu2007-yb} limited to simulation for a spherical object. Overall, the key limitation in this method class is the models and challenging real-time optimization. Both affect the ability to achieve dexterous skills as the models are unavailable or often inaccurate, making them challenging to transfer to real hands.

\section{Reinforcement learning for dexterous in-hand manipulation}
\llabel{sec:RL}
The rise of data-driven sensorimotor learning has led to a wave of complex robotics tasks, including dexterous manipulation. In particular,  direct policy with model-free reinforcement learning has allowed impressive results over the last few years. In seminal works, \textcite{OpenAI2018-bx, OpenAI2019-ng} demonstrated finger-gaiting and finger-pivoting skills with a cube. However, due to the poor sample complexity, learning a policy required hundreds of hours in training covering billions of simulation steps. Hence, towards achieving similar skills efficiently, some prior work \cite{Rajeswaran2017-au, Zhu2019-rk, Radosavovic2021-pg} augment reinforcement learning with human expert trajectories to improve sample complexity for dexterous manipulation.  However, these expert demonstrations are hard to obtain for precision in-hand manipulation tasks and even more so for non-anthropomorphic hands. Towards learning skills across multiple objects \textcite{Huang2021-vo} demonstrated geometry-aware multi-task reinforcement learning with point-cloud object representation. Recently \textcite{Makoviychuk2021-ko} showed that GPU physics could accelerate learning skills similar to OpenAI's. \textcite{Allshire2021-qp} used extensive domain randomization and sim-to-real transfer to re-orient a cube but used a tabletop as an external support surface. With GPU physics, \textcite{Handa2022-fb} demonstrated in-hand re-orientation for a wide range of objects under palm-up and palm-down orientations of the hand with extrinsic sensing providing dense object feedback. 

In most of the cases, these policies do not transfer to arbitrary orientations of the hand as they expect the palm to support the object during manipulation \textemdash a consequence of the policies being trained in a palm-up hand orientation, which simplifies training. In other cases, the policies require extensive external sensing involving multi-camera systems to track the fingers and/or the object, systems that are hard to deploy outside the lab. However, we are interested in achieving these skills exclusively through using fingertip grasps (i.e precision in-hand manipulation \cite{Michelman1998-ye}) without requiring the presence of the palm underneath the object, which enables the policies to be used in arbitrary orientations of the hand. Furthermore, we would like to circumvent the need for cumbersome external sensing by only using internal sensing with proprioceptive and tactile feedback in achieving these skills.

Another limitation of the above works, is that tactile sensing is rarely considered as tactile feedback as tactile sensing available on real hand is often high dimensional and hard to simulate \cite{OpenAI2018-bx}. Hence, van \textcite{Van_Hoof2015-id} proposes learning directly on a real hand equipped with tactile sensing, but this limits them to tasks learnable on real hands. While \textcite{Melnik_undated-wr, Melnik2021-zq} show that tactile feedback improves sample complexity in such tasks, they use high-dimensional tactile feedback with full coverage that is hard to obtain on a real hand. We consider low-dimensional tactile feedback covering only the fingertips.

Alternatively, model-based reinforcement learning can also be considered for some in-hand manipulation tasks: \textcite{Nagabandi2020-hz} manipulate boarding balls but use the palm for support; \textcite{Morgan2021-ny} learn finger-gaiting but with a task-specific under-actuated hand. However, learning a reliable forward model for precision in-hand manipulation with a fully dexterous hand can be challenging. Collecting data involves random exploration, which, as discussed later, has difficulty exploring in this domain.

Various other works leverage hardware towards achieving in-hand object reorientation. \textcite{Morgan2021-ny} achieved finger-gaiting with an under-actuated hand specifically designed for this task. We consider finger-gaiting with a highly actuated hand a harder problem due to poor sample complexity stemming from additional degrees of freedom. \textcite{Bhatt2022-fr} also demonstrate robust finger-gaiting and finger-pivoting manipulation with a soft, compliant hand. Still, these skills were hand-designed and executed in an open-loop fashion rather than autonomously learned.

Only the most recent works consider precision in-hand manipulation with a fully dexterous hand.  Contemporary to our work \cite{Khandate2021-wl}, \textcite{Chen2021-ig}  show in-hand re-orientation without support surfaces that generalizes to novel objects. The policies exhibit complex dynamic behaviors including occasionally throwing the object and re-grasping it in the desired orientation. We differ from this work as our policies only use sensing that is internal to the hand, and always keep the object in a stable grasp so as to be robust to perturbation forces at all times. Furthermore, our policies require a number of training samples that is smaller by multiple orders of magnitude, a feature that we attribute to efficient exploration via appropriate reset state distributions. \textcite{Sievers2022-lb, Pitz2023-kj, Rostel2023-jo} also demonstrated in-hand reorientation of a cube to the desired pose with polices using object state-estimation and tactile feedback.

Next, with improved exploration achieved using reset distributions, \textcite{Qi2022-wy, Yin2023-qw} further used rapid motor adaptation to achieve effective sim-to-real transfer of tactile-only in-hand manipulation skills for continuous reorientation of small cylindrical and cube-like objects. \textcite{Yuan2023-af, Qi2023-bj} use both visual and tactile sensing. Also, \textcite{Chen2022-rc} used dense object feedback with extensive domain randomization to achieve arbitrary reorientation of wide range of objects. However, we primarily focus on improving exploration and learning policies for more difficult tasks, such as in-hand manipulation of non-convex and large shapes, with only intrinsic sensing. We also achieve successful and robust sim-to-real transfer without extensive domain randomization or domain adaptation by closing the sim-to-real gap via tactile feedback \cite{Khandate2023-fc}.

\section{Exploration with sub-skill controllers}

Other prior turn to sub-skill controllers for improving efficiency as challenging dexterous skills can often be decomposed into sub-skills. Many works aim to combine such sub-skill controllers, often possible to use model-based or analytical controllers, to learn a hierarchical policy. \textcite{Veiga2020-zm} learn a higher level policy through RL, while having the low level controllers exclusively deal with tactile feedback. \textcite{Li2019-fq} learn 2D re-orientation using model-based controllers to ensure grasp stability in simulation. 

These approaches all share a common limitation -- the low-level sub-skill controllers are necessary not only during training but also during deployment. This restricts the feedback used by low-level controllers to the sensing modalities available on the hand while simultaneously requiring the controller to be robust to the sim-to-real gap. 

Hence, alternate approaches towards learning an end-to-end policy with the guidance from controllers have been proposed. Popular approach in this direction is to provide sub-skill controller guidance through behavior policy of off-policy RL. \textcite{Kurenkov2019-ye} proposed AC-Teach framework to sample from multiple experts based on the critic's estimate of the action-value for non-prehensile manipulation. \textcite{Jeong2020-pt} derived REQ to better utilize the off-policy transitions sampled from a sub-skill expert. \textcite{Zhang_undated-ad} propose value-based policy-composition. In our work \cite{Khandate2024-ab}, we use too AC-Teach off-policy learning framework, as it provides an effective method to incorporate sub-skill controllers for learning an end-to-end policy, and show that simple controllers and highly sub-optimal controllers can enable exploration even for highly dexterous motor control tasks such as stable in-hand manipulation.

Nonetheless, improving exploration with sub-skill controller guidance is limited in scope. The sub-skills controllers provided may not cover the full span sub-skills, especially for highly challenging dexterous skills, and also the sub-skills can themselves be hard to design. Thus, it is necessary to consider more fundamental approaches that circumvent the limitations of RL for dexterous skills i.e. random  exploration. As introduced, we must consider structured exploration techniques in RL. We discus relevant literature to this next. 

\section{Structured exploration for reinforcement learning}
The challenge of exploration is well known in RL and various schemes have been proposed to address it. Some proposed exploration methods include using intrinsic rewards \cite{Pathak2017-ik, Haarnoja2018-zj} or improving action consistency via temporally correlated noise in policy actions \cite{Amin2021-dm} or parameter space noise \cite{Plappert2017-gc}.  However, these exploration methods operate under the strict assumption that the learning agent cannot teleport between states, mimicking the constraints of the real world.

Fortunately, in cases where the policies are primarily trained in simulation, this requirement can be relaxed, and we can use our knowledge of the relevant state space to design effective exploration strategies. A number of these methods improve exploration by injecting useful states into the reset distribution during training. \textcite{Nair2017-nu} use states from human demonstrations in a block stacking task, while Ecoffet et al. \cite{Ecoffet2019-lk, Ecoffet2021-xs} use states previously visited by the learning agent itself for problems such as Atari games and robot motion planning. \textcite{Tavakoli2018-ah} evaluate various schemes for maintaining and resetting from the buffer of visited states. However, these schemes were evaluated only on benchmark continuous control tasks \cite{Duan2016-em}. From a theoretical perspective, \textcite{Agarwal2020-pp} show that a favorable reset state distribution provides a means to circumvent worst-case exploration issues using sample complexity analysis of policy gradients. 

Finding feasible trajectories through a complex state space is a well-studied motion planning problem. Of particular interest to us are sampling-based methods such as Rapidly-exploring Random Trees (RRT) \cite{LaValle1998-kn, Karaman2010-zi, Webb2013-nt} and Probabilistic Road Maps (PRM) \cite{Kavraki1996-gr, Kavraki1998-wl}. These families of methods have proven highly effective and are still being expanded. Stable Sparse-RRT (SST) and its optimal variant SST* \cite{Li2021-lv} are examples of recent sampling-based methods for high-dimensional motion planning with physics. However, the goal of these methods is finding (kinodynamic) trajectories between known start and goal states rather than closed-loop control policies which can handle deviations from the expected states.

Several approaches have tried to combine the exploratory ability of SBP with RL, leveraging planning for global exploration while learning a local control policy via RL \cite{Chiang2019-vm,Francis2020-qe,Schramm2022-hs}. These methods were primarily developed for and tested on navigation tasks, where nearby state space samples are generally easy to connect by an RL agent acting as a local planner. The LeaPER algorithm \cite{Pinto2018-xi} also uses plans obtained by RRT as reset state distribution and learns policies for simple non-prehensile manipulation. However, the state space for the prehensile in-hand manipulation tasks we show here is highly constrained, with small useful regions and non-holonomic transitions. Other approaches use trajectories planned by SBP as expert demonstrations. \textcite{Morere2020-gq} recommend using a policy trained with SBP as expert demonstrations as an initial policy. Alternatively, \textcite{Jurgenson2019-ye} and \textcite{Ha2020-od} use planned trajectories in the replay buffer of an off-policy RL agent for multi-arm motion planning. First, these methods requires that planned trajectories also include the actions used to achieve transitions, which SBP does not always provide. Next, it is unclear how off-policy RL can be combined with the extensive physics parallelism that has been vital in the recent success of on-policy methods for learning manipulation \cite{Allshire2021-qp, Makoviychuk2021-ko, Chen2021-ig}. 

\section{Dexterous manipulation from robot-demonstrations}
\llabel{sec:IL}

Learning from demonstrations \cite{Schaal1996-us} is a popular approach for achieving dexterous skills. These efforts primarily differ in the setups for collecting robot demonstrations and the learning algorithms employed to extract policies.

Robot demonstration can be collected using two broadly different approaches. Human hand pose can mapped to robot pose in real-time and used for teleoperation \cite{Handa2019-me, Qin2023-bx, Arunachalam2022-qh} of robot in simulation or the real robot.  When demonstrations are collected on the real robot, it helps mitigate the sim-to-real gap later when used to extract policies.

Imitation learning \cite{Florence2022-hw, Shafiullah_undated-gu, Chi2023-vp}is a popular approach for achieving multi-fingered manipulation. DIME \cite{Arunachalam2022-qh} and other similar methods \cite{Qin2023-bx} then standard imitation learning techniques to demonstrate dexterous manipulation skills.  \textcite{Guzey2023-zq} use demonstrations consisting of tactile sensing on the robot and achieve dexterous skills. On the other hand, when reward-labeled demonstrations are available.

Offline RL \cite{Levine2020-pe} can also be used to achieve dexterous manipulation \textcite{Nair2020-gq}, but it is not widely adopted primarily due to the difficulty of reward labels in robot demonstrations. However, if reward labels are available offline, RL can be advantageous as it can improve upon \cite{Kumar2020-cy, Kumar2021-tk} the often sub-optimal trajectories in the demonstrations.

One of the critical advantages of learning from demonstrations is that it enables the development of dexterous skills for long-horizon tasks. These tasks often involve a sequence of actions, such as grasping, in-hand reorientation, object transfer, and re-grasping. While model-free reinforcement learning (RL) has facilitated the acquisition of various dexterous skills, it has proven less effective for such complex, multi-step tasks. Another significant advantage of learning from demonstrations, particularly when these demonstrations are performed on a real robot, is the ability to circumvent the sim-to-real gap that is commonly encountered in reinforcement learning-based approaches.

There are several limitations to imitation learning as well. One major challenge is the distribution shift, which necessitates a large number of demonstrations to mitigate its effects. This challenge likely contributes to the difficulty in achieving tasks that require a high degree of coordination, such as in-hand reorientation. Additionally, there is a fundamental limitation due to the inherent constraints on the dexterity of demonstrations that can be collected. Multi-fingered dexterity, where tactile sensing plays a critical role, exemplifies this issue. Without tactile feedback being rendered back to the human operator, it becomes impossible to collect accurate demonstrations for such tasks.

\section{Learning from human demonstrations}

Directly learning from human demonstrations—without the need for teleoperation of the robot—offers a promising avenue for enabling imitation learning of dexterous skills that require tactile feedback. This approach allows for demonstrations that involve complex multi-fingered dexterity, such as in-hand reorientation. Additionally, because human demonstrations are easier to collect, it becomes feasible to gather a large number of them, improving training.

Learning from human demonstrations typically involves learning from videos, a longstanding area of research with approaches ranging from reinforcement learning and imitation learning to planning with world models. After briefly reviewing imitation learning from human videos more broadly, we will delve into the relevant work on learning from visuo-tactile human demonstrations.

One of the common ways is to use videos to learn a reward model for reinforcement learning.  \textcite{Shao2021-pm} learn a classify human demonstrations into task categories and use the classification score as a reward for training a robot policy. \textcite{Chen2021-zp} learn discriminator to predict if two videos are completing the same task or not. Using the similarity score as a reward alongside action conditioned video prediction model for planning, a new task based on human demonstration can be executed. ViPER \cite{Escontrela2023-mm} train video prediction transformer and use log-likelihood as reward functions.

Alternatively, other works use imitation learning. Many propose a frameworks for imitation learning at the level of semantic skills and use it for cross-domain imitation learning \textcite{Pertsch2022-yp, Xu2023-jg, Zakka2021-in, Xu2023-dd}. 

To leverage large-scale human video datasets such as SomethingSomething-V2 \cite{Goyal2017-xg, noauthor_undated-oj} and Ego4D \cite{Grauman2021-vl} several prior works use self-supervised learning to learn visual representations and demonstrate benefits for downstream policy learning. \textcite{Parisi2022-ea} use MoCo representations and R3M \cite{Nair2022-ss} uses MAE for learning such representations. Other works predict affordance \cite{Bahl2023-ud} from human videos. Such methods are also limited to relatively simple non-prehensile manipulation tasks.

Towards extracting action representation from video, several recent works have proposed learning world models. \textcite{Mendonca2023-lh} learn a "structured-world model" using an affordance action space and employ model-predictive control for deployment. Similarly, \textcite{Yuan2024-at} propose predicting object 3D flow from RGBD data.

While the above approaches considered relatively simple non-prehensile manipulation tasks, other works have focused on multi-fingered manipulation from human videos. \textcite{Qin2022-zb} demonstrated that human hand and object poses can be extracted from a video to facilitate simulations. VideoDex \cite{Shaw2022-pk} uses explicit pose retargeting of human to robot hand and Neural Dynamic Policies \cite{Bahl2020-cv} for imitation. \textcite{Haldar2023-cd} imitate by using optimal transport matching with demonstrations. As human video data for dexterous manipulation can be hard to collect, Mimic-Play \cite{Wang2023-ka} instead proposes using human hand-in-play data, which is much easier to collect.

The embodiment gap between the human and robot can be particularly acute for multi-fingered manipulation, limiting the success of explicit retargeting-based imitation learning. Therefore, we need to extract actions in the latent space without labeled actions. \textcite{EdwardsUnknown-lz} proposed simultaneously learning a world model and a latent policy from only visual demonstrations. \textcite{Schmidt2023-me} learn a forward model while also learning an inverse dynamics model with an information bottleneck. \textcite{Yang2024-sc} design a shared latent space to learn cross-embodiment policies across manipulation and navigation tasks for robotic arms and drones. However, these methods consider either simple simulated control tasks or non-prehensile manipulation. We focus on learning latent actions for multi-fingered dexterity.

Visual demonstrations alone, without tactile sensing, are not very helpful for challenging multi-fingered dexterous skills such as in-hand reorientation. Hence, we aim to collect tactile readings alongside visual demonstrations to achieve dexterous contact-rich skills via human demonstrations. Several devices and gloves have been devised to collect tactile feedback \cite{Ruppel2024-ew, Sundaram2019-ju, Sundaram2019-qn, Wei2023-rc, Sagisaka2012-es, Martin2004-dl, Sagisaka2011-tu, Yeo2016-mn, Kim2010-fz, McCaw2018-zv}, but they do not use it for extracting robot skills. \textcite{Wei2023-rc} build a hand exoskeleton device to achieve high-quality demonstrations for robots but do not demonstrate dexterous skills beyond simple in-hand manipulation. In contrast to these setups, we equip the human hand with low-cost tactile sensors. Additionally, we extract visuo-tactile representations to enable imitation learning with a small number of robot demonstrations.


\chapter{Exploration with State Priors}

In this work, we focus on learning finger-gaiting (manipulation involving finger substitution and re-grasping) and finger-pivoting (manipulation involving the object in hinge-grasp) skills. Both skills are important towards enabling large-angle in-hand object re-orientation: achieving an arbitrarily large rotation of the grasped object around a given axis, up to or even exceeding a full revolution. Such a task is generally not achievable by in-grasp manipulation (i.e. without breaking the contacts of the original grasp) and requires finger-gaiting or finger-pivoting (i.e. breaking and re-establishing contacts during manipulation); these are not restricted by the kinematic constraints of the hand and can achieve potentially limitless object re-orientation.

We are interested in achieving these skills exclusively through using fingertip grasps (i.e precision in-hand manipulation \cite{Michelman1998-ye}) without requiring the presence of the palm underneath the object, which enables the policies to be used in arbitrary orientations of the hand. However, the task of learning to manipulate only via such precision grasps is a significantly harder problem: action randomization, responsible for exploration in RL, often fails as the hand can easily drop the object.

Furthermore, we would like to circumvent the need for cumbersome external sensing by only using internal sensing in achieving these skills. The challenge here is that the absence of external sensing implies we do not have information regarding the object such as its global shape and pose. However, we posit that internal sensing by itself can provide object information sufficient towards our goal. 

We set out to determine if we can even achieve finger-gaiting and finger-pivoting skills purely through intrinsic sensing in simulation, where we evaluate both proprioceptive feedback and tactile feedback. To this end, we consider the task of continuous object re-orientation about a given axis, aiming to learn finger-gaiting and finger-pivoting without object pose information. With this approach, we hope to learn policies to rotate object about cardinal axes and combine them for arbitrary in-hand object re-orientation. To overcome challenges in exploration, we propose collecting training trajectories starting from a wide range of grasps sampled from appropriately designed initial state distributions as an alternative exploration mechanism. We summarize the contributions of this work as follows:

\begin{enumerate}
\item We learn finger-gaiting and finger-pivoting policies that can achieve large angle in-hand re-orientation of a range of simulated objects. Our policies learn to grasp and manipulate only via precision fingertip grasps using a highly dexterous and fully actuated hand, allowing us to keep the object in a stable grasp without the need for passive support at any instance during manipulation.
\item We are the first to achieve these skills by making use of only intrinsic sensing such as proprioception and touch, while also generalizing to multiple object shapes.
\item We present an exhaustive analysis of the importance of different internal sensor feedback for learning finger-gaiting and finger-pivoting policies in a simulated environment using our approach.
\end{enumerate}

\llabel{ch:resetdist}

\section{Learning precision in-hand re-orientation}
We address two important challenges for precision in-hand reorientation using reinforcement learning. First, we propose a hand-centric decomposition method for achieving arbitrary in-hand reorientation in an object-agnostic fashion. Next, we identify that a key challenge of exploration for learning precision in-hand manipulation skills can be alleviated by collecting training trajectories starting at varied stable grasps. We use these grasps to design appropriate initial state distributions for training. Our approach assumes a fully-actuated and position-controlled (torque-limited) hand.

\section{Hand-centric decomposition}
\llabel{sec:decompostion}
Our aim is to push the limits on manipulation with only intrinsic sensing, and do this in a general fashion without assuming object knowledge. Thus, we do so in a hand-centric way: we learn to rotate around axes grounded in the hand frame. This means we do not need external tracking (which presumably needs to be trained for each individual object) to provide object-pose.\footnote{We note that there exist applications where specific object poses are needed at a task level, and for such cases we envision future work where a high-level object-specific tracker makes use of our hand-centric object-agnostic policies to achieve it.} We also find that rewarding angular velocity about desired axis of rotation is conducive to learning finger-gaiting and finger-pivoting policies. However, learning a single policy for any arbitrary axis is challenging as it involves learning goal-conditioned policies, which is difficult for model-free RL.


\begin{figure}[h]
    \centering
    \includegraphics[width=0.8\columnwidth]{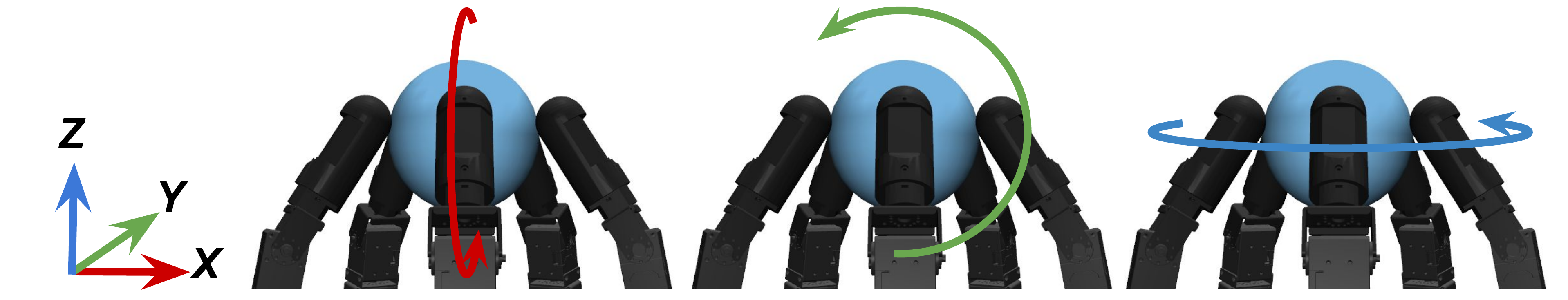}
    \caption{Hand-centric decomposition of in-hand re-orientation into re-orientation about cardinal axes.}
    \llabel{fig:xyz_policies}
    \vspace{-5pt}
\end{figure}

Our proposed method for large-angle arbitrary in-hand reorientation is thus to decompose the problem of achieving arbitrary angular velocity of the object into learning separate policies about the cardinal axes as shown in Fig.~ \lref{fig:xyz_policies}. The finger-gaiting policies obtained for each axis can then be combined in the appropriate sequence to achieve the desired change in object orientation, while sidestepping the difficulty of learning a goal-conditioned policy. 

We assume that proprioceptive sensing can provide current positions $\boldsymbol{q}$ and controller set-point positions $\boldsymbol{q}^s$. We note that the combination of desired positions and current positions can be considered as a proxy for motor forces, if the characteristics of the underlying controller are fixed. More importantly, we assume tactile sensing to provide absolute contact positions $\boldsymbol{p}_i \in \mathbb{R}^3$ and magnitude of contact forces $c_i \in \mathbb{R}$ at each fingertip $i$. With known fingertip geometry, the contact normals $\boldsymbol{n}_i \in \mathbb{R}^3$  can be derived from contact positions $\boldsymbol{p}_i$. 

Our axis-specific re-orientation policies are conditioned only on proprioceptive and tactile feedback as given by the observation vector $\mathbf{o}$:
\begin{equation}
    \mathbf{o} = [\boldsymbol{q}, \boldsymbol{q}^s, \boldsymbol{p}_1 \ldots \boldsymbol{p}_m, c_1 \ldots c_m , \boldsymbol{n}_1 \ldots \boldsymbol{n}_m]
    \llabel{eq:observation}
\end{equation}
where $m$ is the number of fingers. Our policies command set-point changes $\Delta \mathbf{q}_d$. 

\section{Learning axis-specific re-orientation}

\begin{figure}[t]
    \centering
    \includegraphics[width=0.3\columnwidth]{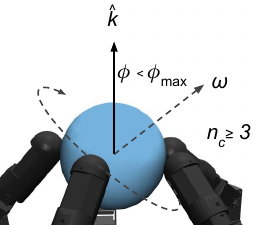}
    \caption{Learning axis conditional continuous re-orientation $\hat{\boldsymbol{k}}$. We use the component of angular velocity $\boldsymbol{\omega}$ about $\hat{\boldsymbol{k}}$ as reward when the object is in a grasp with 3 or more fingertips, i.e $n_c \geq 3$.}
    \llabel{fig:reward}
    \vspace{-5pt}
\end{figure}

We now describe the procedure for learning in-hand reorientation policies for an arbitrary but fixed axis. Let $\hat{\boldsymbol{k}}$ be the desired axis of rotation. 
To learn axis-specific policy $\pi^{\hat{\boldsymbol{k}}}$ that continuously re-orients the object about the desired axis $\hat{\boldsymbol{k}}$ we use the object's angular velocity $\boldsymbol{\omega}$ along $\hat{\boldsymbol{k}}$ as reward as shown in Fig \lref{fig:reward}. However, to ensure that the policy learns to only use precision fingertip grasps to reorient the object, we provide this reward if only fingertips are in contact with the object. In addition, we require that at least 3 fingertips are in contact with the object. In addition, we encourage alignment of the object's axis of rotation with the desired axis by requiring the separation to be limited to $\phi_{max}$.

\setlength{\arraycolsep}{0.0em}
\begin{eqnarray}
r&{}={}&\min(r_{max}, {\boldsymbol{\omega}}.\hat{\boldsymbol{k}}) \ \mathbf{I}[n_c \geq 3 \land \phi \leq \phi_{max} ]\nonumber\\
&&{+}\:\min(0, {\boldsymbol{\omega}}.\hat{\boldsymbol{k}}) \ \mathbf{I}[n_c < 3 \vee \phi> \phi_{max} ]
\llabel{eq:reward}
\end{eqnarray}
\setlength{\arraycolsep}{5pt}

The reward function is described in (\lref{eq:reward}), where $n_c$ is the number of fingertip contacts and $\phi$ is the separation between the desired and current axis of rotation. Symbols $\land$, $\lor$, $\mathbf{I}$ are the logical \textit{and}, the logical \textit{or}, and indicator function, respectively. Notice that we also use reward clipping to avoid local optima and idiosyncratic behaviors. In our setup, $r_{max}$ and $\phi_{max}$ are both set to 0.5. Although the reward uses the object's angular velocity, we do not need additional sensing to measure it as we only train in simulation with the intent of zero-shot transfer to hardware.


\section{Enabling exploration with human-provided state-priors}
\llabel{sec:exploration}
A key issue in using reinforcement learning for learning precision in-hand manipulation skills is that a random exploratory action can easily disturb the stability of the object held in a precision grasp, causing it to be dropped. This difficulty is particularly acute for finger-gaiting, which requires fingertips to break contact with the object and transition between different grasps, involving different fingertips, all while re-orienting the object. Intuitively, the likelihood of selecting a sequence of random actions that can accomplish this feat and obtain a useful reward signal is very low.


For a policy to learn finger-gaiting, it must encounter these diverse grasps within its training samples so that the policy's action distributions can improve at these states.
Consider taking a sequence of random actions starting from a stable $l$-finger grasp. While it is possible to reach a stable grasp with an additional finger in contact (if available), it is more likely to lose one finger contact, then another and so on until the object is dropped. Over multiple trials, we can expect to encounter most combinations of $l-1$ grasps. In this setting, it can be argued that starting from a stable grasp with all $m$ fingers in contact leads to maximum exploration. Interestingly, as we will demonstrate in Sec \lref{sec:expts}, we found this to be insufficient.

Our insight is to observe that through domain knowledge we are already aware of the states that a sufficiently exploratory policy might visit. Using domain knowledge in designing initial distributions is a known way of improving sample complexity \cite{Kakade2003-ja}\cite{De_Farias2003-ai}. Thus, we use our knowledge of relevant states in designing the initial states used for episode rollouts and show that it is critical for learning precision finger-gaiting and finger-pivoting. 

We propose sampling sufficiently-varied stable grasps relevant to re-orienting the object about the desired axis and use them as initial states for collecting training trajectories. These grasps must be well distributed in terms of number of contacts, contact positions relative to the object, and object poses relevant to the task. To this end, we first initialize the object in an random pose and then sample fingertip positions until we find a stable grasp as described in Stable Grasp Sampling (SGS) in Alg. \lref{algo:sampling}.

\begin{figure}[t]
    \centering
    \includegraphics[width=0.8\columnwidth]{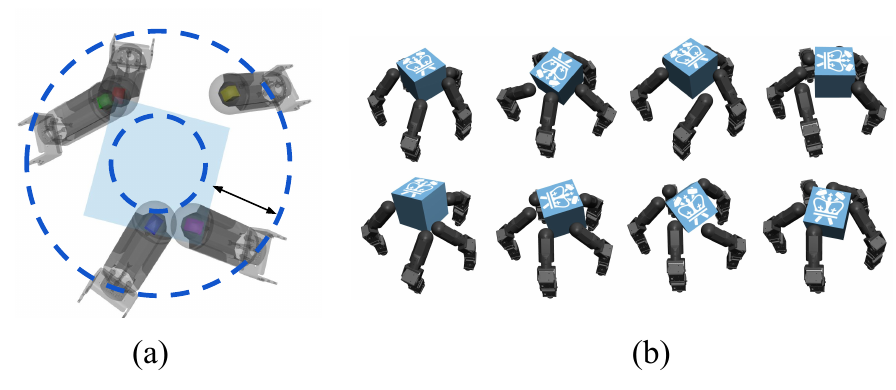}
    \caption{(a) Sampling fingertips around the object. (b) Diverse relevant initial grasps sampled for efficient exploration.}
    \llabel{fig:init_dist}
    \vspace{-5pt}
\end{figure}

\begin{algorithm}[t]
    \caption{Stable Grasp Sampling (SGS)} \llabel{algo:sampling}
    \textbf{Input:}$\rho_{obj}$, $\rho_{hand}$, $t_s$, $n_{c,\min}$ \Comment{object pose distribution, hand pose distribution, simulation settling time, minimum number of contacts} \\
    \textbf{Output:} $\mathbf{s}_{g}$ \Comment{simulator state of the sampled grasp} 
    \begin{algorithmic}[1]
    \Repeat
    \State Sample object and hand pose: $\boldsymbol{x}_s$ $\sim$ $\rho_{obj}$, $\boldsymbol{q}_s$ $\sim$ $\rho_{hand}$
    \State Set object pose in the simulator with $\boldsymbol{x}_s$
    \State Set joint positions and controller set-points with $\boldsymbol{q}_s$
    \State Step the simulation forward by $t_s$ seconds
    \State Find number of fingertips in contact with object, $n_c$
    \Until{$n_c \geq n_{c,\min}$}
    \State Save simulator state as $\boldsymbol{s}_{g}$
    \end{algorithmic}
\end{algorithm}

In SGS, we first sample an object pose and a hand pose, then update the simulator with the sampled poses towards obtaining a grasp. We advance the simulation for a short duration, $t_s$, to let any transients die down. If the object has settled into a grasp with at least two contacts, the pose is used for collecting training trajectories. Note that the fingertips could be overlapping with the object or with each other as we do not explicitly check this. However, due to the soft-contact model used by the simulator (MuJoCo \cite{Todorov2012-xe}) the inter-penetrations are resolved during simulation.  An illustrative set of grasps sampled by SGS are shown in Fig \lref{fig:init_dist}b.

To sample the hand pose, we start by sampling fingertip locations within an annulus centered on and partially overlaps with the object (Fig \lref{fig:init_dist}a). Thus, the probabilities of each fingertip making contact with the object and of staying free are roughly the same. With this procedure, not only do we find stable grasps relevant to finger-gaiting and finger-pivoting, we improve the likelihood of discovering them, thus minimizing training wall-clock time.

\section{Experiments and Results}

For evaluating our method, we focus on learning precision in-hand re-orientation about the z- and x- axes for a range of regular object shapes. (The y-axis is similar to x-, given the symmetry of our hand model.) Our object set, which consists of a cylinder, sphere, icosahedron, dodecahedron and cube, is designed so that we have objects of varying difficulty with the sphere and cube being the easiest and hardest, respectively. For training, we use PPO \cite{Schulman2017-td}. We chose PPO over other state-of-the-art methods such as SAC primarily for training stability .

For the following analysis, we use z-axis re-orientation as a case study. In addition to the above, we also train z-axis re-orientation policies without assuming joint set-point feedback $\mathbf{q}_d$. For all these policies, we study their robustness properties by adding noise and also by applying perturbation forces on the object (Sec \lref{sec:robust}). We also study the zero-shot generalization properties of these policies (Sec \lref{sec:gen}). Finally, through ablation studies we present a detailed analysis ascertaining the importance of different components of feedback for achieving finger-pivoting (Sec \lref{sec:fb_study}).

We note that, in simulation, the combination of $\mathbf{q}_d$ and $\mathbf{q}$ can be considered a good proxy for torque, since simulated controllers have stable and known stiffness. However, this feature might not transfer to a real hand, where transmissions exhibit friction, stiction and other hard to model effects. We thus evaluate our policies both with and without joint set-point observations.

\begin{figure}[h]
  \centering
  \includegraphics[clip, width=\columnwidth]{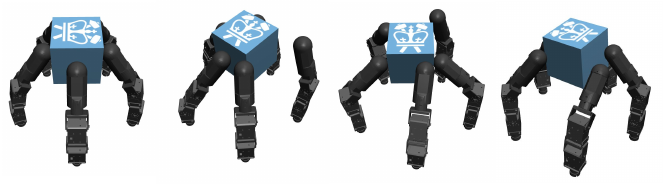}
  \caption{A learned finger-gaiting policy that can continuously re-orient the target object about the hand z-axis. The policy only uses sensing modalities intrinsic to the hand (such as touch and proprioception), and does not require explicit object pose information from external sensors.}
\end{figure}

\begin{figure*}[ht!]
    \centering
    \includegraphics[clip, width=0.9\textwidth]{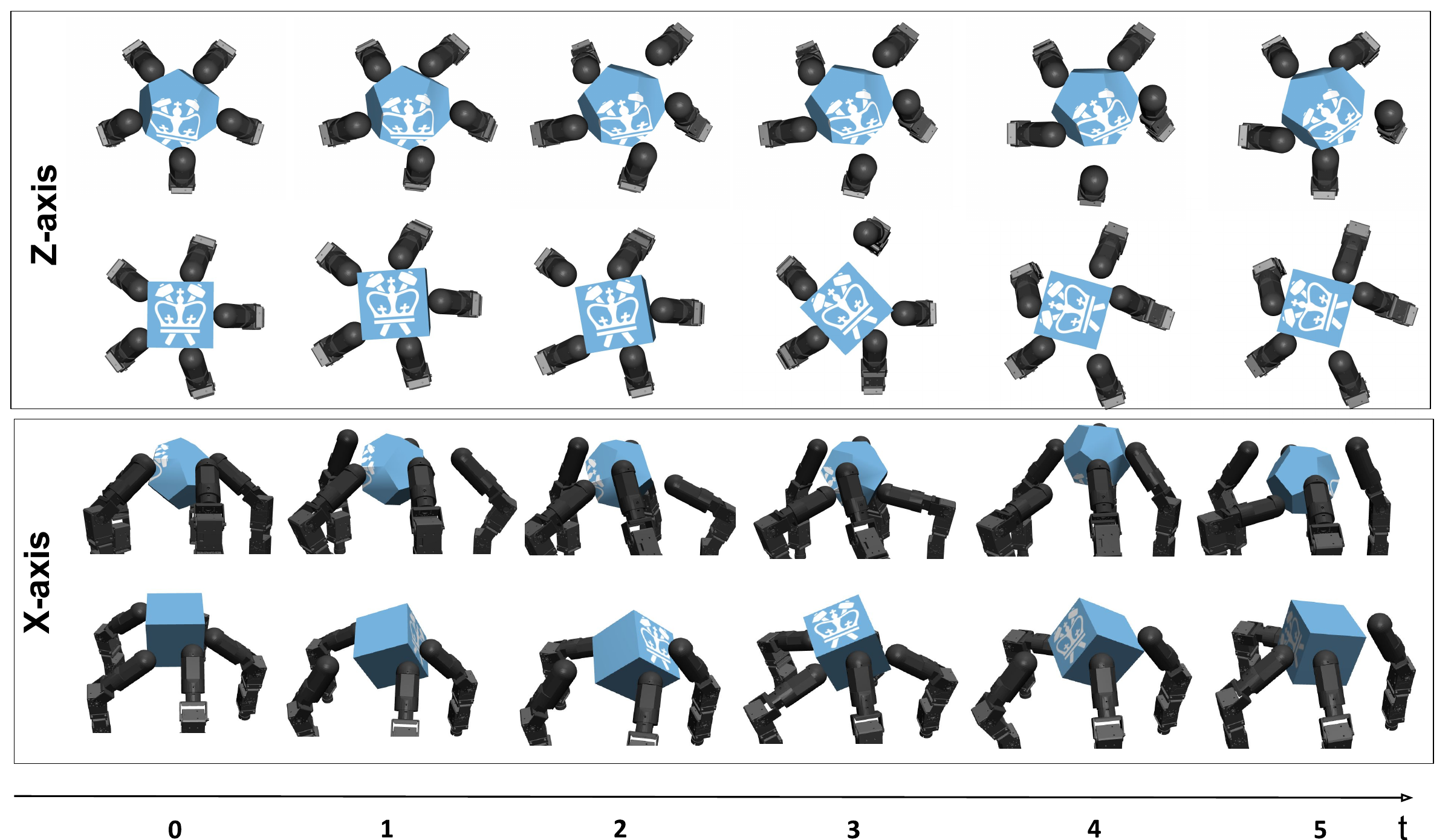}
    \caption{Finger-gaiting and finger-pivoting our policies achieve to re-orient about z-axis and x-axis respectively. Key frames are shown for two objects, dodecahedron and cube. }
    \llabel{fig:gaitframes}
\end{figure*}

\begin{figure}[t]
    \centering
    \includegraphics[clip, width=0.6\columnwidth]{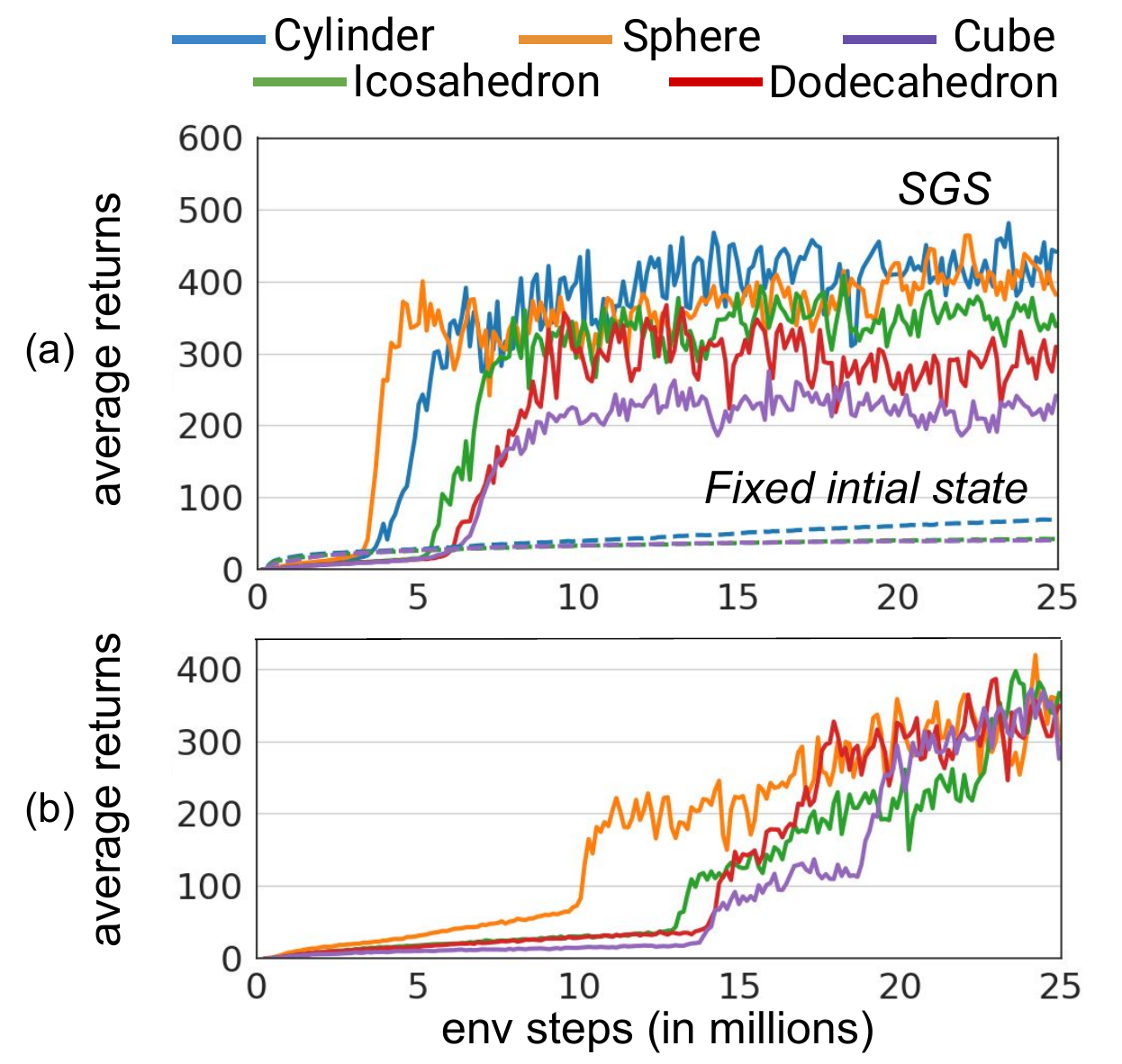}
    \caption{Average returns for (a) z-axis re-orientation and (b) x-axis re-orientation. Learning with wide range of initial grasps sampled via SGS succeeds, while using a fixed initial state fails.}
    \llabel{fig:train_return}
    \vspace{-10pt}
\end{figure}

Fig \lref{fig:train_return}a shows the learning curves for object re-orientation about the z-axis for a range of objects from using our method of sampling stable initial grasps to improve exploration. We also show learning curves using a fixed initial state (grasp with all fingers) for representative objects. First, we notice that the latter approach does not succeed. These policies only achieve small re-orientation via in-grasp manipulation and drop the object after maximum re-orientation achievable without breaking contacts. 
\llabel{sec:expts}
However, when using a wide initial distribution of grasps (sampled via SGS), the policies learn finger-gaiting and achieve continuous re-orientation of the object with significantly higher returns. With our approach, we also learn finger-pivoting for re-orientation about the x-axis, with learning curves shown in Fig \lref{fig:train_return}b. Thus, we empirically see that using a wide initial distribution consisting of relevant grasps is critical for learning continuous in-hand re-orientation and that our method results in superior sample-complexity over the state-of-the-art i.e PPO without the use of initial state distribution. Fig \lref{fig:gaitframes} shows our finger-gaiting and finger-pivoting policies performing continuous object re-orientation about z-axis and x-axis respectively.


As expected, difficulty of rotating the objects increases as we consider objects of lower rotational symmetry from sphere to cube. In the training curves in Fig \lref{fig:train_return}, we can observe this trend not only in the final returns achieved by the respective policies, but also in the number of samples required to learn continuous re-orientation. We also successfully learn policies for in-hand re-orientation without joint set-point position feedback, but these policies achieve slightly lower returns. However, they may have interesting consequences for generalization as we discussed in Sec \lref{sec:gen}.

\subsection{Robustness}
\llabel{sec:robust}
Fig. \lref{fig:robustness} shows the performance of our policy for the most difficult object in our set (cube) as we artificially add white noise with increasing variance to different sensors' feedback. We also increasingly add perturbation forces on the object. Overall, we notice that our policies are robust to noise and perturbation forces of magnitudes that can be expected on a real hand.

In particular, our policies show little drop in performance for noise in joint positions, but are more sensitive to noise in contact feedback. Nevertheless, they are still robust, and achieve high returns even at $5$mm error in contact position and $25\%$ error in contact force.  Interestingly, for noise in contact position, we found that drop in performance arises indirectly through the error in contact normal $\hat{\mathbf{t}}_n^i$ (computed from contact position $\mathbf{c}_n^i$). As for perturbation forces on the object, we observe high returns even for high perturbation forces ($1$N) equivalent to the weight of our objects. Our policies are robust event without joint-setpoint $\mathbf{q}_d$ feedback with similar robustness profiles.

\begin{figure}[h]
    \centering
    \includegraphics[clip,width=0.5\columnwidth]{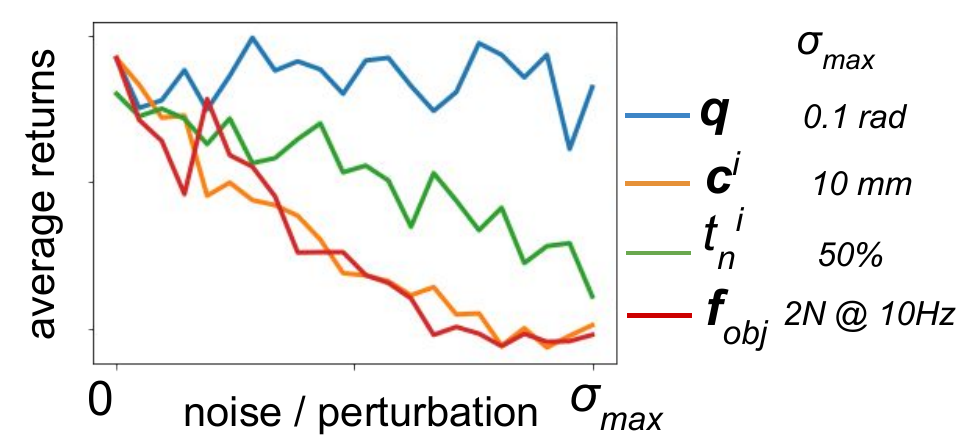}
    \caption{Robustness of our policies with increasing sensor noise and perturbation forces on the object.}
    \llabel{fig:robustness}
\end{figure}

\subsection{Generalization}
\llabel{sec:gen}
We study generalization properties of our policies by evaluating it on  different objects in the object set. We consider the transfer score, which is the ratio $R_{ij}/R_{ii}$ where $R_{ij}$ is the average returns obtained when evaluating the policy learned with object $i$ on object $j$. 

Fig. \lref{fig:cross_transfer_a} shows the cross transfer performance for policies trained with all feedback. We note that the policy trained on the sphere transfers to the cylinder and vice versa. Also, the policies trained on icosahedron and dodecahedron transfer well between themselves and also perform well on sphere and cylinder. Interestingly, the policy trained on the cube does not transfer well to the other objects. When not using joint set-point position feedback $\mathbf{q}_d$, the policy learned on the cube transfers to more objects. With no way to infer motor forces, the policy potentially learns to rely more on contact feedback which aids generalization.

\begin{figure}[h]
    \centering
    \includegraphics[clip, width=0.65\columnwidth]{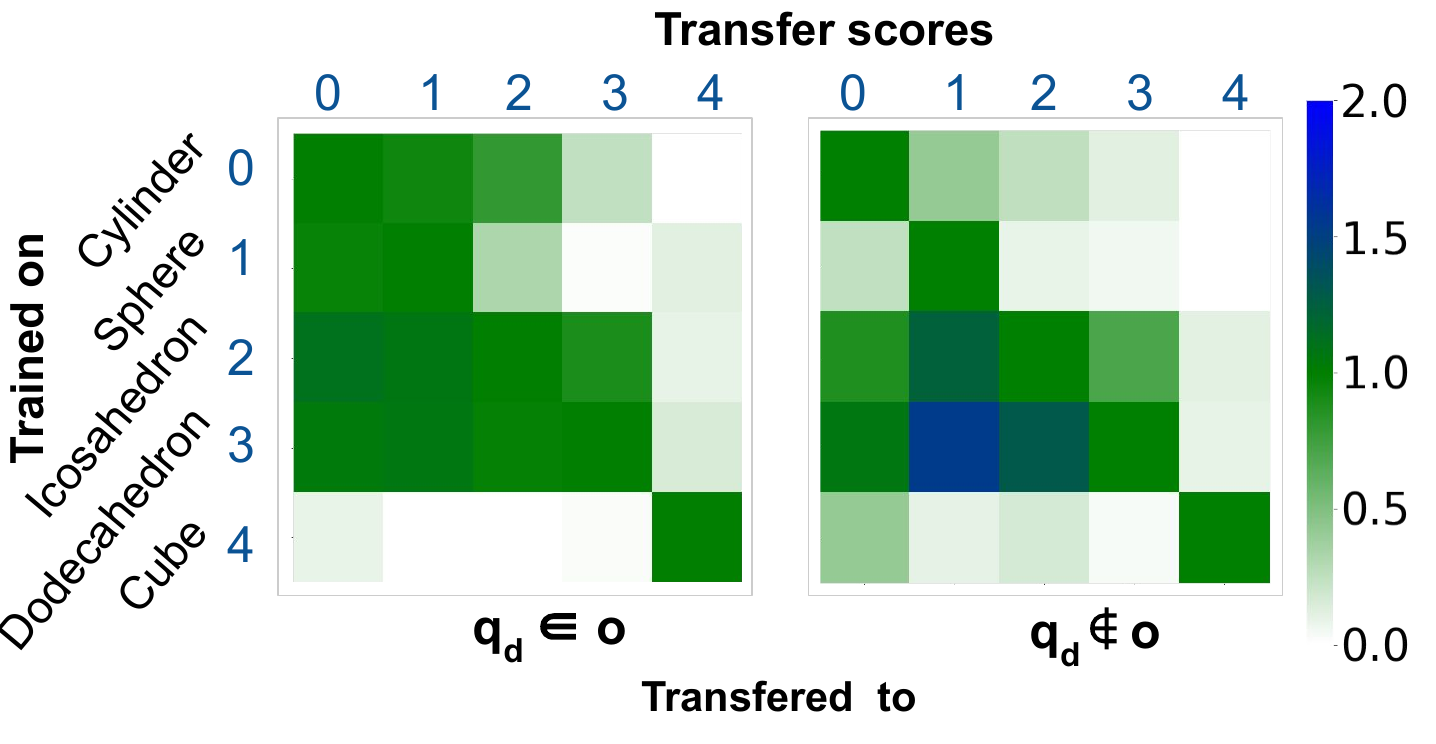}
    \caption{Cross transfer scores for policies with and without $\mathbf{q}_d$ in feedback.}
    \llabel{fig:cross_transfer_a}
    \vspace{-5pt}
\end{figure}

\subsection{Observations on feedback}
\llabel{sec:fb_study}
While our work provides some insight w.r.t the important components of our feedback through our robustness and generalization results, many interesting questions remain. We are particularly interested to discover what aspects matter most in contact feedback. To answer such questions, we run a series of ablations holding out different components.  For this, we again consider learning finger-gaiting on the cube as shown in Fig \lref{fig:fb_study}. 

Based on this ablation study, we can make a number of observations. As expected, contact feedback is essential for learning in-hand re-orientation via finger-gaiting; the policy does not learn finger-gaiting with just proprioceptive feedback (\#4). More interesting, and also more surprising, is that explicitly computing contact normal $\mathbf{t}_n^i$ and providing it as feedback is critical when excluding  joint position set-point $\mathbf{q}_d$ (\#6 to \#10). In fact, the policy learns finger-gaiting with just contact normal and joint position feedback (\#10). However, while not critical, contact position and force feedback are still beneficial as they improve sample efficiency (\#6, \#7).

\begin{figure}
    \centering
    \includegraphics[width=0.7\columnwidth]{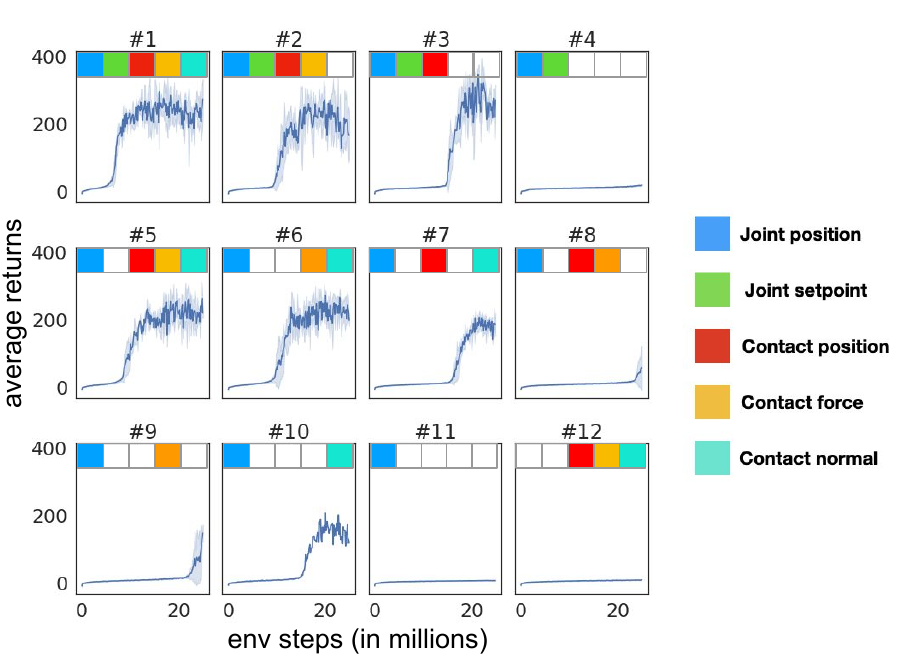}
    \caption{Ablations holding out different components of feedback. For each experiment, color filled cells in the observation vector shown above the training curve indicate which of the components of the observation vector are provided to the policy. The key takeaway from these ablations is that we found that explicitly computed contact normal feedback is critical contact feedback required by the policy to successfully learn finger-gaiting.}
    \llabel{fig:fb_study}
    \vspace{-5pt}
\end{figure}

\section{Conclusion}
In this paper, we focus on the problem of learning in-hand manipulation policies that can achieve large-angle object re-orientation via finger-gaiting. To facilitate future deployment in real scenarios, we restrict ourselves to using sensing modalities intrinsic to the hand, such as touch and proprioception, with no external vision or tracking sensor providing object-specific information. Furthermore, we aim for policies that can achieve manipulation skills without using a palm or other surfaces for passive support, and which instead need to maintain the object in a stable grasp.

A critical component of our approach is the use of appropriate initial state distributions during training, used to alleviate the intrinsic instability of precision grasping. We also decompose the manipulation problem into axis-specific rotation policies in the hand coordinate frame, allowing for object-agnostic policies. Combining these, we are able to achieve the desired skills in a simulated environment, the first instance in the literature of such policies being successfully trained with intrinsic sensor data.

We consider this work to be a useful step towards future sim-to-real transfer. To this end, we engage in an exhaustive empirical analysis of the role that each sensing modality plays in enabling our manipulation skills. Specifically, we show that tactile feedback in addition to proprioceptive sensing is critical in enabling such performance. Finally, our analysis of the policies shows that they generalize to novel objects and are also sufficiently robust to force perturbations and sensing noise, suggesting the possibility of future sim-to-real transfer. 


\chapter{Exploration with Action Priors}

In this work, we explored an alternative method for bootstrapping reinforcement learning. We notice that it simple sub-skills controller for dexterous manipulation and the actions provided by these sub-skill controllers may enable exploration.

\begin{figure}[h]
\centering
\includegraphics[width=0.45\textwidth]{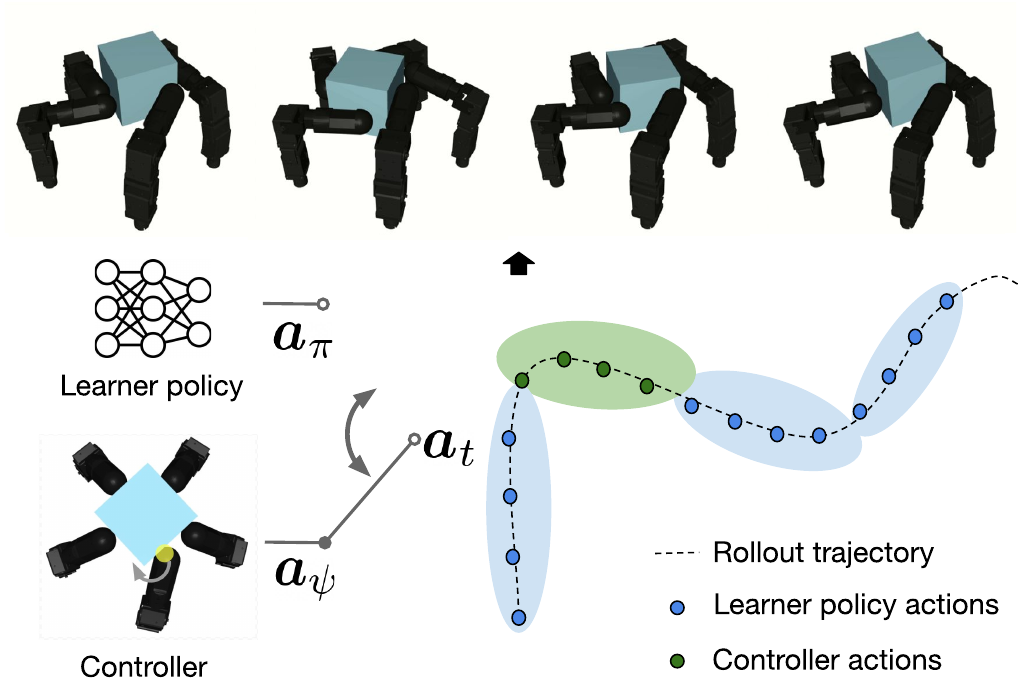}
\caption{Our method of interweaving the policy and sub-skill controller (ex. contact switching controller) during training allows the policy to effectively learn dexterous finger-gaiting skills as shown. Videos can be found at project page: \href{https://roamlab.github.io/vge}{roamlab.github.io/vge}}
\llabel{fig:eyecandy}
\end{figure}

In this work, we seek to leverage model-based controllers to improve sample efficiency while retaining the benefits of learning an end-to-end policy. The key idea that we propose is to interleave learner policies and sub-skill controllers during episode rollout during the initial phase and use only the learner policy during the later phase of training. Consequently, we use the sub-skill controllers only during training and not at run-time. We show that our method guides exploration towards relevant regions of the state-space. We also use the current action-value estimates provided by the critic to guide action selection which further improves exploration. 

Critically, we use controllers that are simple to design and inexpensive to query actions from. Our controllers do not involve multi-step horizon motion planning or other sources of complexity and computational cost. This makes them inexpensive to compute - favorably reducing the cost of querying the expert. However, inexpensive controllers will often be sub-optimal in their behavior. Still, our method enables learning despite using such sub-optimal controllers. Overall, our main contributions here include:
\begin{itemize}
    \item We demonstrate that a set of simple sub-skill controllers can be leveraged to learn dexterous in-hand manipulation tasks such as finger-gaiting.
    \item In contrast to previous work combining model-based controllers and learning for manipulation, we use the proposed controllers for training an end-to-end policy that does not require these controllers for deployment. Our approach alleviates many constraints on the controllers we use, in terms of both performance and sensory input.
    \item We also show that sub-optimal controllers can enable effective exploration of the complex state space of our problem, without exploratory reset distributions used in previous studies with similar goals. 
\end{itemize}

We will demonstrate our method for learning dexterous manipulation skills on a task exhibiting a particularly challenging problem in exploration. Specifically, we consider the task of finger-gaiting in-hand manipulation with only fingertip grasps and no support surfaces - a necessary skill for continuous object re-orientation in arbitrary orientations of the hand. Learning this skill purely with reinforcement learning is challenging as random exploration from a fixed state does not sufficiently explore the desired state-space \cite{Khandate2021-wl}. Due to the complexity and contact-rich nature of the task, it also challenging to use model-based methods to design an expert for finger-gaiting.

Finger-gaiting is a rich skill that consists of a range of behaviors that we can split into two sub-skills: in-grasp manipulation, and contact switching. In-grasp manipulation entails maintaining the object in a stable grasp and reorienting it without breaking or making contact. Therefore, due to limits on the range of motion of the hand, the maximum object rotation is also limited. Contact switching involves breaking and making contact to re-grasp the object in order to further object reorientation via in-grasp manipulation.

Thus, instead of designing a single controller for finger-gaiting we can decompose it into two controllers, an in-grasp manipulation controller and a contact switching controller. Still, ensuring optimality and robustness is challenging for these controllers. Nevertheless, sub-optimal model-based controllers obtained from simplified models or heuristics can still generate reasonable actions and are able to generate exploratory sequences of state transitions. 

In this work, we will (i) design simple controllers for in-grasp manipulation and contact switching, and (ii) learn an effective end-to-end policy for finger-gaiting using these sub-optimal controllers by following their actions.




\section{Problem definition and sub-skill controllers}
\llabel{sec:learnfg}
We consider the task of finger-gaiting in-hand manipulation by learning continuous re-orientation about z-axis in a hand-centric frame. We aim to learn such a policy by rewarding angular velocity of the object along the z-axis. Let us consider a fully dexterous hand with $m$ fingers and $d$ dofs. Assuming position control, our policy outputs joint target updates from observations $\bm{s}$:
\begin{equation}
    \bm{s} = [\bm{q}, \bm{q}', \bm{p}_1, .. , \bm{p}_m, \bm{n}_1, .., \bm{n}_m, c_1, .., c_m]
\end{equation}
where $\bm{q} \in \mathcal{R}^d$, $\bm{q}'\in \mathcal{R}^d$ are current and target joint positions, $\bm{p}_i \in \mathcal{R}^3$, $\bm{n}_i \in \mathcal{R}^3$ are contact positions, contact normal in the global coordinate frame for contacts on all fingertips and $c_i \in \mathcal{R}$ is the magnitude of contact forces. Similar to~\cite{Khandate2022-qt}, we train this policy by rewarding the angular velocity of the object about the z-axis when making at least 3 contacts between the object and the fingertips.

As previously discussed, finger-gaiting skill can be broken down into two sub-skills - in-grasp manipulation and contact switching. We will now design controllers for each.

\begin{figure}[t]
\centering
\includegraphics[width=0.4\textwidth]{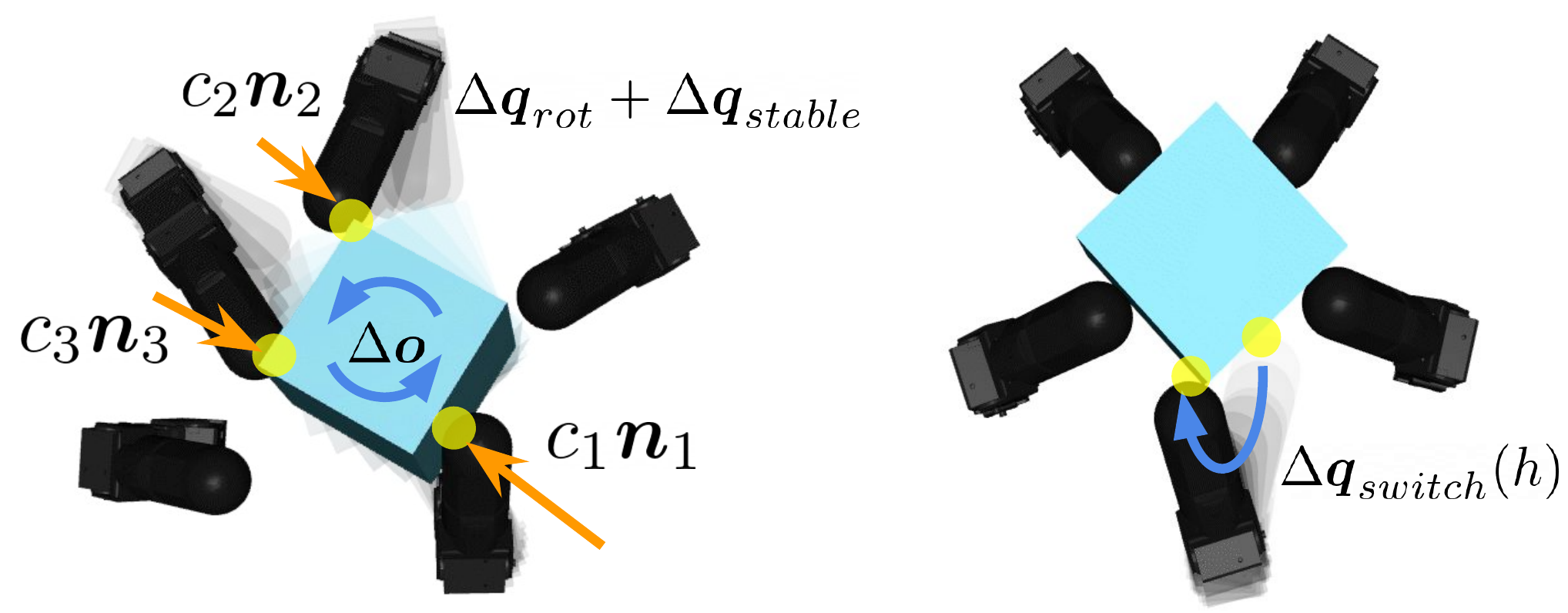}
\caption{Sub-skill controllers used for learning finger-gaiting: (left) In-grasp manipulation controller (right) Contact switching controller as described in Sec \lref{sec:learnfg}}.
\llabel{fig:experts}
\vspace{-5mm}
\end{figure}

The in-grasp manipulation sub-skill maintains a stable grasp on the object and, simultaneously, reorients the object about the desired axis. We construct this controller by superposition of two lower-level controllers - one responsible for stability and another responsible for object re-orientation. 

To maintain grasp stability, we assume at least three contacts and desire the net object wrench be close to zero. We use the kinematic model of the hand to derive our stability controller as follows. Let $S$ be the set of contact points and $\bm{n}_1, \ldots, \bm{n}_k$ ($k\geq3$) be the contact normals. $\bm{G}_S$ and $\bm{J}_S$ are the grasp map matrix and the Jacobian for the set of contacts.  Our stability controller keeps the object in equilibrium by minimizing the net wrench applied to the object. This can be achieved by solving the following Quadratic Program (QP):
\begin{align}
\text{unknowns: normal}&\text{ force magnitudes }c_i,~i=1 \ldots k \nonumber \\
\text{minimize}~\|\bm{w}\|&~\text{subject to:} \nonumber \\
\bm{w} &= \bm{G}_S^T \left[ c_1 \bm{n_1} \ldots c_k \bm{n_k} \right]^T \llabel{eq:stabstart}\\
c_i & \geq  0~\forall i \\
\exists j~\text{such that}~c_j &= 1~\text{(ensure non-zero solution)}
\end{align}

Let $c_i^*$'s be the desired contact forces obtained as solution of above QP. We can compute the actions $\Delta q_{stable}$ needed for stability (expressed as joint position setpoint changes) by solving the following system: 
\begin{eqnarray}
    \Delta \bm{p}_i &=& \alpha (c^*_i - c_i)\bm{n}_i\\
    \bm{J}_S \Delta q_{stable} &=& [\Delta \bm{p}_1, \ldots, \Delta \bm{p}_k]^T 
    \llabel{eq:dqstab}
\end{eqnarray}
where $\alpha$ is the stability controller gain.

Separately, let $\Delta \bm{o} \in \mathcal{R}^6$  denote the desired change in object pose. We can compute the action $\Delta \bm{q}_{rot}$ required to  achieve this object re-orientation by solving
\begin{align}
    \bm{J}_S \Delta \bm{q}_{rot} = \bm{G}_S^{T} \Delta {\bm{o}}
    \llabel{eq:dqrot}
\end{align}

Finally, $\Delta \bm{q}_{stable}$ and $\Delta \bm{q}_{rot}$ are combined together simply via superposition. The complete set of steps involved in the in-grasp manipulation controller is outlined in Alg.~\lref{algo:ingrasp}. When used by itself in a rollout the cumulative reward achieved by the in-grasp manipulation controller before reaching joint limits is far lower than the cumulative reward achievable by a successful finger-gaiting policy within the same duration. 


\begin{algorithm}[t]
\caption{In-grasp manipulation (IGM) controller}
\begin{algorithmic}[1]
\State Get Jacobian, $\bm{J}_S$ and grasp map matrix $\bm{G}_S$
\State Compute optimal contact force $c^*_i$'s for stability from QP
\State Solve for $\Delta \bm{q}_{stable}$ using Eq~\lref{eq:dqstab}
\State Solve for $\Delta \bm{q}_{rot}$ from sampled $\Delta \bm{o}$ using Eq~\lref{eq:dqrot}
\State $\psi_{IGM}(\bm{s}) = \Delta \bm{q}_{stable} + \Delta \bm{q}_{rot}$ 
\end{algorithmic}
\llabel{algo:ingrasp}
\end{algorithm}

Our insight here is to simply use a fixed trajectory as a controller, 
\begin{equation}
    \psi_{CS}(\bm{s}) = \Delta \bm{q}_{switch}(h)  
\end{equation}
where $h=1, \ldots, H$ and $H$ is the length of the controller trajectory. 

This trajectory is a simple primitive for breaking and making one contact as described next. We design a trajectory that only breaks and makes contact with one selected finger at a given time even though contact switching as a skill can involve multiple contact switches.  We select a finger in contact at random and follow a hand-designed trajectory as shown in Fig~\lref{fig:experts} for the selected finger to break contact and remake contact with the object at a different location. This trajectory is achieved as shown in Fig~\lref{fig:switchtraj}. As one can expect, the switching controller by itself collects no meaningful reward.

Finally, we also construct a third controller. Combining the in-grasp manipulation and contact switching controller we construct a finger-gaiting controller. While a randomly selected fingertip breaks and makes contact, this controller also reorients the object with the fingertips in contact. Although this controller comes closer than the rest, it does not achieve the whole finger-gaiting skill we are looking to learn. Continuous object re-orientation is not possible with this controller as the fingers selected to switch contacts are still selected at random without any coordination.

We construct this controller again by superposition. We superpose in-grasp manipulation and contact switching controller.
\begin{align}
    \psi_{FG}(\bm{s}) &= \psi_{IGM}(\bm{s})  + \psi_{CS}(\bm{s}) 
\end{align}


This finger-gaiting controller also achieves a far lower return relative to a successful finger-gaiting policy.

\section{Learning from action priors from sub-skill controllers}


We use off-policy actor-critic reinforcement learning, where the policy used for episode rollouts (i.e the behavior policy) can be significantly different from the end-to-end policy being learned (i.e the learner policy). We use this dichotomy to enable exploration by constructing a behavior policy that is a composition of the learner policy and the sub-skill controllers. Importantly, this approach also sidesteps the need for the controller during deployment - once trained, the learner policy by itself is sufficient for deployment.

We construct the behavior policy to achieve sufficient exploration as follows. Our behavior policy periodically selects between the controllers and learner policy. We bias this selection towards higher value controllers using the critic's action-value estimate of the actions sampled from these controllers. This heuristic allows the probability of selecting a controller action to vary across the state-space. We show that this heuristic of using action-value estimates to select between available controllers and learner policy improves exploration. Optionally, we use value-weighted behavior cloning in addition to the nominal RL objective of the chosen off-policy RL method to update the policy. The method is detailed in the following paragaphs.



Let $\pi_\theta$ represent the learner policy and $\psi$ represent the controller. At every step of the rollout, we query actions from both the learner policy and controllers. Let these actions be $\bm{a}_\pi$, $\bm{a}_{\psi}$. The probability $p_{\psi}$ of selecting the action from a controller, as well as the probability $p_{\pi}$ of selecting the action prescribed by the learner policy, are computed as:


\begin{align}
p_{\psi} &= \frac{\exp(Q(\bm{s},\bm{a}_{\psi}))}{\exp(Q(\bm{s},\bm{a}_\pi)) + \exp(Q(\bm{s},\bm{a}_{\psi}))} \llabel{eq:softmaxpsi}\\
p_\pi &= 1 - p_{\psi} \llabel{eq:softmaxpi}
\end{align}

As $Q(\bm{a}_\pi, \bm{s})$ increases, $p_{\psi}$ decreases, i.e., as the learner policy improves, the probability of following the controller decays exponentially. In practice, we also impose a hard stop, i.e., we stop querying the controllers altogether after a few million steps of training and only use the learner policy for exploration. This point typically corresponds to the convergence of the critic. 

\begin{algorithm}[t]
\caption{Learning from sub-skill controllers}
\llabel{algo:qsampling}
\begin{algorithmic}[1]
\Require Controller $\psi$ \\
Initialize policy $\pi_{\theta}$, $Q_{\phi}$ and off-policy RL algorithm with replay buffer $\mathcal{D}$
\For{each iteration}
\For{$t= 0, \ldots, T-1$}
\For{every $H$ steps} \llabel{alg:line:behavstart}
\State Compute $p_{\pi}$, $p_{\psi}$ from $\bm{a}_\pi, \bm{a}_{\psi}$ (Eq~\lref{eq:softmaxpsi} \& \lref{eq:softmaxpi})
\State Select $\mu$: $\mu \sim \{ \pi, \psi \}$ given $p_{\pi}$, $p_{\psi}$ 
\EndFor
\State Follow selected $\mu$ for next H steps: $\bm{a}_t \sim \mu(\bm{s}_t)$ \llabel{alg:line:behavend} 
\State $\mathcal{D} \leftarrow \mathcal{D} \cup \{\bm{s}_t, \bm{a}_t, r(\bm{s}_t, \bm{a}_t), \bm{s}_{t+1}\}$
\EndFor
\For{each gradient step}
\State $\theta \leftarrow \theta - \nabla_{\theta} \mathcal{L}_{RL}(\theta)$ 
\State $\theta \leftarrow \theta  -  \nabla_{\theta} \mathcal{L}_{BC}(\theta)$ (Optional)
\llabel{alg:line:bcloss}
\EndFor
\EndFor
\end{algorithmic}
\end{algorithm}

Our method is summarized in Alg~\lref{algo:qsampling}. The first difference w.r.t the standard off-policy learning framework is steps \lref{alg:line:behavstart}-\lref{alg:line:behavend} when the behavior policy is different from the learning policy and is composed as described previously. The second difference, in step \lref{alg:line:bcloss}, entails updating the learner policy parameters using a value weighted behavior cloning loss $\mathcal{L}_{BC}$ (see Eq \lref{eq:bcloss}), in addition to the nominal actor loss $\mathcal{L}_{RL}$ as determined by the off-policy algorithm of choice. We follow a linear decay schedule for $\beta$.

\begin{equation}
    \mathcal{L}_{BC}(\theta) = \sum_{(\bm{s}, \bm{a}),~\bm{a} \neq \bm{a}_\pi }\beta \log \pi_{\theta}(\bm{a} | \bm{s}) Q(\bm{s}, \bm{a})
    \llabel{eq:bcloss}
    \vspace{-5mm}
\end{equation}


\section{Experiments and Results}

\llabel{sec:results}
Our main experimental goals are: (a) illustrating the importance of sampling actions from controllers for enabling exploration and (b) studying the effect of the different sub-optimal controllers proposed.  


\begin{figure*}
\centering
\includegraphics[width=1.0\textwidth]{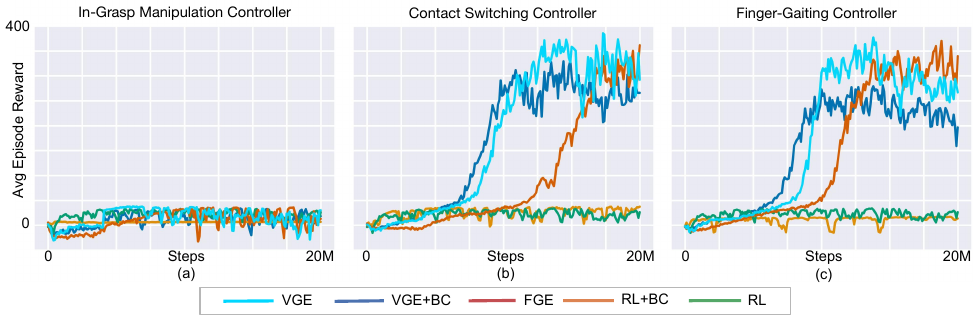}
\caption{Training curves for using (a) In-grasp manipulation controller (b) Contact switching controller (c) Finger-gaiting controller over all the evaluation conditions listed in Sec~\lref{sec:comp}. Our method that interleaves following controllers with policy (VGE, VGE + BC) learns finger-gaiting, while standard off-policy RL that does not follow controller actions fails. This shows it is essential to follow actions sampled by the sub-skill controllers to enable exploration.}
\llabel{fig:results}
\vspace{-5mm}
\end{figure*}

\begin{figure}
\centering
\begin{subfigure}[b]{0.2\textwidth}
\includegraphics[width=\textwidth]{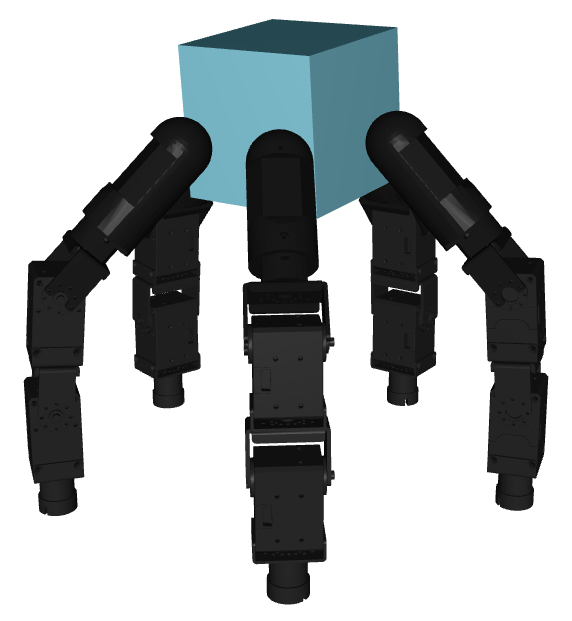}
\caption{15-dof hand}
\llabel{fig:hand}
\end{subfigure}
~~
\begin{subfigure}[b]{0.3\textwidth}
\centering
\includegraphics[width=\textwidth]{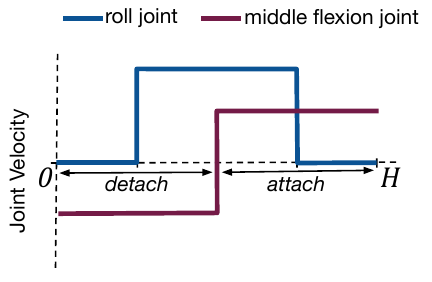}
\caption{Joint velocity for contact switching}
\llabel{fig:switchtraj}
\end{subfigure}
\caption{(a) The 5-fingered, 15-dof dexterous hand grasping the cube we use in our experiments (b) Joint velocity profile of the joints of the finger making and breaking contact. It involves detach and attach phases. Note that the distal flexion joint is held fixed.}
\end{figure}

\subsection{Experimental setup}
\llabel{sec:comp}

We use a dexterous hand in a simulated environment as shown in Fig~\lref{fig:hand}. It is a fully-actuated $15$-DOF hand consisting of five fingers where each finger consists of one roll joint followed by two flexion joints. Although our method is effective for a range of objects, in our analysis, we use a cube of side $10cm$ unless otherwise specified.  

The key hyperparameters used for the sub-optimal controller are as follows. For the in-grasp manipulation controller, we set the desired change in object pose $\Delta\bm{o}$ so that it represents a small reorientation about the z-axis. For the contact switching controller, which follows a fixed hand-designed trajectory, the velocity profile of the trajectory is as shown in Fig~\lref{fig:switchtraj}. It is designed to detach and re-attach the fingertip with the object. The roll-joint and middle-flexion-joint velocities are randomly sampled in the range of $[0.1, 0.4]$ rad/s.

Finally, as our off-policy RL algorithm, we used SAC \cite{Haarnoja2018-zj} and TD3 \cite{Fujimoto2018-lk} both with similar results. However, for brevity, we show results only for SAC. Note that with SAC, we set the exploration co-efficient $\alpha = 0$ -- using any non-zero value fails for all the baselines and also our method(s). Additionally, we do not query the controllers for actions after 4M steps of training in all of our evaluations.

\subsection{Evaluated conditions}
\llabel{sec:evalcond}

\begin{table}
    \centering
    \caption{Training robustness calculated as the percentage of seeds that successfully learn with about 10 seeds for each method.}
    \ra{1.1}
    \begin{tabular}{ccc}
        \midrule
         \phantom{} & CS Controller &  FG Controller\\
         \midrule
         VGE & $27\%$ &  $\mathbf{100}\%$ \\
         VGE + BC &  $36\%$ & $\mathbf{100}\%$ \\
         FGE & $67\%$  & $\mathbf{73}\%$\\
         \midrule
    \end{tabular}
    \llabel{tab:robust}
\end{table}

\llabel{sec:resultdisc}
\begin{figure}[t]
\centering
\includegraphics[width=0.35\textwidth]{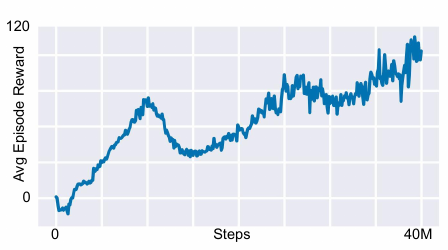}
\caption{Training curve for multiple objects using finger-gaiting controller.}
\llabel{fig:multiobj}
\vspace{-5mm}
\end{figure}

\begin{figure*}[t]
\centering
\includegraphics[width=0.8\textwidth]{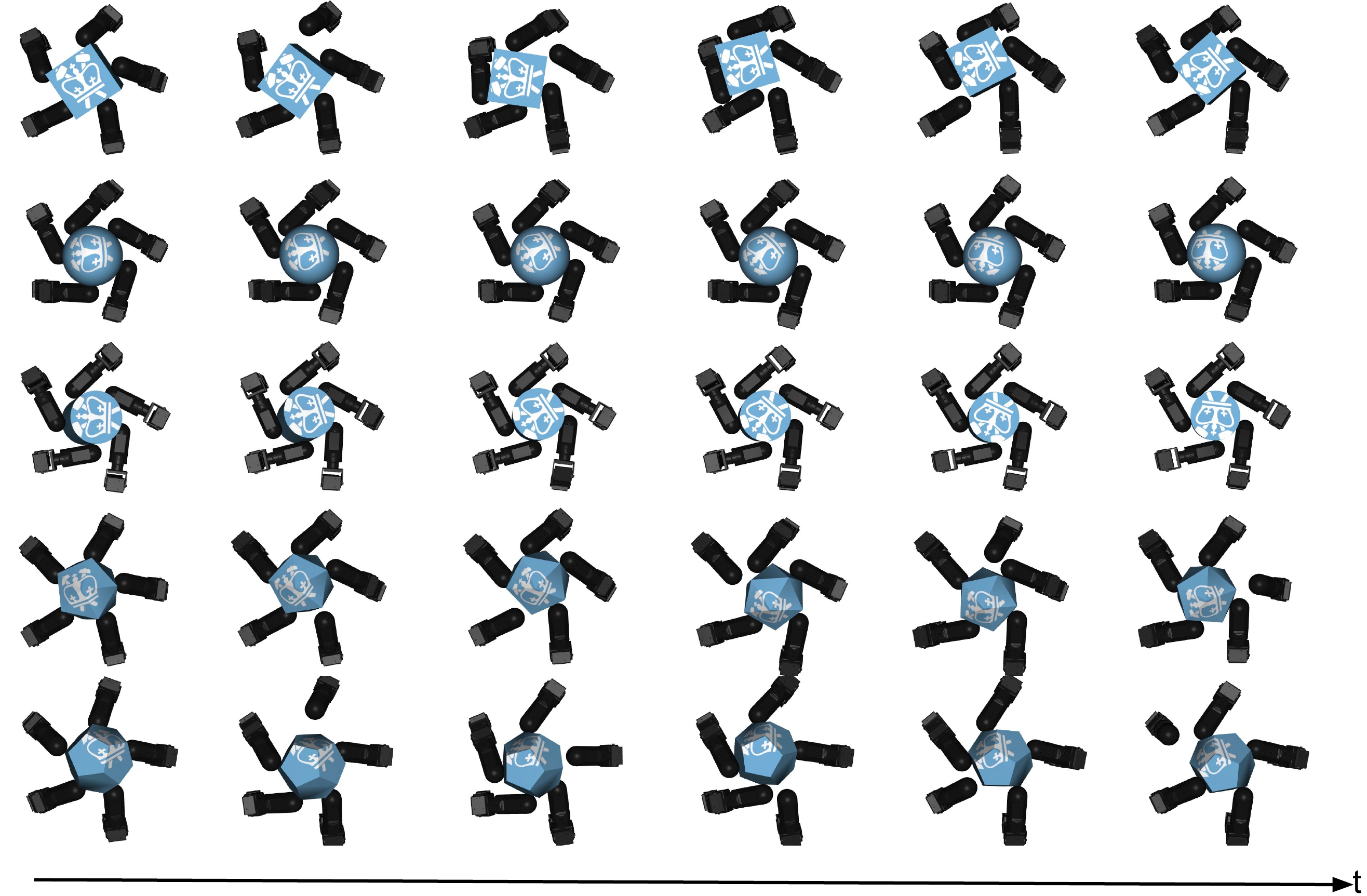}
\caption{Keyframes of the gaits achieved for multiple objects simultaneously with a single policy.}
\llabel{fig:gaits}
\end{figure*}

We evaluate and compare the following approaches:
\begin{itemize}
    \item \emph{RL}: Standard off-policy RL without the use of controllers and without any modification to the behavior policy or loss function is the first baseline. 
    \item \emph{RL + BC loss}: We also test a version of off-policy RL augmented using the same behavior cloning loss w.r.t the behavior policy as used in our method.
    \item \emph{Fixed Guided Exploration (FGE)}: We consider this baseline to study the advantage of allowing the probability of following the controllers to vary across the state-space. Unlike our method, controller probability is independent of the state as it is a fixed decay schedule based on training progress. 
    \item \emph{Value Guided Exploration (VGE)}: This is the method described in this paper, combining sub-skill controllers and value-guided controller sampling.
    \item \emph{VGE + BC loss}: This is a variant of our method that additionally uses behavior cloning with loss as in Eq. (\lref{eq:bcloss}).
\end{itemize}




\subsection{Results and discussion}

We evaluated the above experimental conditions individually for each of the sub-optimal controllers described in Sec \lref{sec:learnfg} (Fig ~\lref{sec:results}a-c). As we see, our method generally outperforms the baselines considered for each controller as proposed.  Here, the inability of baselines RL + BC loss and RL to learn is important to note. It underscores the key idea of our method that following actions sampled by the sub-skill controllers enables an otherwise very difficult exploration. 


Our evaluations also provide insights regarding the properties of the controllers that are critical for exploration. Fig \lref{fig:results} also demonstrates, that contact-switching behavior is critical for such exploration, as this controller enables learning finger-gaiting whenever it is used. 

While the in-grasp manipulation (IGM) controller is not critical for learning finger-gaiting, it is still beneficial as it improves training robustness. As seen from Table~\lref{tab:robust}, the success rate of learning to gait w.r.t varying seeds is markedly improved which may be attributable to the stability controller used as part of the in-grasp manipulation controller. 



Finally, we can also learn a single finger-gaiting policy for multiple objects with our method. In Fig~\lref{fig:multiobj}, we show that the finger-gaiting (FG) controller enables learning the finger-gaiting policy simultaneously for the sphere, cylinder, dodecahedron, icosahedron, and cube. The keyframes of the gaits achieved with this policy as visualized in Fig \lref{fig:gaits}.

\subsection{Sim-to-real transfer feasibility}
\llabel{sec:sim2real}
While we do not attempt sim-to-real transfer in this work, we believe that these policies can be transferred to the real hand using established methods from literature. As demonstrated by recent work \cite{OpenAI2018-bx, Chen2022-rc, Qi2022-wy, Khandate2023-gy}, domain randomization has been effective for transferring dexterous manipulation skills learned in simulation to the real hand. To investigate if our policies are suitable for such methods, we also tested their robustness to both perturbation forces on the object and noise in sensory feedback. As shown in Fig \lref{fig:perturbation}, our policies can sustain large perturbation forces equivalent to the full weight of the object, and are also robust to a high degree of sensor noise.  This characteristic leads us to believe that training with a curriculum of domain randomization will make transferring these robust policies to the real hand viable, and we hope to demonstrate this in future work.

\begin{figure}[h]
\centering
\includegraphics[width=0.55\textwidth]{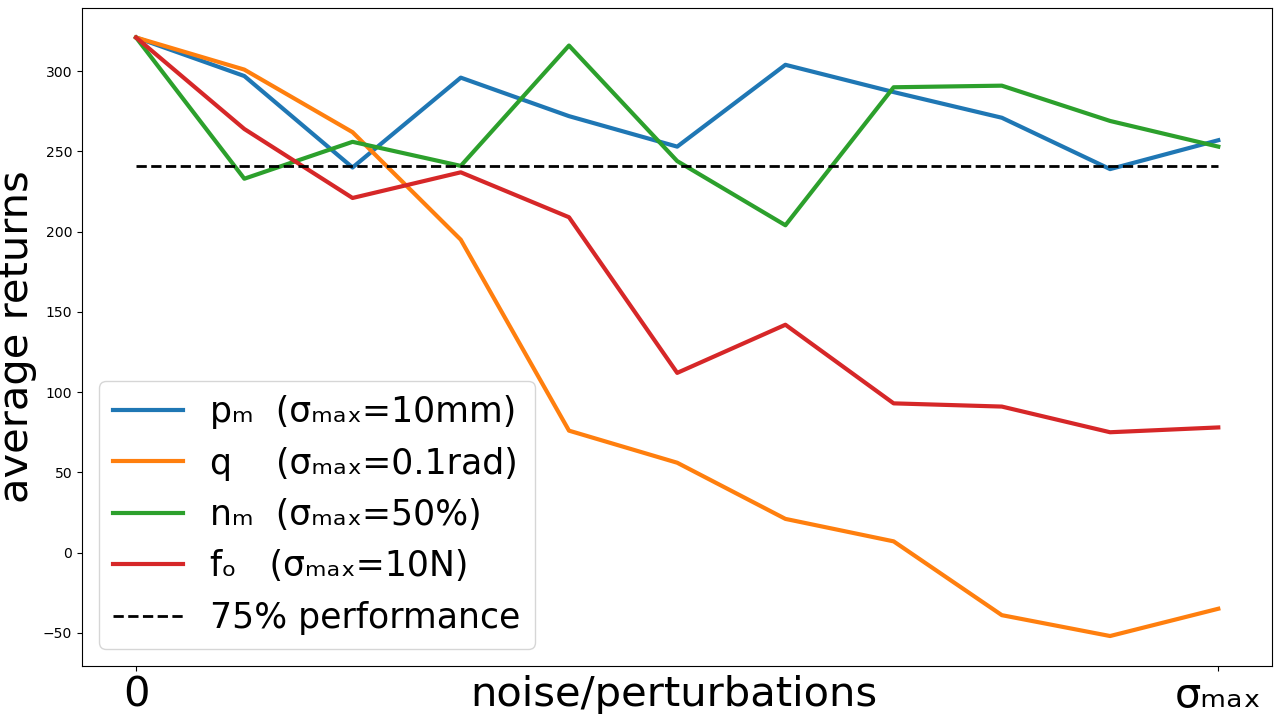}
\caption{Average episode returns of a policy trained on the cube with increasing perturbation forces and noise evaluated separately as shown. The policy sustains $0.1 rad$ of noise in joint positions $\boldsymbol{q}$, up to $50\%$  error in contact normal  $\boldsymbol{n}_m$ and a large error of $3mm$ in contact positions $\boldsymbol{p}_m$ -- all while suffering only  $25\%$  percent drop in average episode returns.}
\llabel{fig:perturbation}
\vspace{-5mm}
\end{figure}

\section{Conclusion}
Our work proposed the use of model-based controllers to assist exploration in reinforcement learning of dexterous manipulation tasks. To this end, we use off-policy RL and take advantage of the freedom with the construction of behavior policy to follow actions from the controller while collecting rollout trajectories to update the learning policy with such data to learn a successful policy.

We evaluated our method with the challenging dexterous manipulation task of learning finger-gaiting in-hand manipulation with only fingertip grasps. We designed simple controllers for key sub-skills in finger-gaiting - in-grasp manipulation and contact switching - and used these to enable learning effective finger-gaiting skills without any other additional means of improving exploration. Moreover, we show that our method benefits from model-based controllers even if the controllers are significantly sub-optimal.


More generally, we demonstrated a method to use domain expertise for learning dexterous manipulation tasks with improved sample efficiency. We believe our method of using sub-skill controllers for exploration is a promising approach for achieving dexterous skills with complex dynamics. While this work uses only model-based controllers, our method can also use learned sub-skill experts when available. It can also be extended to consider multiple experts. We hope to explore these promising directions in future work.
\llabel{ch:controller}



\chapter{Structured Exploration}

\llabel{ch:rrt}
The human-provided state priors are insufficient for exploration, especially for objects with complex geometries. The state space is composed of a very narrow labyrinthine structure.  

Reinforcement Learning (RL) of robot sensorimotor control policies has seen great advances in recent years, demonstrated for a wide range of motor tasks such as dexterous manipulation. This has translated into higher levels of dexterity than previously possible, typically demonstrated by the ability to reorient a grasped object in-hand using complex finger movements \cite{OpenAI2019-ng, Chen2021-ig, Qi2022-wy}. 


However, training a sensorimotor policy is still a difficult process, particularly for hard-exploration problems where the underlying state space exhibits complex structure, such as "narrow passages" between parts of the space that are accessible or useful. Manipulation is indeed such a problem: even when starting with the object secured between the digits, a random action can easily lead to a drop, and thus to an irrecoverable state. Finger-gaiting \cite{Ma2011-fo} further implies transitions between different subsets of fingers used to hold the object, all while maintaining stability. This leads to difficulty in exploration during training, since random perturbations in the policy action space are unlikely to discover narrow passages in state space. Current studies address this difficulty through a variety of means: using simple, convex objects and non-convex objects of significantly reduced size to reduce the difficulty of the task, reliance on support surfaces to reduce the chances of a drop, object pose tracking through extrinsic sensing, etc.

The difficulty of exploring problems with labyrinthine state space structures is far from new in robotics. In fact, the large and highly effective family of Sampling-Based Planning (SBP) algorithms was developed to address this exact problem. By expanding a known structure towards targets randomly sampled in the state space of the problem, SBP methods can explore even very high-dimensional state spaces in ways that are probabilistically complete, or guaranteed to converge to optimal trajectories. However, SBP algorithms are traditionally designed to find trajectories rather than policies. For problems with computationally demanding dynamics, SBP cannot be used online for previously unseen start states or to quickly correct when unexpected perturbations are encountered along the way.

Broadly, this work draws on the strength of both RL and SBP methods in order to train motor control policies for in-hand manipulation with finger gaiting. We aim to manipulate more difficult objects, including concave shapes, while securing them at all times without relying on support surfaces. Furthermore, we aim to achieve large reorientation of the grasped object with purely intrinsic (tactile and proprioceptive) sensing. To achieve that we use a non-holonomic RRT algorithm with added constraints to find approximate trajectories that explore the useful parts of the problem state space. Then, we use these trajectories towards training complete RL policies based on the full dynamics of the problem. In particular, we use state data sampled from the tree to construct exploratory reset distribution and use action data from tree paths to obtain a warm-start policy through imitation learning and illustrate this in Fig \lref{fig:rxr}.

\begin{figure*}[t]
    \centering
    \includegraphics[width=\textwidth]{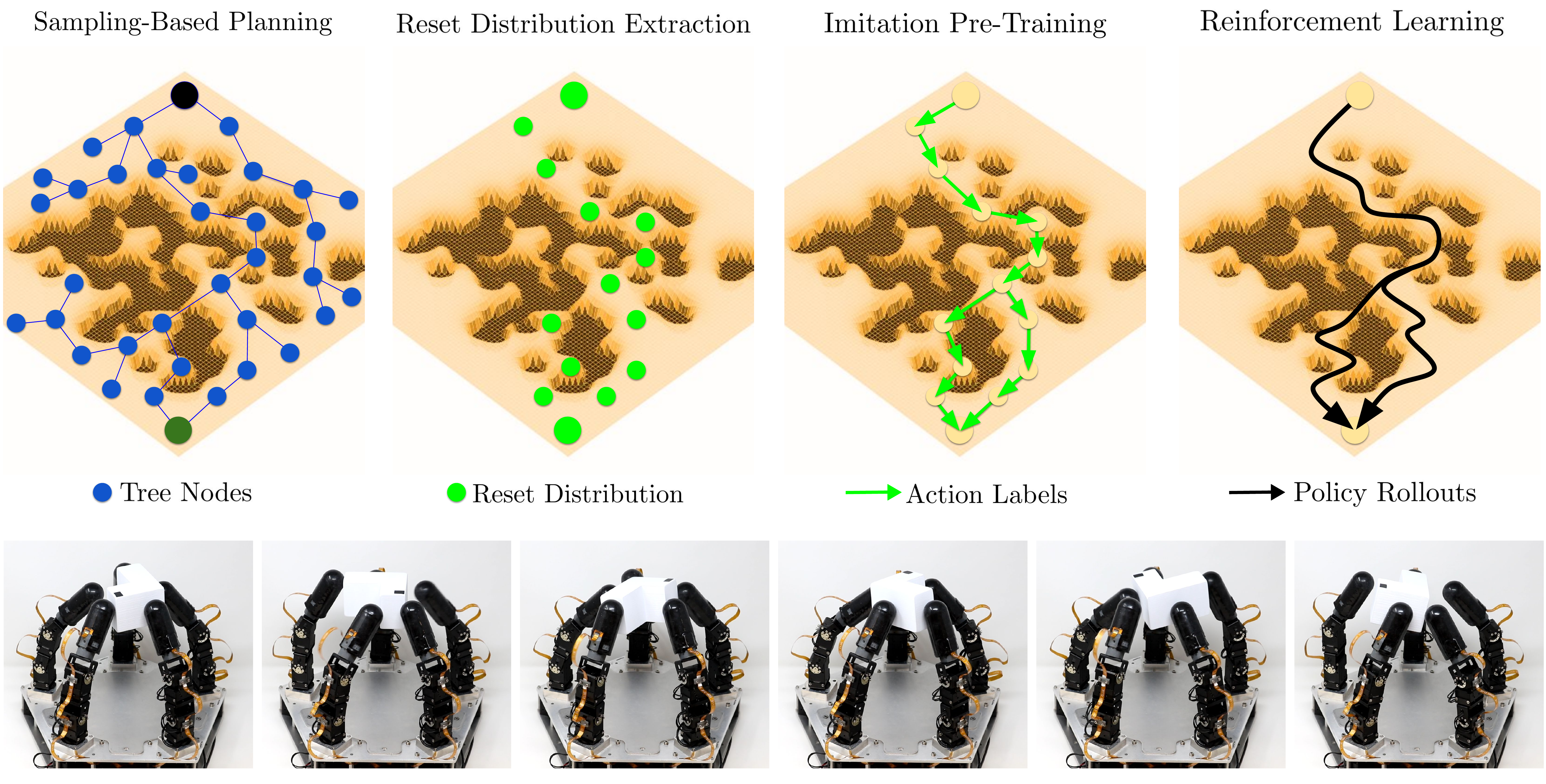}
    \caption{Our method illustrated with an abstract state-space consisting of narrow stable regions (beige) between large unstable regions (black). The proposed two-stage approach uses sampling-based planning to explore the challenging state-space and leverages the information within best paths by using state data as reset distribution and action data for imitation pre-training for efficient reinforcement learning.}
    \llabel{fig:rxr}
\end{figure*}

This extended version of our study contains a number of additions compared to the previous conference proceedings version~\cite{Khandate2023-fc}. We introduce Imitation Pre-training, a method for leveraging actions sampled from the RRT tree in order to further bootstrap learning and show that it further improves training efficiency. We extend our applications to the more difficult in-hand manipulation task of reorientation to a desired pose, which can be either a canonical pose known at training time (for example, as needed by downstream steps in an assembly process) or arbitrary and specified as a goal at run-time. Finally, we provide a number of additional insights obtained from an extensive ablation of our approach.

Overall, the main contributions of this work include:
\begin{itemize}
    \item To the best of our knowledge, we are the first to show that reset distributions generated via SBP with kinematic constraints can enable more efficient training of RL control policies for dexterous in-hand manipulation. 
    \item We show that SBP can explore useful parts of the manipulation state space, allowing RL to later fill in the gaps between approximate trajectories by learning appropriate actions under more realistic dynamic constraints. A warm-start policy obtained by imitating the actions from sampling-based trajectories provides an additional boost in training speed. 
    \item The exploration boost from SBP allows us to train policies for dexterous skills not previously shown, such as in-hand manipulation of concave shapes, with only intrinsic sensing and no support surfaces. We demonstrate these skills both in simulation and on real hardware.
\end{itemize}

\section{Related Work}
\vspace{3mm}
\noindent \textbf{Exploration in RL}.  Exploration methods for general RL operate under the strict assumption that the learning agent cannot teleport between states, mimicking the constraints of the real world. Under such constraints, proposed exploration methods include using intrinsic rewards \cite{Pathak2017-ik, Haarnoja2018-zj} or improving action consistency via temporally correlated noise in policy actions \cite{Amin2021-dm} or parameter space noise \cite{Plappert2017-gc}. 

Fortunately, in cases where the policies are primarily trained in simulation, this requirement can be relaxed, and we can use our knowledge of the relevant state space to design effective exploration strategies. A number of these methods improve exploration by injecting useful states into the reset distribution during training. \textcite{Nair2017-nu} use states from human demonstrations in a block stacking task, while Ecoffet et al. \cite{Ecoffet2019-lk, Ecoffet2021-xs} use states previously visited by the learning agent itself for problems such as Atari games and robot motion planning. \textcite{Tavakoli2018-ah} evaluate various schemes for maintaining and resetting from the buffer of visited states. However, these schemes were evaluated only on benchmark continuous control tasks \cite{Duan2016-em}. From a theoretical perspective, \textcite{Agarwal2020-pp} show that a favorable reset state distribution provides a means to circumvent worst-case exploration issues using sample complexity analysis of policy gradients. 

\vspace{3mm}
\noindent \textbf{Combining SBP and RL}. Finding feasible trajectories through a complex state space is a well-studied motion planning problem. Of particular interest to us are sampling-based methods such as Rapidly-exploring Random Trees (RRT) \cite{LaValle1998-kn, Karaman2010-zi, Webb2013-nt} and Probabilistic Road Maps (PRM) \cite{Kavraki1996-gr, Kavraki1998-wl}. These families of methods have proven highly effective and are still being expanded. Stable Sparse-RRT (SST) and its optimal variant SST* \cite{Li2021-lv} are examples of recent sampling-based methods for high-dimensional motion planning with physics. However, the goal of these methods is finding (kinodynamic) trajectories between known start and goal states rather than closed-loop control policies which can handle deviations from the expected states.

Several approaches have tried to combine the exploratory ability of SBP with RL, leveraging planning for global exploration while learning a local control policy via RL \cite{Chiang2019-vm,Francis2020-qe,Schramm2022-hs}. These methods were primarily developed for and tested on navigation tasks, where nearby state space samples are generally easy to connect by an RL agent acting as a local planner. The LeaPER algorithm \cite{Pinto2018-xi} also uses plans obtained by RRT as reset state distribution and learns policies for simple non-prehensile manipulation. However, the state space for the prehensile in-hand manipulation tasks we show here is highly constrained, with small useful regions and non-holonomic transitions. Other approaches use trajectories planned by SBP as expert demonstrations. \textcite{Morere2020-gq} recommend using a policy trained with SBP as expert demonstrations as an initial policy. Alternatively, \textcite{Jurgenson2019-ye} and \textcite{Ha2020-od} use planned trajectories in the replay buffer of an off-policy RL agent for multi-arm motion planning. First, these methods requires that planned trajectories also include the actions used to achieve transitions, which SBP does not always provide. Next, it is unclear how off-policy RL can be combined with the extensive physics parallelism that has been vital in the recent success of on-policy methods for learning manipulation \cite{Allshire2021-qp, Makoviychuk2021-ko, Chen2021-ig}. 

\vspace{3mm}
\noindent \textbf{Dexterous Manipulation}.
Turning specifically to the problem of dexterous manipulation, a number of methods have been used to advance the state of the art, including planning, learning, and leveraging the mechanical properties of the manipulator. \textcite{Leveroni1996-iy} build a map of valid grasps and use search methods to generate gaits for planar reorientation, while \textcite{Han1998-xj} consider finger-gaiting of a sphere and identify the non-holonomic nature of the problem. Some methods have also considered RRT for finger-gaiting in-hand manipulation \cite{Yashima2003-lw, Xu2007-yb}, but limited to simulation for a spherical object. More recently, Morgan et al. demonstrate robust finger-gaiting for object reorientation using actor-critic reinforcement learning~\cite{Morgan2021-ny} and multi-modal motion planning~\cite{Morgan2022-xt}, both in conjunction with a compliant, highly underactuated hand designed explicitly for this task. \textcite{Bhatt2022-fr} also demonstrate robust finger-gaiting and finger-pivoting manipulation with a soft compliant hand, but these skills were hand-designed and executed in an open-loop fashion rather than autonomously learned.

Model-free RL has also led to significant progress in dexterous manipulation, starting with {OpenAI's} demonstration of finger-gaiting and finger-pivoting~\cite{OpenAI2019-ng} trained in simulation and translated to real hardware. However, this approach uses extensive extrinsic sensing infeasible outside a lab setting, and relies on support surfaces such as the palm underneath the object. \textcite{Khandate2022-qt} show dexterous finger-gaiting and finger-pivoting skills using only precision fingertip grasps to enable both palm-up and palm-down operation, but only on a range of simple convex shapes and in a simulated environment. 

Recently \textcite{Makoviychuk2021-ko} showed that GPU physics could be used to accelerate learning skills similar to OpenAI's. \textcite{Allshire2021-qp}  used extensive domain randomization and sim-to-real transfer to re-orient a cube but used a table top as an external support surface. \textcite{Chen2021-ig, Chen2022-rc, Handa2022-fb} demonstrated in-hand re-orientation for a wide range of objects under palm-up and palm-down orientations of the hand with extrinsic sensing providing dense object feedback. 

\textcite{Sievers2022-lb, Pitz2023-kj, Rostel2023-jo} demonstrated in-hand cube reorientation to desired pose with purely tactile feedback. \textcite{Qi2022-wy, Yin2023-qw} used rapid motor adaptation to achieve effective sim-to-real transfer of tactile only in-hand manipulation skills for continuous reorientation of small cylindrical and cube-like objects. \textcite{Yuan2023-af, Qi2023-bj} use both visual and tactile sensing. We learn tactile in-hand manipulation skills critically, in our case, the exploration ability of SBP allows learning of policies for more difficult tasks, such as in-hand manipulation of non-convex and large shapes, with only intrinsic sensing. We also achieve successful and robust sim-to-real transfer without extensive domain randomization or domain adaptation by closing the sim-to-real gap via tactile feedback.

\section{Problem Description}
\llabel{sec:nonhol}

We focus on the problem of achieving dexterous in-hand manipulation while simultaneously securing the manipulated object in a precision grasp. Keeping the object stable in the grasp during manipulation is needed in cases where a support surface is not available or the skill must be performed under different directions for gravity (i.e. palm up or palm down). However, it also creates a difficult class of manipulation problems, combining movement of both the fingers and the object with a constant requirement of maintaining stability.

Formally, our goal is to obtain a policy for issuing finger motor commands to achieve a desired object transformation. The state of our system at time $t$ is denoted by $\bm{x}_t=(\bm{q}_t, \bm{p}_t)$, where $\bm{q} \in \mathcal{R}^d$ is a vector containing the positions of the hand's $d$ degrees of freedom (joints), and $\bm{p} \in \mathcal{R}^6$ contains the position and orientation of the object with respect to the hand. An action (or command) is denoted by the vector $\bm{a} \in \mathcal{R}^d$ comprising new setpoints for the position controllers running at every joint.

For parts of our approach, we assume that a model of the forward dynamics of our environment (\textit{i.e.} a physics simulator) is available for planning or training. We denote this model by $\bm{x}_{t+1} = F(\bm{x}_t, \bm{a}_t)$. We will show however that our results transfer to real robots using standard sim-to-real methods.  

As discussed above, we also require that the hand maintains a stable precision grasp of the manipulated object at all times. Overall, this means that our problem is characterized by a high-dimensional state space, but only small parts of this state space are accessible to us: those where the hand is holding the object in a stable precision grasp. Furthermore, the transition function of our problem is non-holonomic: the subset of fingers that are tasked with holding the object at a specific moment, as well as the object itself, must move in a concerted fashion. Conceptually, the hand-object system must evolve on the complex union of high-dimensional manifolds that form our accessible states. Still, the problem state space must be effectively explored if we are to achieve dexterous manipulation with large object reorientation and finger gaiting.

\section{Sampling-based State Space Exploration}
\llabel{sec:sbp}

To effectively explore our high-dimensional state space characterized by non-holonomic transitions, we turn to the well-known Rapidly-exploring Random Trees (RRT) algorithm. We leverage our knowledge of the manipulation domain to induce tree growth along the desired manifolds in state space. In particular, we expect two conditions to be met for any state: (1) the hand is in contact with the object only via fingertips, (2) the distribution of these contacts must be such that a stable grasp is possible.

\begin{algorithm}[t]
\caption{General-purpose non-holonomic RRT (G-RRT)}\llabel{alg:g-rrt}
\begin{algorithmic}[1]
\Require Tree containing root node, $G$; $N \gets 1$

\While{$N<N_{max}$} \llabel{line:mainloop}
\State $\bm{x}_{sample} \gets$ random point in state space
\State $\bm{x}_{node} \gets$ node closest to $\bm{x}_{sample}$ currently in G
\State $d_{min} \gets \infty; \bm{x}_{new} \gets $ NULL
\While{$k<K_{max}$} \llabel{line:loop}
\State $\bm{a} \gets \mathcal{N}(0, \alpha\bm{I})$ random action
\State $\bm{x}_a \gets F(\bm{x}_{node},\bm{a})$
\If{Stable($\bm{x}_a$) \textbf{and} $dist(\bm{x}_{sample},\bm{x}_a) < d_{min}$ }\llabel{line:stable}
\State $d_{min} \gets dist(\bm{x}_{sample},\bm{x}_a)$
\State $\bm{x}_{new} \gets \bm{x}_a$
\EndIf
\State $k \gets k+1$
\EndWhile
\If{$\bm{x}_{new}$ is not NULL}
\State Add $\bm{x}_{new}$ to G with $\bm{x}_{node}$ as parent
\State $N\gets N+1$
\EndIf
\EndWhile
\Statex
\Return G
\end{algorithmic}
\end{algorithm}

Assuming system dynamics $F()$ are available and fast to evaluate, we use here the general non-holonomic version of the RRT algorithm, which is able to determine an action that moves the agent towards a desired sample in state space via random sampling. We use the same version of this algorithm as described, for example, by ~\textcite{King2016-iv}, which we recapitulate here in Alg.~\lref{alg:g-rrt} and refer to as G-RRT.

The essence of this algorithm is the \textbf{while} loop in line~\lref{line:loop}: it is able to grow the tree in a desired direction by sampling a number $K_{max}$ of random actions, then using the transition function $F()$ of our problem to evaluate which of these produces a new node that is as close as possible to a sampled target. 

Our only addition to the general-purpose algorithm is the stability check in line \lref{line:stable}: a new node gets added to the tree only if it passes a stability check. This check consists of advancing the simulation for an additional 2s with no change in the action; if, at the end of this interval, the object has not been dropped (i.e. the height of the object is above a threshold) then the new node is deemed stable and added to the tree. Assuming a typical simulation step of 2 ms, this implies 1000 additional calls to $F()$ for each sample.

Overall, the great advantage of this algorithm lies in its simplicity and generality. The only manipulation-specific component is the aforementioned stability check. However, its performance can be dependent on $K_{max}$ (i.e. number of action samples at each iteration), and each of these samples requires a call to the transition function. This problem can be alleviated by the advent of highly efficient and massively parallel physics engines implementing the transition function, which is an important research direction complementary to our study.

We note that in previous work \cite{Khandate2023-fc} we used an additional condition on the state: the hand must maintain at least three fingers in contact with the object\footnote{Three contacts are the fewest that can achieve stable grasps without relying on torsional friction, which is highly sensitive to the material properties of the objects in contact}. Empirically, we noticed that the three contact constraint had minimal impact on the results of G-RRT. Thus, we discontinued the contact constraint in this work. 

In the same prior work, we also introduced a manipulation-specific version of the non-holonomic RRT algorithm dubbed M-RRT, which does not require the use of the transition function for stability checks. Instead, M-RRT uses manipulation kinematics alone to explore manifolds defined by the contact constraints. This version also required that the hand must maintain at least three fingers in contact with the object. While the M-RRT version has the potential to obtain significant speedups by foregoing dynamic simulation, G-RRT is appealing in its generality and simplicity. As both versions are able to effectively explore the state space, we focus this study on the G-RRT version of our algorithm and plan additional comparisons between these two versions for future work.

\section{Reinforcement Learning}
\llabel{sec:rl}

While the exploration method we have discussed so far is capable of exploring the complex state space of in-hand manipulation and identifying approximate transitions that follow the complex manifold structure of this space, it does not provide directly usable policies.  The state transitions themselves provide no mechanism to act in states that are not part of the tree, or to act under slightly different transition functions.

To generate closed-loop policies able to handle variability in the encountered states, we turn to RL algorithms. Critically, we rely on the trees generated by our sampling-based algorithms to ensure effective exploration of the state space during policy training.

\begin{algorithm}[t]
\caption{R$\times$R (+ IPT)}\llabel{alg:rxr+ipt}
\begin{algorithmic}[1]
\Require Randomly initialized actor $\pi_\theta$ and critic $V_\psi$, learning rates $\eta_1, \nu_1, \eta_2, \nu_2$
\Statex \texttt{\# Execute Sampling-based Planning}
\State Tree, $G$ $\gets$ G-RRT() \Comment{Alg \lref{alg:g-rrt}}
\Statex \texttt{\# Extract top trajectories}
\State Let $\bm{x}_{root}$ be the root state and  $\bm{x}_{goal}$ be the desired goal state towards achieving the manipulation task of interest.
\llabel{line:rxr_extract_start}
\State $D \gets \{\}, P \gets 1$
\While{$P < P_{max}$}
\State Sample goal $\bm{x}_{goal}$ \Comment{fixed or randomized}
\State $\bm{x}_{node} \gets$ node closest to $\bm{x}_{goal}$ now in G
\State $\tau = \{\bm{x}_{node}\}$ 
\While {$\bm{x}_{node} \neq \bm{x}_{root}$}
\State \textbf{delete}  $\bm{x}_{node}$ from $G$
\State Let $\bm{x}_{parent}$ be parent of $\bm{x}_{node}$
\State $\tau \gets \bm{x}_{parent} \cup \tau$  \Comment{nodes in reverse}
\State $\bm{x}_{node} \gets \bm{x}_{parent}$
\EndWhile
\State $D \gets D \cup \tau$,  $P \gets P + 1$
\EndWhile
\llabel{line:rxr_extract_end}

\Statex \texttt{\# Imitation Pre-training} \Comment{Optional}
\Statex Get warm-start policy and critic 
\State $\pi_{\theta}$, $V_\psi$ $\gets$ IPT($\mathcal{D}$, $\pi_\theta$, $V_\psi$, $\eta_1$, $\nu_1$) \Comment{ from Alg \lref{alg:impt}}
\llabel{line:rxr_ipt}

\Statex \texttt{\# Reinforcement Learning}
\State $\pi_{\theta}$, $V_\psi$ $\gets$ RL($\mathcal{D}$, $\pi_\theta$, $V_\psi$, $\eta_2$, $\nu_2$)
\For{$N$ iterations}
\llabel{line:rxx_rl_start}
\State Initialize/clear rollout buffer $\mathcal{R}$
\For{$E$ episodes}
\State Collect rollout $\tau$ with $\pi_\theta$ from initial state $\bm{x}_{0}$ where $\bm{x}_{0} \sim \mathcal{D}$ the buffer of best states from sampling-based planning.
\State $\mathcal{R} \leftarrow \mathcal{R} \cup \tau$
\EndFor
\Statex Update $\pi_{\theta}$ w.r.t policy loss $\mathcal{L}_{\pi}$ on $\mathcal{R}$ with RL algorithm of choice
\State $\theta \leftarrow \theta - \eta_2\nabla_{\theta} \mathcal{L}_{\pi}(\theta)$ 
\Statex Update critic $V_{\psi}$
\State $\mathcal{L}_{V} = \displaystyle\mathop{\mathbb{E}}_{\bm{o}_t \sim R}\left[(V_\psi(\bm{o}_t)-V^{targ}_t)^2\right]$
\State $\psi \leftarrow \psi - \nu_2\nabla_{\psi}L_{V}$
\EndFor
\llabel{line:rxx_rl_end}
\end{algorithmic}
\end{algorithm}

\subsection{Sampling-based Reset Distribution}
\llabel{sec:RL}

One mechanism we use to transition information from the sampling-based tree to the policy training method is via the reset distribution: we select relevant paths from the planned tree and then use the nodes therein as reset states for policy training.

We note that the sampling-based trees as described here are task-agnostic. Their effectiveness lies in achieving good coverage of the state space (usually within pre-specified limits along each dimension). Once a specific task is prescribed (e.g. via a reward function), we must select states from paths through the tree that are relevant to the task. After selecting said states, we use a uniform distribution over these states as a reset distribution for RL. We describe this concretely in Alg \lref{alg:rxr+ipt}. In particular, lines \lref{line:rxr_extract_start} - \lref{line:rxr_extract_end} show the heuristic approach we use to select paths through the tree to compose the set of reset states. Lines  \lref{line:rxx_rl_start} - \lref{line:rxx_rl_end} show reinforcement learning from using these reset states for exploration. We note that other selection mechanisms are also possible; a promising and more general direction for future studies is to select tree branches that accumulate the highest reward.

This approach has a theoretical grounding in related work showing that derived reset distributions can be used to aid exploration. Let $\rho$ denote the initial state distribution of the MDP to be solved with policy-gradient RL. Recent results~\cite{Agarwal2020-pp} show that it is beneficial to compute policy gradients for policy $\pi$ under a different initial distribution $\mu$ if it enables sufficient exploration. If $d^{\pi}_{\rho}$ denotes the stationary state distribution induced by policy $\pi$ under the initial state distribution $\rho$, then $d^{\pi^*}_{\rho}$ is the stationary distribution of the optimal policy $\pi^*$ with initial state distribution $\rho$. Improved exploration can be achieved if $\mu$ sufficiently covers $d^{\pi^*}_{\rho}$. Here, we obtain such a favorable distribution $\mu$ from plans derived via sampling-based planning.

As this approach is compatible with online RL wherein policy rollouts are collected from a new set of states every episode, both off-policy and on-policy RL are equally feasible. However, we use on-policy learning due to its compatibility with GPU physics simulators and relative training stability.

\subsection{Imitation Pre-training}

\begin{algorithm}[t]
\caption{Imitation Pre-training (IPT)}\llabel{alg:impt}
\begin{algorithmic}[1]
\Require Sampling-based trajectories (state-only) as demonstration $\mathcal{D}$, randomly initialized policy $\pi$ and value network $V$ with parameters $\theta, \psi$, learning rates $\eta$ and $\nu$
\Statex \texttt{\# Assemble buffer with actions $\mathcal{D}'$}
\For {all consecutive state pairs $\bm{x}_k, \bm{x}_{k+1}$}
\State Compute $\bm{o}_k$ observation vector at $\bm{x}_k$
\State Get $\bm{a}_k$ with Eq \leqref{eq:inv}
\State $\mathcal{D}' \leftarrow \mathcal{D} \cup (\bm{o}_k, \bm{a}_k) $
\EndFor
\Statex \texttt{\# BC with MSE loss $\mathcal{D}'$}
\For{$E_1$ epochs}
\State $L_\pi =\displaystyle \mathop{\mathbb{E}}_{\bm{o},\bm{a} \sim D'} \left[(\bm{a} - \pi_{\theta}(\bm{o}))^2\right]$
\State $\theta \leftarrow \theta - \eta \nabla_{\theta}L_{\pi}(\theta)$
\EndFor
\Statex \texttt{\# Value pre-training}
\For{$E_2$ epochs}
\State Collect rollouts with $\pi_\theta$ and store in $\mathcal{R}$ 
\State $L_{V} = \displaystyle\mathop{\mathbb{E}}_{\bm{o}_t \sim R}\left[(V_\psi(\bm{o}_t)-V^{targ}_t)^2\right]$
\State $\psi \leftarrow \psi - \nu\nabla_{\psi}L_{V}$
\EndFor
\Statex
\Return $\pi_\theta$, $V_\psi$ 
\end{algorithmic}
\end{algorithm}

Relying on sampling-based trajectories exclusively for the reset state distribution disregards potentially valuable information concerning actions and the connectivity between states within exploration trees. Imitation learning, which maps states to actions via supervised learning, can be effective in using such action information embedded within the exploration trees. However, this is not straightforward to execute. The sampling-based trajectories cannot be treated as authentic demonstrations. Instead, they are quasi-static demonstrations composed of a sequence of stable states as a consequence of the stability constraint (Alg \lref{alg:g-rrt}, line \lref{line:stable}) we enforce during sampling-based planning. Due to this stability check, the actions are a distorted version of the desired actions under full dynamics. Thus, imitation learning, for instance, with Behavioral Cloning (BC), is unlikely to yield a successful policy due to the approximate nature of demonstrations and inherent challenges arising from distribution shift. Nonetheless, the resulting policies can be used as a warm-start policy for reinforcement learning. 

Here we provide a description of this approach. First, we observe that the difference in joint angles between node transitions is a reliable approximation of the desired action, particularly its direction. Augmented by a scaling hyper-parameter, these differences can serve as action labels. For illustration, let $\bm{q}_k$ and $\bm{q}_{k+1}$ be joint angles at successive states. The action labels $\bm{a}_k$ can be obtained by simply 
scaling the difference with a scaling factor $\beta$, i.e $\bm{a}_k = \beta (\bm{q}_{k+1} - \bm{q}_{k})$. In cases where obtaining such action labels is more complex, an inverse dynamics model can be of assistance. Eq \leqref{eq:inv} restates this generally
\begin{align}
\llabel{eq:inv}
    \bm{a}_k &= g(\bm{x}_{k}, \bm{x}_{k+1})
\end{align}
where $g$ is the inverse function to compute action labels between two successive states of a demonstration.

We then utilize these demonstrations via learning an imitation policy to bootstrap reinforcement learning. To retain the benefits from any generalization properties achieved via imitation learning, we pre-train the critic network on rollouts of the imitation policy to avoid washing it out with a randomly initialized critic \cite{Hansen2022-pc}. This method aligns with contemporary approaches for integrating reinforcement learning with imitation learning which go beyond simply augmenting the online replay buffer with demonstrations \cite{Hu2023-jt}. Our main approach, supplemented with imitation pre-training, is depicted in Alg. \lref{alg:impt}. Lines \lref{line:rxr_ipt} and \lref{line:rxx_rl_start} show imitation pre-training with sampling-based demonstrations and warm-starting reinforcement learning.

\section{Experimental Setup and Tasks}

\vspace{3mm}
\noindent \textbf{Hardware.} We use the robot hand shown in Fig.~\lref{fig:rxr}, consisting of five identical fingers. Each finger comprises a roll joint and two flexion joints for a total of 15 fully actuated position-controlled joints. For the real hardware setup, each joint is powered by a Dynamixel XM430-210T servo motor. The distal link of each finger consists of an optics-based tactile fingertip as introduced by \textcite{Piacenza2020-tk}. 

\begin{figure}
    \centering
    \includegraphics[trim=0mm 20mm 0mm 0mm,clip,width=0.3\textwidth]{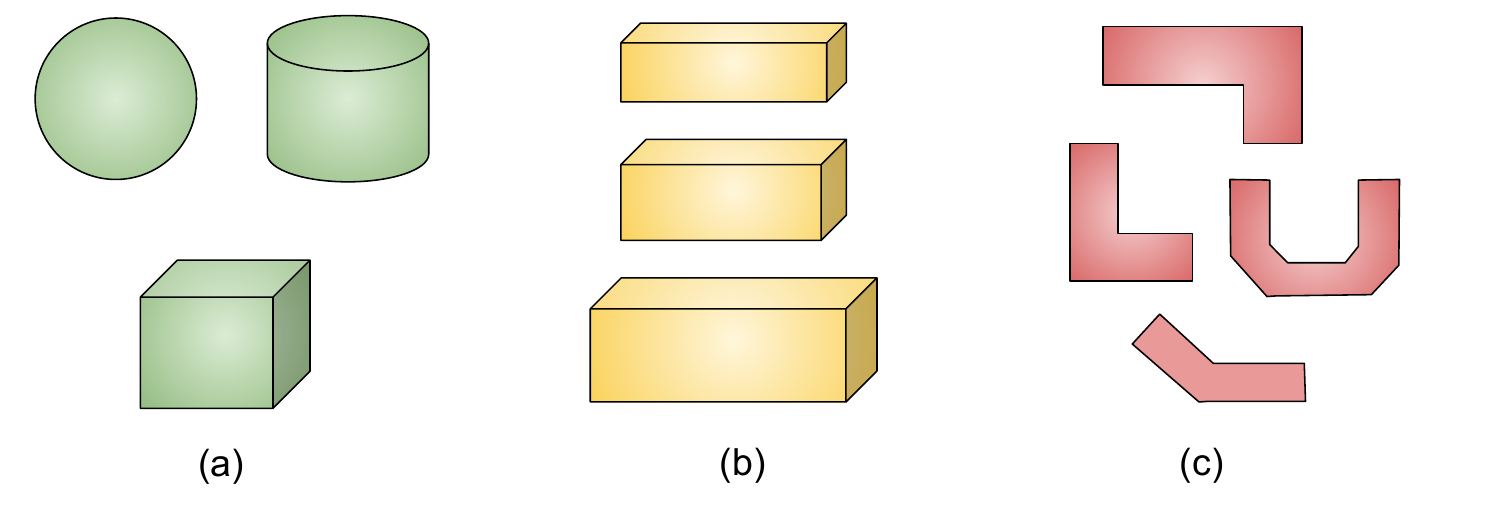}
    \caption{The object shapes for which we learn finger-gaiting. From left to right: the easy, medium and hard categories.}
    \llabel{fig:objset}
\end{figure}

We test our methods on the object shapes illustrated in Fig.~\lref{fig:objset}. We split this into categories: "easy" objects (sphere, cube, cylinder), "moderate" objects (cuboids with elongated aspect ratios), and "hard" objects (either concave L- or U-shapes) and train object-specific policies  for each of these objects with the proposed methods and also the baseline methods to be discussed later.
We note that in-hand manipulation of the objects in the "hard" category has not been previously demonstrated in the literature.

\vspace{3mm}
\noindent\textbf{Exploration Trees.} The extensive sampling of possible actions, which is the main computational expense of G-RRT (line \lref{line:loop}) lends itself well to parallelization. In practice, we use the IsaacGym \cite{Makoviychuk2021-ko} simulator to parallelize this algorithm at both these levels (16 parallel evaluations of the main loop, and 1024 parallel evaluations of the action sampling loop). 
In general, given the advent of increasingly more powerful parallel architectures for physics simulation, we expect that more general methods that are easier to parallelize might win out in the long term over more problem-specific solutions that are more sample efficient at the individual thread level.

\vspace{3mm} 
\noindent\textbf{Reinforcement Learning}. We train our policies using Asymmetric Actor Critic PPO \cite{Pinto2017-ji, Schulman2017-td} where we separate actor and critic training for improved performance. All training is done in the IsaacGym simulator and the critic uses object pose $\boldsymbol{p}$, object velocity $\dot{\boldsymbol{p}}$, and net contact force on each fingertip $\boldsymbol{t}_1 \ldots \boldsymbol{t}_{m}$ as feedback in addition to the feedback already provided to the policy network.

For Imitation Pre-training, we compute action labels from state transitions using $\beta=2$ in Eq \leqref{eq:inv}.
We train the critic with rollouts of the imitation policy for 2M steps to mitigate forgetting issues inherent in a randomly initialized critic network. 

\subsection{Tasks}
We evaluate our method on a variety of challenging dexterous in-hand manipulation tasks.
\\~\\
\noindent\textbf{Finger-gaiting}.
First, we focus on the task of achieving large in-hand object rotation about a desired axis. We, as others before~\cite{Qi2022-wy}, believe this to be representative of this general class of problems, since it requires extensive finger gaiting and object reorientation.

We chose to focus on the case where the only sensing available is hand-centric, either tactile or proprioceptive. Achieving dexterity with only proprioceptive sensing, as biological organisms are clearly capable of, can lead to skills that are robust to occlusion and lighting and can operate in very constrained settings. With this directional goal in mind, the observation available to our policy consists of tactile and proprioceptive data collected by the hand, and no global object pose information. Formally,  the observation vector is
\begin{equation}
\bm{o}_t = [\bm{q}_t, \bm{q}^s_t, \bm{c}_t]    
\end{equation}
where $\bm{q}_t, \bm{q}^s_t \in \mathcal{R}^d$ are the current positions and setpoints of the joints, and $\bm{c}_t \in [0, 1]^m$ is the vector representing binary (contact / no-contact) touch feedback for each of $m$ fingers of the hand.

In this task, where our goal is finger-gaiting for z-axis object rotation, we plan trees where object rotation around the x- and y-axes was restricted to 0.2 radians. Then we select $2 \times 10^4$ nodes from the paths that exhibit the most rotation around the z-axis to construct our reset distribution and for imitation pre-training. On average, each such path comprises 20-30 nodes. We recall that all tree nodes are subjected to an explicit stability check under full system dynamics before being added to the tree; we can thus use each of them as is. 

Similar to our previous work \cite{Khandate2022-qt}, we use a reward function that rewards object angular velocity about the z-axis. In addition, we include penalties for the object's translational velocity and its deviation from the initial position \cite{Qi2022-wy}.
\\~\\
\noindent\textbf{Arbitrary Reorientation}. While finger-gaiting to achieve maximum rotation is a good proof-of-concept task with several applications, we now consider the canonically studied and versatile in-hand manipulation skill of reorientation to a desired pose. We continue with the requirement to achieve in-hand manipulation with only stable fingertip grasps, as it enables reorientation in arbitrary orientations of the hand. 

We take on the challenging task of orienting an object from a randomly assigned initial pose to a desired arbitrary orientation, a frequently studied task in dexterous in-hand manipulation \cite{Andrychowicz2020-le, Chen2021-ig}.  In addition to proprioceptive and tactile data collected by the hand, we include current and desired object orientations as inputs to the policy resulting in the following observation vector,
\begin{equation}
\bm{o}_t = [\bm{q}_t, \bm{q}^s_t, \bm{c}_t, \bm{\phi}_t, \bm{\phi}_g]    
\end{equation}
where $\bm{\phi}_t, \bm{\phi}_g \in \mathcal{R}^4$ are the current and desired object orientation respectively. Our reward function is a modification of the one proposed in \cite{Chen2021-ig}, as we found that the original reward function fails in our setup. Eq. \lref{eq:reward_ar} describes the reward function that we use for this task, 

\begin{equation}
    \llabel{eq:reward_ar}
    r_t = c \cdot \max(\min(\Delta_t - \Delta_{t-1}, \epsilon), -\epsilon) + c_{\text{success}} \cdot \mathbbm{1}[\text{success}]
\end{equation}

where $\Delta_t$ is the angular distance between the current and desired object orientation, $\epsilon > 0$ is a threshold coefficient and $c < 0$ is a scaling coefficient. A large positive reward, denoted $c_{\text{success}}$, is added if the agent successfully completes the task. Task completion is achieved through successful reorientation plus satisfaction of heuristics adapted from previous work \cite{Chen2021-ig}. The criteria are described here:
\begin{enumerate}
    \item Reorientation criterion: $\Delta_t < \theta_{\text{thresh}}$
    \item Joint angular velocity criterion: $\|\bm{\dot{q}}_t\|_2 < \dot{q}_{\text{thresh}}$
    \item Object linear velocity criterion: $\|\bm{\dot{x}}_t\|_2 < \dot{x}_{\text{thresh}}$
    \item Object angular velocity criterion: $\|\bm{\omega}_t\|_2 < \omega_{\text{thresh}}$
\end{enumerate}

 Unlike the previous task where a subset of the tree needs to be extracted to aid exploration, here the full tree, i.e. all nodes of the tree, can be used towards assembling the buffer of exploratory reset states. To ensure complete exploration of the state space, we use G-RRT trees generated without any constraints on object orientation.
\\~\\
\noindent\textbf{Go-to-root}. While our method improves on sample efficiency for learning policies for arbitrary reorientation, obtaining such policies remains computationally expensive due to the requirement of a few tens of billion simulation training steps. Learning such policies can be overkill; often, in practice, it is sufficient to reorient the object to a fixed canonical pose starting from an arbitrary initial pose. In a packaging line, for example, items arrive at arbitrary orientations. The essential in-hand manipulation task is to reorient objects to a canonical pose before inserting them in the packaging container. 

However, learning to reorient the object to a fixed desired orientation starting from an arbitrary initial pose is still difficult as it suffers the same exploration challenges seen in previous tasks. Fortunately, our method of using exploration trees is naturally well suited for this problem. The robot state with the desired canonical orientation can itself be used as the root node while generating the exploration tree. Thus, we refer to this as the "Go-to-root" task. 

In this task, we aim to learn a policy to reorient the grasped object to reach a desired canonical / root orientation. The inputs to the policy network are similar to the arbitrary reorientation task, consisting of proprioceptive, tactile, and current object pose feedback but excluding the desired goal orientation as it is fixed, resulting in the observation vector in \leqref{eq:g2r_obs}. The reward function and success criteria are identical to the arbitrary reorientation task except with fixed canonical orientation $\bm{\phi}_g$ as the goal.
\begin{equation}
\llabel{eq:g2r_obs}
\bm{o}_t = [\bm{q}_t, \bm{q}^s_t, \bm{c}_t, \bm{\phi}_t]    
\end{equation}

For imitation pre-training, we extract plans by backtracking from randomly selected nodes with large displacement from the root. In each task, this amounts about about 3K unique observation-action pairs for finger-gaiting and about 15k unique observation-action pairs for go-to-root task.

\subsection{Algorithms and Baselines}
In our experiments, we compare the following approaches:
\\~\\
\noindent\textbf{Ours, R$\times$R}: In this variant, we use the method presented in this paper, relying on exploratory reset states obtained by growing the tree via G-RRT. In all cases, we use a tree comprising $10^5$ nodes as informed the ablation study in Fig~\lref{fig:treesize}.
\\~\\
\noindent\textbf{Ours, R$\times$R + IPT}: In this variant of our method, we use Imitation Pre-training with labels obtained from the grown G-RRT tree, alongside using exploratory reset distribution extracted from the same tree. 
\\~\\
\noindent\textbf{Stable Grasp Sampler (SGS)}: This baseline represents an alternative to the method presented in this paper: we use a reset distribution consisting of stable grasps generated by sampling random joint angles and object orientations. This approach has been effective in demonstrating precision in-hand manipulation with only intrinsic sensing \cite{Khandate2022-qt, Qi2022-wy} for simple shapes. 
\\~\\
\noindent\textbf{Explored Restarts (ER)}: This method selects states explored by the policy itself during random exploration to use as reset states~\cite{Tavakoli2018-ah}. It is highly general, with no manipulation-specific component, and requires no additional step on top of RL training. We implement the "uniform restart" scheme as it was shown to have superior performance on high dimensional continuous control tasks. However, we have found it to be insufficient for the complex state space of our problem: it fails to learn a viable policy even for simple objects.
\\~\\
\noindent\textbf{Fixed Initialization (FI)}: For completeness, we also tried resetting from a single fixed state.
\\~\\
\noindent\textbf{Fixed Initialization (FI + IPT)}: Towards understanding the effect of pre-training, we also tried FI baseline but with a warm start policy obtained from Imitation Pre-training using the G-RRT tree.
\\~\\
\noindent\textbf{Gravity Curriculum (GC)}: Additionally, we evaluated Fixed Initialization with gravity curriculum, linearly annealed from $0 \frac{\text{m}}{\text{s}^2}$ to $-9.81 \frac{\text{m}}{\text{s}^2}$ over the course of 50M timesteps.

\section{Results}

\begin{figure*}
    \centering
    \includegraphics[width=\textwidth]{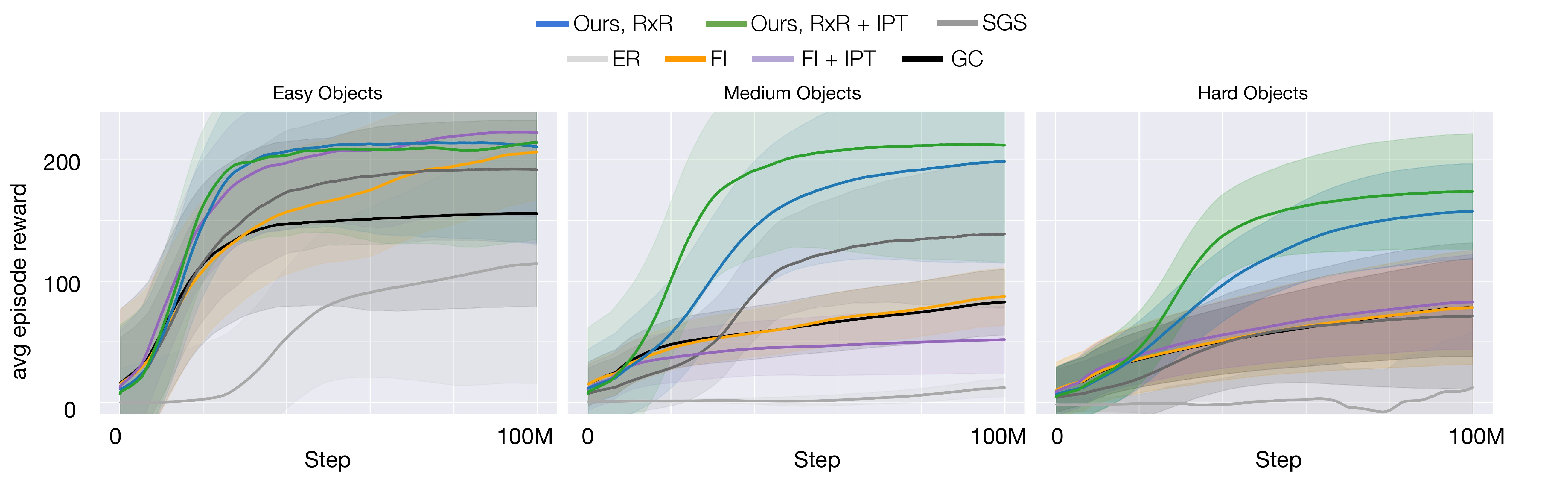}
    \caption{Training performance of our methods and a number of baselines for the Finger-gaiting task on the object categories shown in Fig.~\lref{fig:objset}.}
    \llabel{fig:results}
\end{figure*}
\begin{figure*}
\vspace{-10mm}
    \centering
    \includegraphics[width=\textwidth]{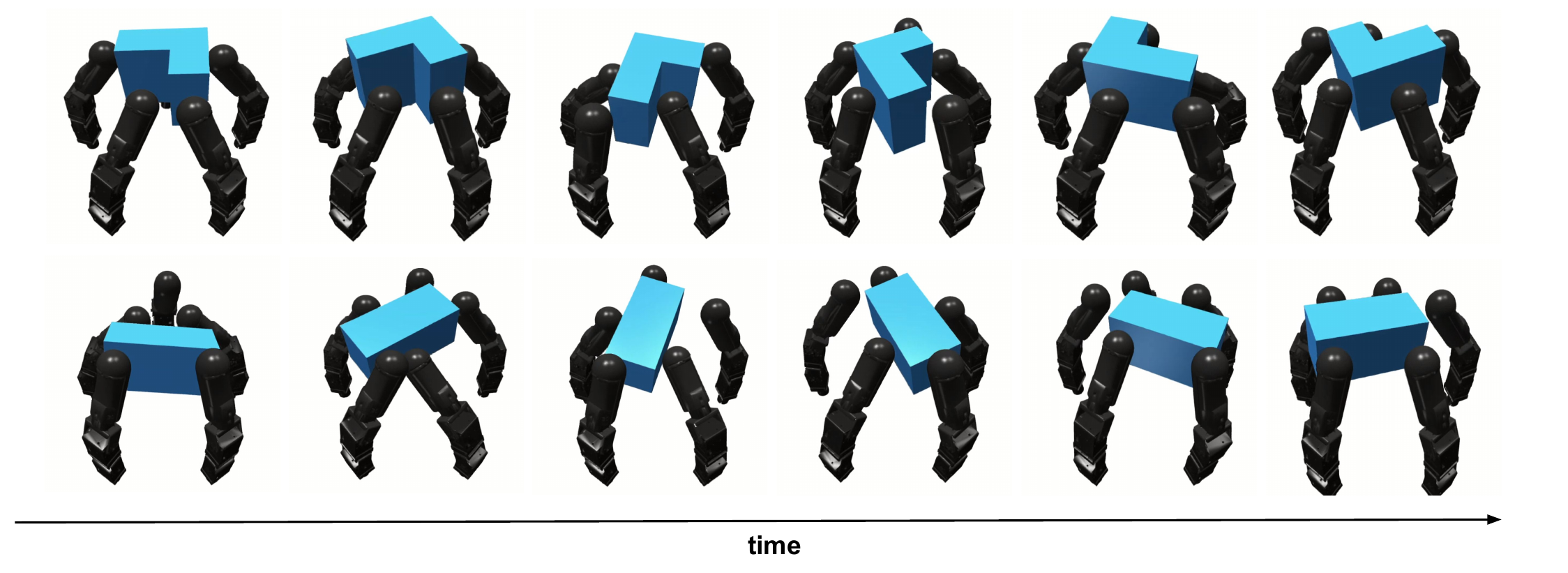}
    \caption{Key frames of the policies for the Finger-gaiting achieved with our method R$\times$R for representative objects in simulation.}
    \llabel{fig:gaiting}
\end{figure*}
\begin{figure}
    \centering
    \includegraphics[width=0.7\textwidth]{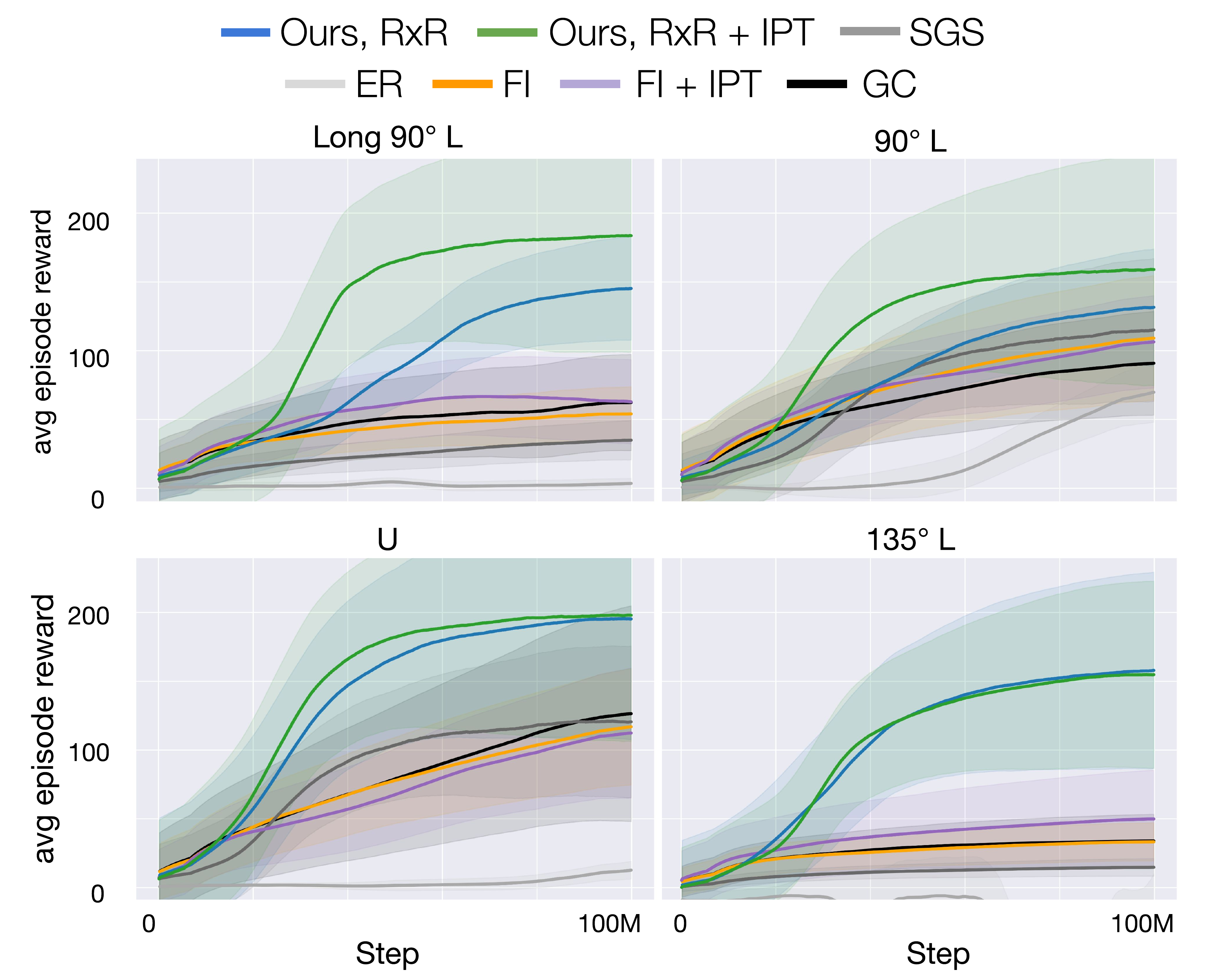}
    \caption{Training performance of our methods and a number of baselines for the Finger-gaiting task on hard objects shown in Fig.~\lref{fig:objset}.}
    \llabel{fig:results_hard}
\end{figure}

\subsection{Evaluation in simulation}

\llabel{sec:result_fg}
\noindent\textbf{Finger-gaiting}. Our training results aggregated over 3 seeds are summarized in Fig.~\lref{fig:results} \& Fig.~\lref{fig:results_hard}. We find that our methods (R$\times$R, R$\times$R + IPT) converge within 50M timesteps across all objects for the Finger-gaiting task, consistently outperforming baselines.

\begin{figure*}
    \centering
    \includegraphics[width=\textwidth]{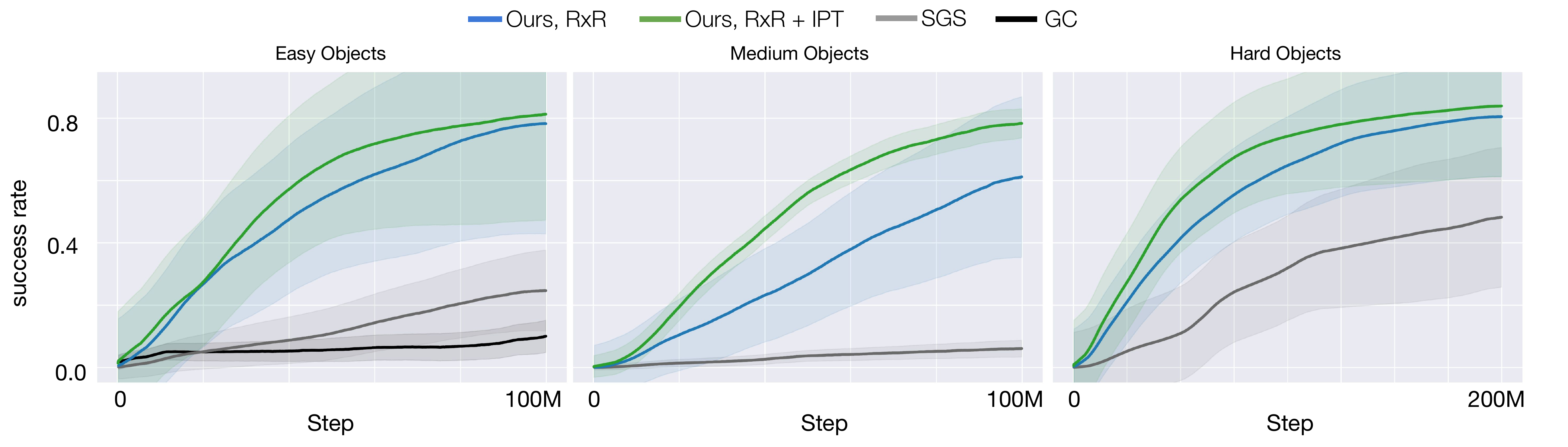}
    \caption{Training performance for the Go-to-root task with our methods and other best-performing baselines.}
    \llabel{fig:results_g2r}
\end{figure*}
\begin{figure*}
    \centering
    \includegraphics[width=\textwidth]{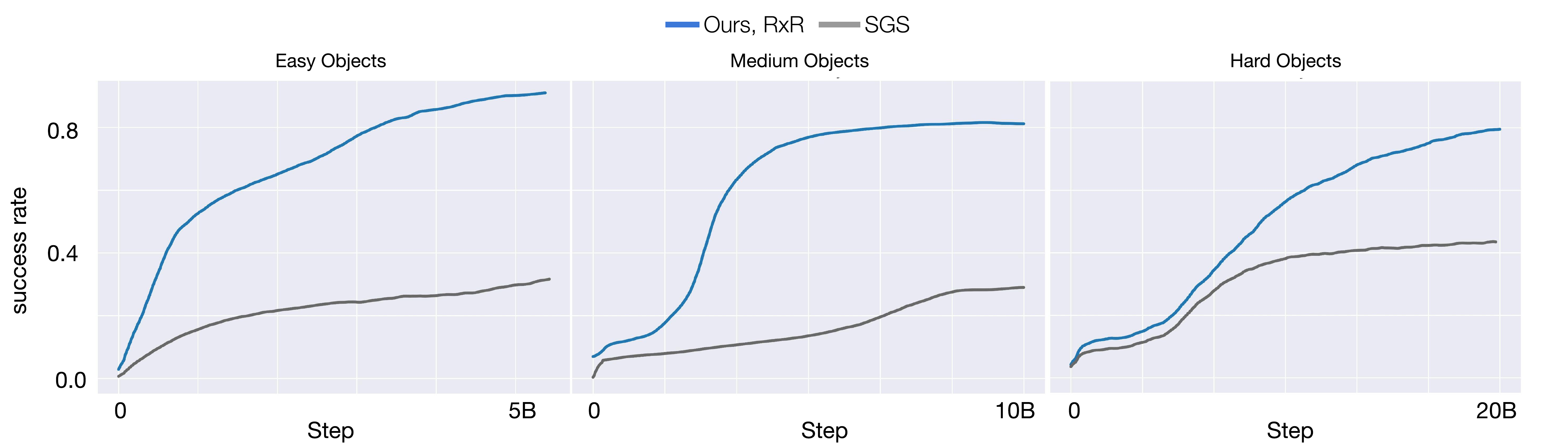}
    \caption{Training performance for the Arbitrary Reorientation task comparing our methods (R$\times$R, R$\times$R + IPT) with Stable Grasp Sampler (SGS) baseline.}
    \llabel{fig:results_ar}
\end{figure*}
\begin{figure}
    \centering
    \includegraphics[width=0.5\textwidth]{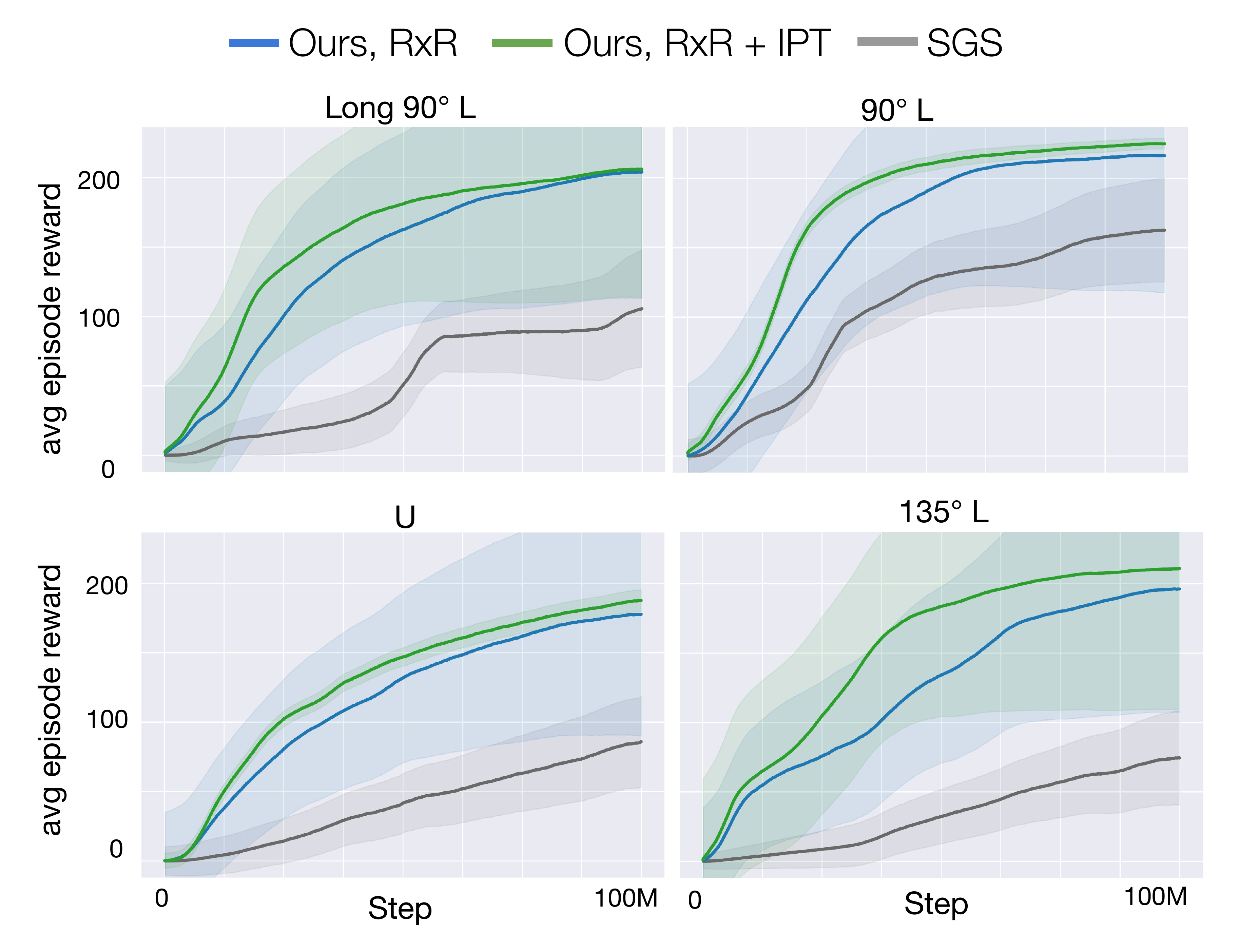}
    \caption{Training performance for the Go-to-root task with our methods and other best-performing baselines for objects in hard category.}
    \llabel{fig:results_g2r_hard}
\end{figure}
\begin{figure*}
    \centering
    \includegraphics[width=\textwidth]{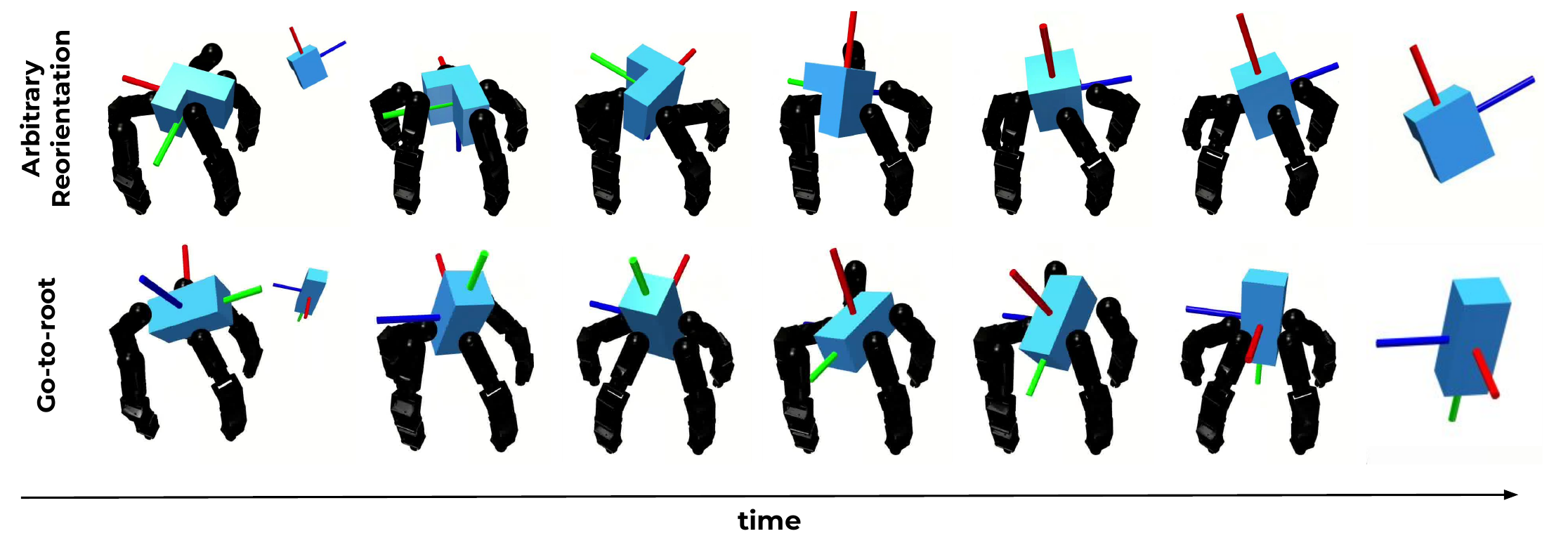}
    \caption{Key frames of goal-reaching tasks, Go-to-root and Arbitrary Reorientation, in simulation.}
    \llabel{fig:gaiting_ar}
\end{figure*}

On the easy object set, we find that all the methods except the ER baseline learn to gait. However, policies that use a random grasp reset distribution (SGS) and sampling-based reset distribution (R$\times$R) achieve higher performance with robust gaits. We also observe that warm starting with Imitation Pre-training (R$\times$R + IPT) results in faster convergence. Unlike our previous studies, we find that FI also learns finger-gaiting. We attribute this to the improved implementation of our asymmetric actor-critic PPO and increased training steps made possible with the help of GPU physics.

However, as we move to more difficult objects in the medium and hard categories, the performance gap between our methods and baselines broadens. For medium difficulty objects, we find that R$\times$R, R$\times$R + IPT, and SGS again all learn to gait, but the policies learned via our methods, R$\times$R and R$\times$R + IPT, are more effective with higher returns.  Again, we observe that warm starting with Imitation Pre-training (R$\times$R + IPT) results in faster convergence and also achieves improved final performance.

Finally, for the hardest object set, a random grasp-based reset distribution is no longer successful. Only R$\times$R and R$\times$R + IPT converge to stable and consistent gaits. Once again, R$\times$R + IPT improves on R$\times$R with respect to rate of convergence and final performance.

Interestingly, FI + IPT baseline that uses a warm start policy (obtained from Imitation Pre-training using the G-RRT tree) performs similarly to FI baseline, failing on all but easy objects. This suggests that exploratory reset distributions may be necessary to derive benefits from Imitation Pre-training. 

To verify the scalability of the method to train policies for multiple objects, we successfully train object-agnostic policies with our R$\times$R approach. To achieve this we simply use object-specific initial distributions in multi-object training. We found the training difficulty to be limited to the hardest object in the set. Importantly, in Sec \lref{sec:realhandeval}, we also verify our policies can be transferred to the real hand.

Next, we present our results in learning policies to reach a desired object orientation i.e. go-to-root and arbitrary reorientation tasks. Additionally, due to the poor performance of many baselines in the previous task, we continue with the best-performing ones. In particular, we continue with SGS and GC baselines. We drop the ER baseline as it uniformly fails across object classes and also drop the FI baseline as fixed initialization is incompatible with both tasks by definition.  
~
\\~\\
\noindent\textbf{Go-to-root}.
We evaluate R$\times$R and R$\times$R + IPT on the Go-to-root task and compare it with SGS and GC baselines. For this task we augment the Gravity Curriculum baseline to use 20 hand-crafted grasps as the reset distribution. These grasps are constructed to appropriately cover the orientation space of the object. We do this to ensure fair comparison between methods, as learning reorientation from a fixed initial orientation to a fixed target is a significantly easier task. 

Our results are summarized in Fig. \lref{fig:results_g2r} \& Fig.~\lref{fig:results_g2r_hard}. As shown by the curves, go-to-root is a harder task requiring between 100M-200M steps. However, these results indicate that our methods consistently outperform baseline methods and produce effective control policies for the task. Again we find that R$\times$R + IPT improves on R$\times$R in both convergence and final performance across all object sets. 

~
\\~\\
\noindent\textbf{Arbitrary Reorientation}. We evaluate R$\times$R on the Arbitrary Reorientation task. Our results in Fig. \lref{fig:results_ar} plot the success rate. Note that we considered the hand in "palm-down" orientation as it tends to be the desired hand orientation in many applications. We demonstrate that our method outperforms SGS on all objects. Interestingly, our method significantly outperforms SGS baseline even for objects in the easy category, with increasing performance for objects in the medium and hard categories. The keyframes of our policy executing this task are shown in Fig. \lref{fig:gaiting_ar}. 

Overall, a common thread of all experiments presented here is that our methods (R$\times$R, R$\times$R + IPT) enable learning a range of challenging dexterous manipulation tasks, while none of the domain-agnostic methods (ER, FI, GC) are able to learn in-hand manipulation on objects beyond the easy set. 


\subsection{Evaluation on real hand}
\llabel{sec:evalrealfg}

\begin{figure*}
    \centering
    \includegraphics[width=\textwidth]{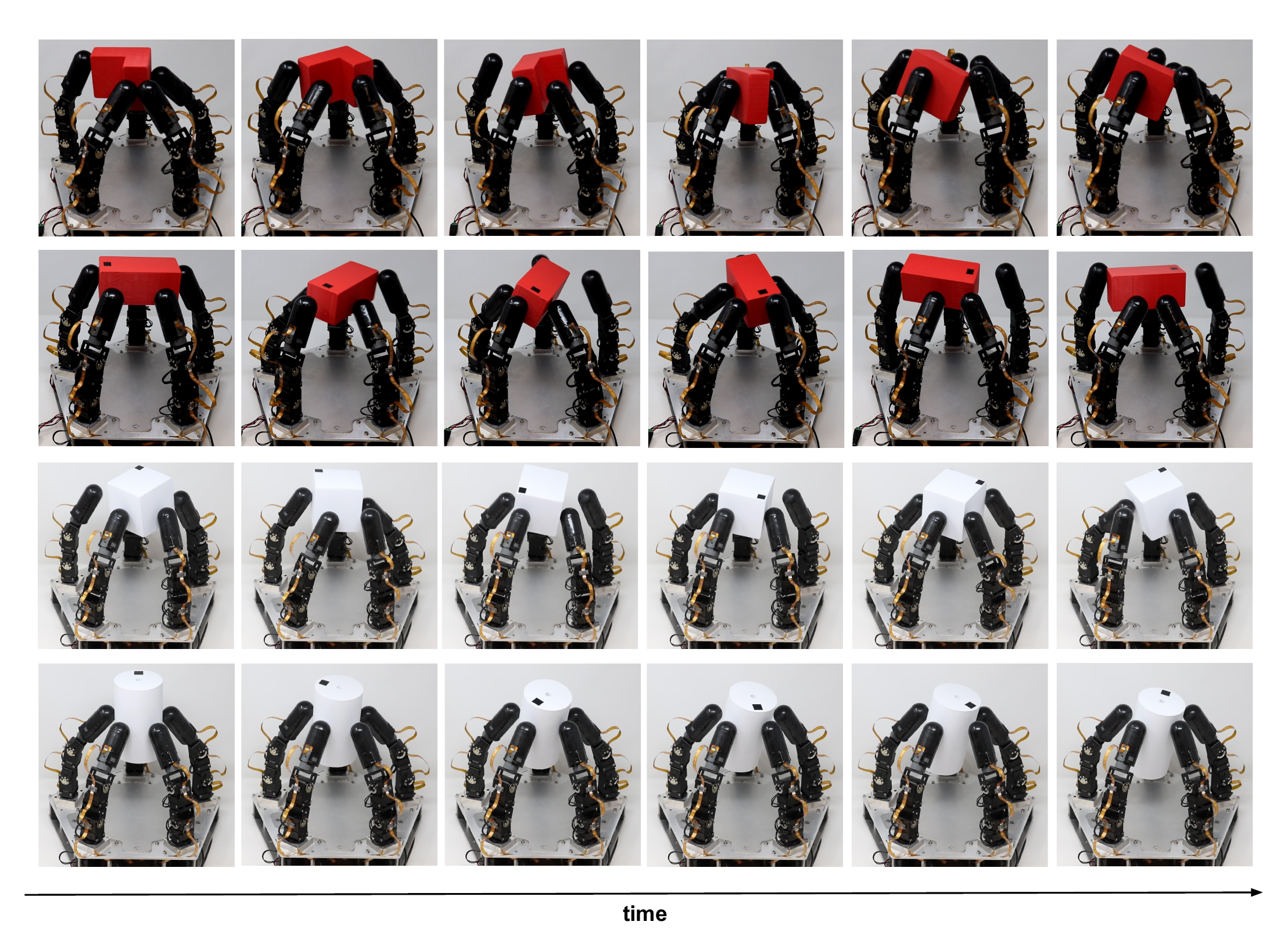}
    \caption{Key frames of the Finger-gaiting policies transferred to the real hand.}
    \llabel{fig:gaiting_real}
\end{figure*}

\begin{figure*}
    \centering
    \includegraphics[width=\textwidth]{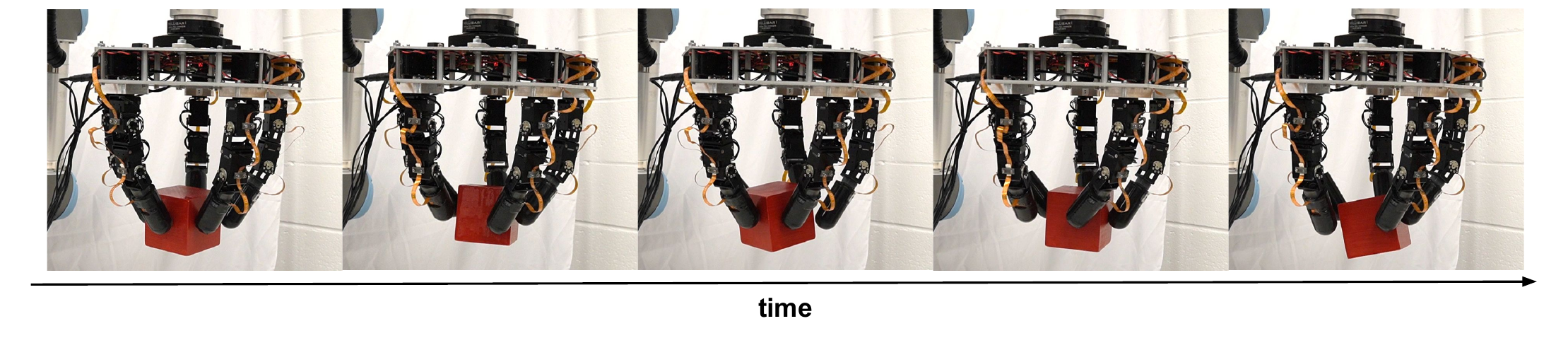}
    \caption{Key frames of a Finger-gaiting policy transferred to the hand in the "palm-down" orientation.}
    \llabel{fig:gaiting_upside_down_real}
\end{figure*}

\llabel{sec:realhandeval}
To test the applicability of our method on real hardware, we attempted to transfer the learned Finger-gaiting policies for a subset of representative objects: cylinder, cube, cuboid \& L-shape. We chose these objects to span the range from simpler to more difficult manipulation skills.

We achieve transfer of the reinforcement learning policies to the real-hardware via domain randomization introduced via a curriculum. We adapt the policies achieved with nominal system parameters i.e. the system parameters used during prior tree generation and reinforcement learning phases with an additional fine-tuning phase where we introduce domain randomization of relevant system parameters. A thorough list of such parameters in detailed next. It is important to note that while randomizing these parameters, the exploratory tree generated with nominal parameters still remains valid as  these parameter perturbations only mildly affected the stability of states. Hence, it is sufficient to use states obtained from this tree generated with with nominal system parameters. 

 First, we impose velocity and torque limits in the simulation, mirroring those used on the real motors ($0.6$ rad/s and $0.5$ N$\cdot$m, respectively). We found that our hardware has a significant latency of $0.05$s, which we included in the simulation. In addition, we modified the angular velocity reward to maintain a desired velocity instead of maximizing the object's angular velocity. We also randomize joint origins ($0.1 \text{ rad}$), friction coefficient ($1-40$), and train with perturbation forces (1 N). All these changes are introduced successively via a curriculum. 

For observation, we used the current position and setpoint from the motor controllers with no additional changes. For tactile data, we found that information from our tactile fingers is most reliable for contact forces above 1 N. We thus did not use reported contact data below this threshold and imposed a similar cutoff in simulation. Overall, we believe that a key advantage of exclusively using proprioceptive data is a smaller sim-to-real gap compared to extrinsic sensors such as cameras. We note that an ablation study illustrating the importance of touch feedback is presented in Sec \lref{sec:ablation}.

For the set of representative objects, we ran the respective policy ten consecutive times and counted the number of successful complete object revolutions achieved before a drop. In other words, five revolutions means the policy successfully rotated the object for $1,800^{\circ}$ before dropping it. In addition, we also report the average object rotation speed observed during the trials.

The results of these trials are summarized in Table~\lref{tab:sim2real}. Fig.~\lref{fig:gaiting_real} and Fig.~\lref{fig:gaiting_upside_down_real} show the keyframes of the real hand finger-gaiting we achieved with our policies. Finally, as our finger-gaiting policies do not rely upon vision feedback, our policies are robust to changes in lighting conditions. Examples of our method operating in dynamic lighting conditions can be found in the accompanying video.  

We note that the starting policies used for sim-to-real transfer were trained with trees generated by a version of G-RRT that used the constraint that at least three fingers must be in contact with the object. However, as previously mentioned, we found that this constraint has minimal effect on the trees generated via G-RRT. Nevertheless, we expect similar sim-to-real performance of policies trained using trees generated after forgoing the three contact constraint. These policies are identical to the policies used for sim-to-real transfer, as per visual comparison in simulation. We also note that preliminary attempts to transfer policies for the cylinder and cube also show similar performance.

\begin{table}
    \centering
    \caption{Manipulation performance in real hardware. We report the median number of object rotations achieved before dropping the object in ten consecutive trials, as well as the time needed to perform these rotations.}
    \ra{1.3}
    \begin{tabular}{ccc}
        \midrule
         \phantom{} & Median  &  Mean rotation \\
         \phantom{} &  revolutions & speed (rad/s)\\
         \midrule
         Cylinder & 5 &  0.42 \\
         Cube (s) &  4.5 & 0.44 \\
         Cuboid & 1.5  & 0.44 \\
         L-shape &  1.5 & 0.24 \\
         \midrule
    \end{tabular}
    \llabel{tab:sim2real}
\end{table}

\section{Ablations}
\llabel{sec:ablation}

\noindent\textbf{Sampling-based Exploration}. First, we conduct ablation of G-RRT for our object set. We aim to discover how effectively the tree explores its available state space given the number of iterations through the main loop (i.e. the number of attempted tree expansions towards a random sample). As a measure of tree growth, we look at the maximum object rotation achieved around our target axis. We note that any rotation beyond approximately $\pi/4$ radians can not be done in-grasp, and requires finger repositioning. Thus, we compare the maximum achieved object rotation vs. the number of expansions attempted (on a log scale). The results are shown in Fig.~\lref{fig:rrtcomp}. 

\begin{figure}[t!]
    \centering
    \includegraphics[width=0.45\textwidth]{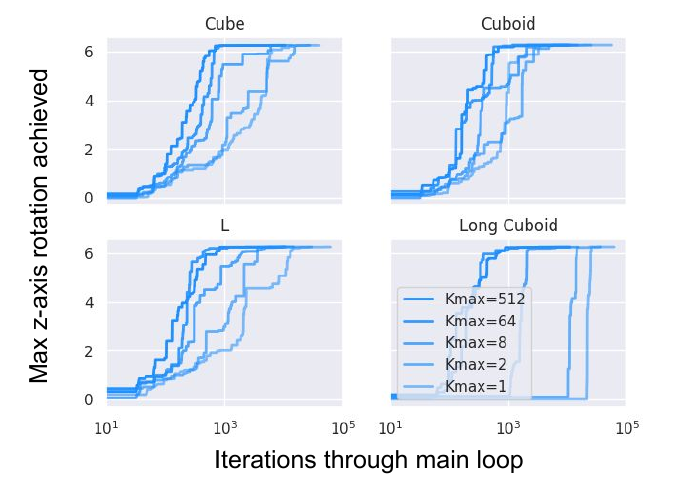}
    \caption{Tree expansion performance for G-RRT. We plot the number of attempted tree expansions (i.e. iterations through the main loop, on a log scale) against the maximum object z-axis rotation achieved by any tree node so far. We plot performance for different values of $K_{max}$, the number of random actions tested at each iteration.}
    \llabel{fig:rrtcomp}
\end{figure}

As expected, the performance of G-RRT improves with the number $K_{max}$ of actions tested at each iteration. Interestingly, the algorithm performs well even with $K_{max}=1$; this is equivalent to a tree node growing in a completely random direction, without any bias towards the intended sample. However, we note that, at each iteration, the node that grows is the closest to the state-space sample taken at the beginning of the loop. This encourages growth at the periphery of the tree and along the real constraint manifolds, and, as shown here, still leads to effective exploration.

We also found that G-RRT is sensitive to the action-scale parameter $\alpha$. Fig \lref{fig:grrt_action_abl} compares exploration as measured by angular distance of the farthest node from the root for varying $\alpha$. Among the various settings evaluated, an $\alpha = 0.15$ was the fastest. Interestingly, higher values of action-scale adversely affects the performance of G-RRT. Furthermore, as we will discusses later, not only does $\alpha$ affect the rate of exploration it also impacts the quality of extracted paths. 

\begin{figure}[h]
    \centering
    \includegraphics[width=0.4\textwidth]{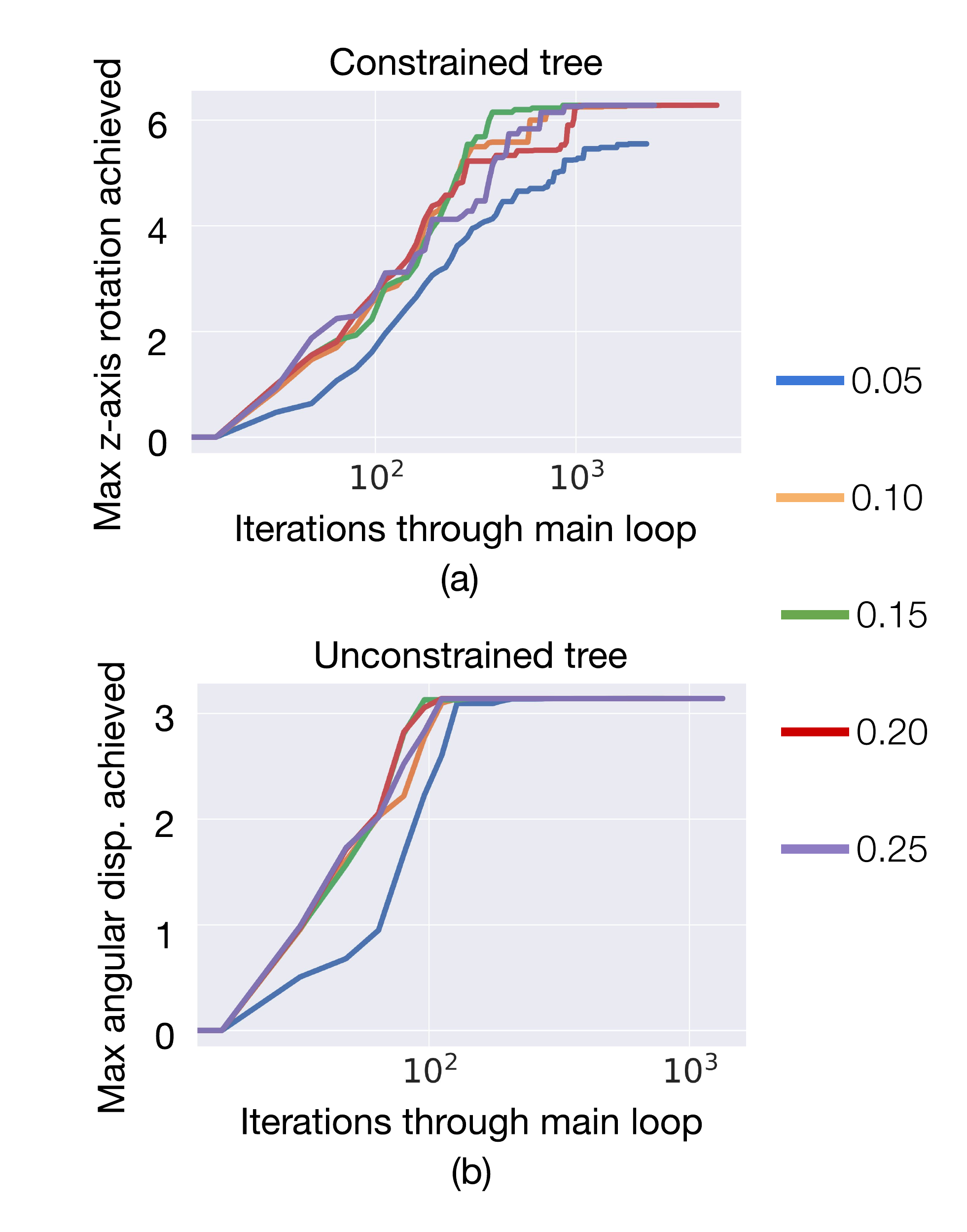}
    \caption{G-RRT action-scale ($\alpha$) ablation. $\alpha = 0.15$ is optimal.}
    \llabel{fig:grrt_action_abl}
\end{figure}

\vspace{3mm}
\noindent\textbf{Reinforcement Learning}. We now consider the ablation of training methods. We perform these ablations on the Finger-gaiting and Go-to-root tasks on a non-convex, L-shaped object. 

We begin by studying the impact of the size of the tree used in extracting reset states. Fig~\lref{fig:treesize}(a) summarizes our results for learning a Finger-gaiting policy using trees of different sizes grown via G-RRT. Qualitatively, we observe that, as the tree grows larger, the top 100-400 paths sampled from the tree contain increasingly more effective gaits, likely closer to the optimal policy. We see that we need a sufficiently large tree with at least $10^4$ nodes to enable learning. However, training is most reliable with $10^5$ nodes. This suggests a strong correlation between the quality of states used for reset distribution and sample efficiency of learning. 

Fig~\lref{fig:treesize}(b) summarizes the tree-size ablation results for learning the Go-to-root task. For a very small tree, we see near 100\% success rate at training time. This is rather unsurprising as, at these small tree sizes, the nodes are still close to the root orientation resulting in a trivial reorientation task. As we expect, validation performance on paths extracted from the largest tree increases as the size of the tree generating the initial state distribution increases.
\begin{figure}[]
    \centering
    \includegraphics[width=0.5\textwidth]{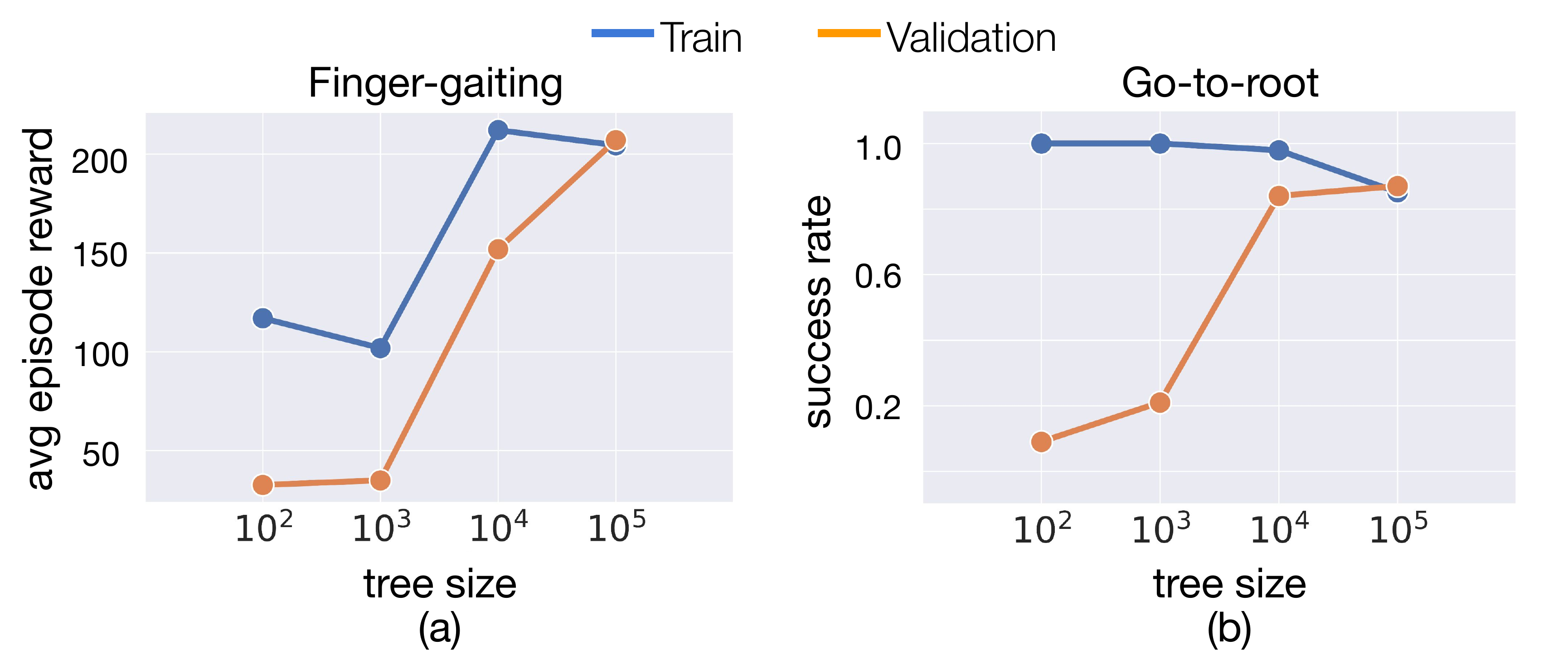}
    \caption{The final performance of policies for Finger-gaiting (left) and Go-to-root (right) tasks using trees of increasing sizes. Validation performance is with reset states from paths extracted from the largest tree.}
    \llabel{fig:treesize}
\end{figure}

We also reconsider the action-scale parameter. Besides affecting the rate of exploration it also impacts the quality of paths.  We can infer this by the impact on training performance in Fig \lref{fig:rl_action_abl}. We see a similar trend, as the training performance peaks again around $\alpha = 0.15$ for both Finger-gaiting and Go-to-root. 

\begin{figure}[t]
    \centering
    \includegraphics[width=0.5\textwidth]{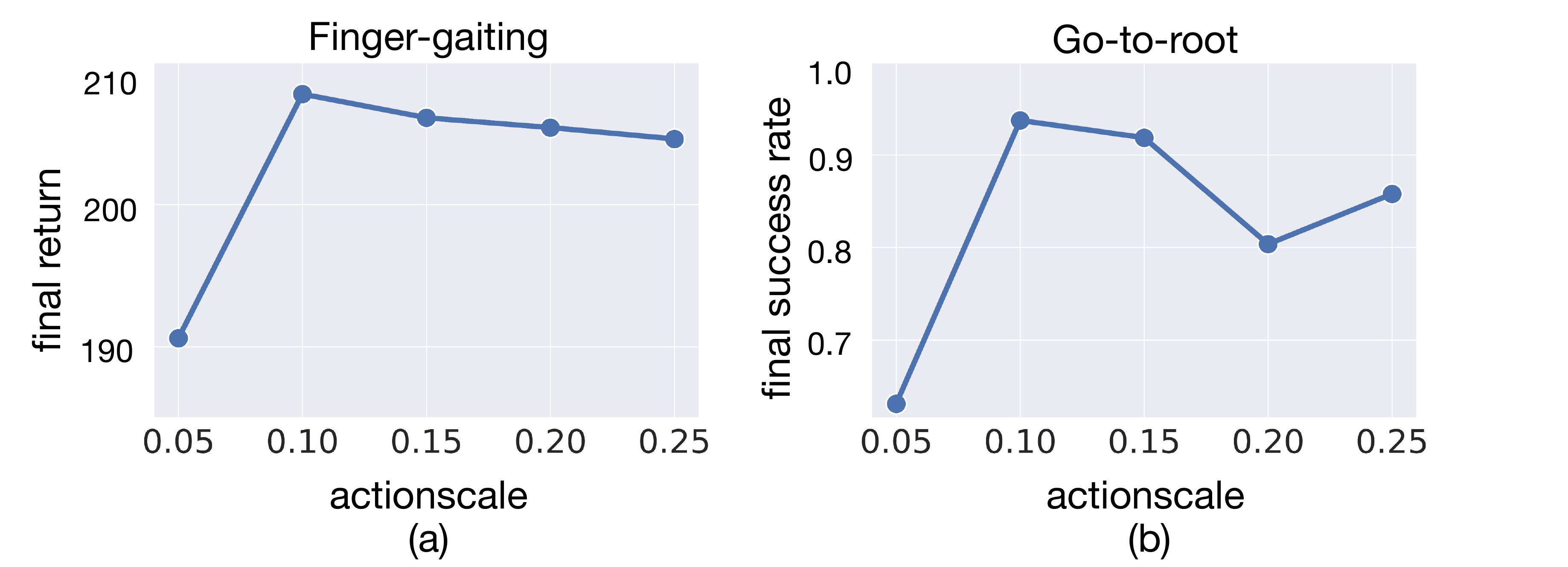}
    \caption{Action scale ablation for the Finger-gaiting and Go-to-root tasks with L.}
    \llabel{fig:rl_action_abl}
\end{figure}

\begin{figure}[t]
    \centering
    \includegraphics[width=0.4\textwidth]{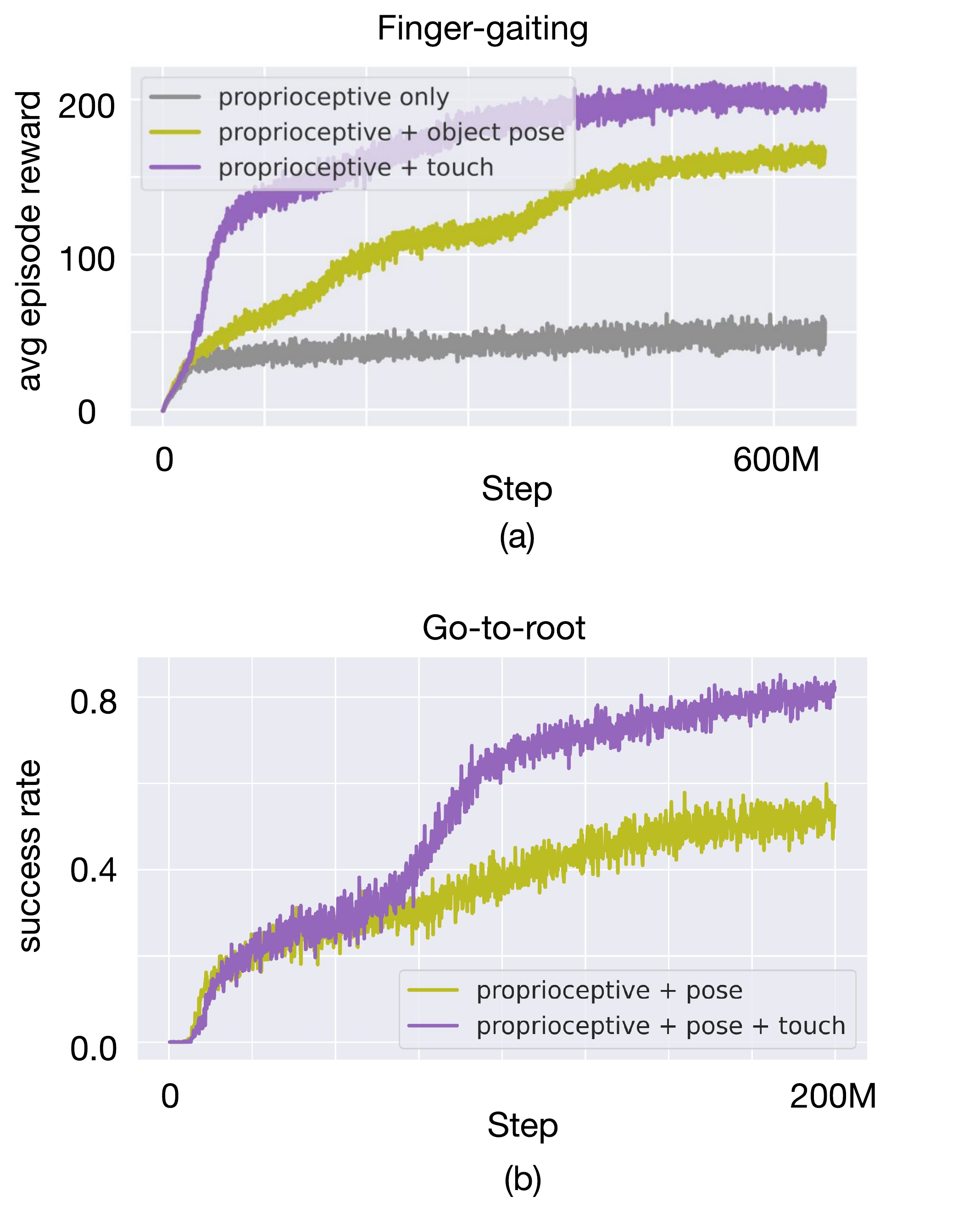}
    \caption{Ablation of policy feedback components which highlights the importance of touch feedback.}
    \llabel{fig:sense_abltn}
\end{figure}

We also conducted an ablation study of policy feedback. Particularly, we aimed to compare intrinsically available tactile feedback vs. object pose feedback that would require external sensing. Fig.~\lref{fig:sense_abltn} summarizes these results which demonstrate the importance of tactile feedback for both Finger-gaiting and Go-to-root tasks. 

In the Finger-gaiting task, we found that touch feedback is essential in the absence of object pose feedback for all moderate and hard objects. For these objects, we also saw that replacing this tactile feedback with object pose feedback results in slower learning. Similarly, in the Go-to-root task, leaving out touch feedback from policy input also results in slower learning. 

These results underscore the importance of touch feedback for in-hand manipulation skills. Richer tactile feedback such as contact position, normals, and force magnitude can be expected to provide even stronger improvements; we hope to explore this in future work.

\section{Discussion and Conclusions}

The results we have presented show that sampling-based exploration methods make it possible to achieve difficult manipulation tasks via RL. In fact, these popular and widely used classes of algorithms are highly complementary in this case. RL is effective at learning closed-loop control policies that maintain the local stability needed for manipulation, and, thanks to training on a large number of examples, are robust to variations in the encountered states. However, the standard RL exploration techniques (random perturbations in action space) are ineffective in the highly constrained state space with a complex manifold structure of manipulation tasks. Conversely, SBP methods, which rely on a fundamentally different approach to exploration, can effectively discover relevant regions of the state space.

We present multiple methods for conveying exploration information derived through SBP to RL training algorithms. In particular, explored states can be used as an effective reset distribution to enable learning. The transitions between states used during sample-based exploration are also useful surrogates for actions, and can thus be used in an imitation pre-training stage to boost learning. Imitation pre-training serves as an effective strategy to leverage the inherent structure accessible through sampling-based plans. This approach becomes particularly valuable in addressing the challenges of hard motor control tasks within reinforcement learning, not only by facilitating training but also by providing a favorable initialization of actor and critic networks. Recent work in this domain extends the use of the imitation policy beyond mere initialization by demonstrating further benefits when incorporated throughout the training process. This is a promising avenue for achieving further performance gains in reinforcement learning scenarios for hard motor control tasks.

We use our approach to demonstrate a number of dexterous manipulation skills: achieving large in-hand reorientation about a given axis via finger-gaiting and reorientation to a desired object pose, which is either pre-set or randomized. Importantly, we demonstrate finger gaiting precision manipulation of both convex and non-convex objects, using only tactile and proprioceptive sensing. Using only these types of intrinsic sensors makes manipulation skills insensitive to occlusion, illumination, or distractors, and reduces the sim-to-real gap. We take advantage of this by demonstrating our approach both in simulation and on real hardware. We note that, while some applications naturally preclude the use of vision (e.g. extracting an object from a bag), we expect that in many real-life situations, future robotic manipulators will achieve the best performance by combining touch, proprioception, and vision. 

In future work, we believe that our approach can be scaled to tackle even more demanding motor control tasks with numerous degrees of freedom, including both prehensile and non-prehensile manipulation. Bi-manual manipulation seems like a natural candidate application, particularly if it involves coordination with multi-fingered dexterous hands. We also believe our approach can be applied beyond dexterous manipulation, such as in learning challenging agile locomotion skills with minimal changes. Towards this aim, we suggest expanding the sampling configuration space to the complete state of the robot with first order derivatives and also replacing the stability check with necessary boundary constraints on the state. With these changes in place, we believe it is feasible recipe to accelerate learning an end-to-end policy for learning agile skills such as robot parkour \cite{Zhuang2023-bi}. Finally,  sampling-based exploration techniques could be integrated directly into the policy training mechanisms, removing the need for two separate stages during training. In fact, we expand on this idea in next chapter.
\\~\\


\chapter{Structured Exploration and Exploitation}

Real-world applications of multi-finger dexterous manipulation are often characterized by two key features. First, the tasks are long-horizon, involving various sequences of dexterous skills such as grasping and in-hand reorientation. Second, the object shapes are highly eccentric, as the manipulation often involves tools like hammers, pens, and other slender objects.

Such long-horizon dexterous manipulation skills for slender objects have primarily been achieved through the use of demonstrations, which help circumvent the issue of exploration. Since demonstrations explore the relevant state space for the task, the need for extensive exploration is minimized. These demonstrations are typically employed with imitation learning, demo-augmented reinforcement learning \cite{Rajeswaran2017-uu, Rajeswaran2017-uu}, or offline reinforcement learning \cite{Nair2020-gq}, each achieving varying degrees of success. High-capacity models predicting action chunks \cite{Zhao2023-jk,Zhao2023-jk, Lee2024-tu}, have further advanced dexterous manipulation skills but with simple grippers.

Nonetheless, as discussed in Chapter 1, a major limitation for long-horizon multi-fingered dexterous manipulation is the difficulty of obtaining demonstrations. Teleoperation, which lacks natural tactile feedback, becomes especially challenging due to the inherently unstable dynamics involved in in-hand manipulation. This instability makes it hard to accurately demonstrate the fine motor control required for complex tasks.

Thus, we explore an alternative source of demonstrations by turning to structured exploration through sampling-based planning. While structured exploration using tree-based planning methods like G-RRT has been crucial for in-hand dexterity (Chapter 3), scaling such approaches to very long-horizon, multi-fingered manipulation tasks is challenging. The main limitation lies in the inability of tree-based sampling to effectively handle long-horizon planning, especially in tasks where exploration is difficult. This issue is further compounded in tasks with unstable dynamics, such as multi-fingered dexterous manipulation. 

However, the well-known Probabilistic Roadmaps (PRMs) \cite{Kavraki1996-gr} do not suffer from these limitations and can be applied to long-horizon planning. That said, there are fundamental challenges in adopting PRMs for high-dimensional state spaces, where it becomes difficult to make the necessary assumptions for their effectiveness.

We propose Diffusion Roadmaps, integrated into an online policy learning framework we call Rapidly-exploring Reinforcement Learning (RRL). The Diffusion Roadmaps algorithm adapts Probabilistic Roadmaps (PRMs) to handle complex high-dimensional states by utilizing diffusion policies. RRL involves applying reinforcement learning to the constructed roadmap graphs, enabling the learning of both a diffusion policy local planner for PRMs and a diffusion policy task policy for long-horizon dexterous manipulation.

\section{Diffusion Roadmaps}

Diffusion Roadmaps address a fundamental limitation in adopting PRMs for high-dimensional state spaces where the useful state space has a manifold structure. PRMs involve building a connected graph of randomly sampled valid configurations with the help of a local planner, wherein the chief issue is that the local planner establishing connectivity between two nodes is hard to construct and often unavailable.  It is also not straightforward to sample random configurations uniformly throughout the state space.

In Diffusion Roadmaps, we use diffusion policy as a local planner, and we build the roadmap graph by inverting the order of operations. First, we add edges obtained by the diffusion policy and then merge intersecting nodes instead of adding nodes first and finding connections between nodes. 

Let $F$ be the forward model, i.e. $\bm{x}_{t+1} = F(\bm{x}_{t+1}, \bm{a}_{t})$ where $\bm{x}_t$ and $\bm{a}_t$ are the state and action at timestep $t$. Our Diffusion Roadmaps algorithm with diffusion policy $\pi_{\eta}$ is described in Alg \lref{alg:DRM} with operations involved described in Alg \lref{alg:drmops}. The key difference between PRM and DRM is that, in DRM, an edge is added by rolling out actions obtained from a diffusion policy, whereas in PRM, an edge is added by checking connectivity between two nodes. This connectivity check routine is not feasible in high-dimensional, complex state spaces.

\begin{algorithm}[t]
\caption{Diffusion Roadmaps}\llabel{alg:DRM}
\begin{algorithmic}[1]
\State Initialize parameters $\eta$ of diffusion policy $\pi_{\eta}$
\State Initialize graph $G \gets \{\} $ or existing graph
\Procedure{DRM}{$G$, $N_{max}$}
\While{$N < N_{max}$}
\State  $\bm{x}_{sample} \gets SAMPLE()$  with probability $\alpha$, OR $\bm{x}_{sample} \gets $ a random node in $G$
\State $\bm{x}_{start} \gets $ ADD-NODE($\bm{x}_{sample}$) \Comment{Returns the added node if merge is performed}
\State Sample $x_{goal}$ in the neighborhood of $\bm{x}_{start}$
\State Sample $K$-step action-chunk
\Statex ~~~~~~~~ $A \sim \pi_{\eta}( . | \bm{x}_{start}, x_{goal})$
\State Rollout for $K$ steps with forward model $F$
\State $\bm{x}_{end} \gets$ Read resulting simulation state
\State $\tau\gets$ rollout from $\bm{x}_{start}$ to $\bm{x}_{end}$
\If{VALID$(\bm{x}_{end})$}
\State $\bm{x}_{end}$ ADD-NODE($\bm{x}_{end}$)
\State ADD-EDGE($\bm{x}_{start}$, $\bm{x}_{end}$, $\tau$)
\EndIf
\EndWhile
\EndProcedure
\end{algorithmic}
\end{algorithm}

\begin{algorithm}[h!]
\caption{}\llabel{alg:drmops}
\begin{algorithmic}[1]
\Statex {\color{gray} \# Verify validity of state}
\Statex \textbf{Require:} State-space bounds $\bm{x}_{low}$ \& $\bm{x}_{high}$, constraint functions $h_1, \ldots, h_C$ 
\Procedure{VALID}{$\bm{x}$}
\State $valid \gets True$
\State $valid \gets  valid \land (\bm{x} > \bm{x}_{low})$ 
\State $valid \gets  valid \land (\bm{x} < \bm{x}_{high})$ 
\State $valid \gets valid \land (h_1(\bm{x}) > 0) \land ... \land (h_C(\bm{x}) > 0)$ \\
\Return $valid$
\EndProcedure
\Statex ~
\Statex {\color{gray} \# Sampling a new state}
\Statex \textbf{Parameters:}  State-space bounds $\bm{x}_{low}$, $\bm{x}_{high}$
\Procedure{SAMPLE}{}
\Repeat
\State $\bm{x}_{sample} \sim \mathcal{U}(\bm{x}_{low}, \bm{x}_{high})$ \Comment{uniform sampling}
\State Apply $\bm{x}_{sample}$ to the simulated environment
\State $\bm{x}'_{sampled} \gets $ Read state from the simulation
\Until{VALID({$\bm{x}'_{sampled}$})}
\State \Return $\bm{x}'_{sampled}$
\EndProcedure
\Statex ~
\Statex {\color{gray} \# Add a new node}
\Statex \textbf{Require:} Graph $G$, merge threshold $\epsilon$
\Procedure{ADD-NODE}{$\bm{x}$}
\State $\bm{x}_{nearest}$ $\gets$ node closest to $\bm{x}$ in $G$
\If{$\|\bm{x}_{nearest} - \bm{x} \| < \epsilon $}
\State \Return $\bm{x}_{nearest}$
\Else
\State $G \gets G \cup \bm{x} $ 
\State \Return $\bm{x}$
\EndIf
\EndProcedure
\Statex ~
\Statex {\color{gray} \# Add a new edge}
\Statex \textbf{Require:} Graph $G$
\Procedure{ADD-EDGE}{$\bm{x}_{start}$, $\bm{x}_{end}$, $\tau$}
\State $\bm{x}_{start}.children \gets \bm{x}_{start}.children \cup \bm{x}_{end}$
\State $\bm{x}_{end}.parent \gets \bm{x}_{start}.parents \cup \bm{x}_{start}$
\State $\bm{x}_{start}.edges \gets \bm{x}_{start}.edges \cup \tau $
\EndProcedure
\end{algorithmic}
\end{algorithm}

\clearpage
 While Alg \lref{alg:DRM} demonstrates how the diffusion planner samples action chunks to grow the roadmap graph, we have not yet addressed a critically important component. To effectively explore high-dimensional state spaces with Diffusion PRM, the diffusion model is integrated into an online learning framework that alternates between planning and learning. Using this framework, we also learn the policy for long-horizon dexterous manipulation skills, which we will describe next.

\section{Rapidly-exploring Reinforcement Learning}

Rapidly-exploring Reinforcement Learning (RRL) is an online RL framework for learning complex motor skills such as long-horizon multi-fingered dexterous manipulation skills. It alternates between structured exploration involving sampling-based planning and structured exploitation involving reinforcement learning on the explored structure representing the useful manifold of the problem state space. An essential element of RRL is a virtuous cycle between the two phases, established by improving the local planner during the exploitation phase, which leads to cascading improvements in the subsequent exploration and exploitation phases.

Consider the task of learning policy $\pi_{\theta} (A_t | \bm{o}_t)$ where $\bm{o}_t$ and $A_t = (\bm{a}_t, ..\bm{a}_{t+K})$ are observation and action chunks at time step $t$. Given reward function $r$ and discount factor $\gamma$, the objective is, as usual, to maximize the expected returns $\mathrm{E}[\sum_{t=0}^{\infty} \gamma^{t-1}r_{t}]$,  starting from state $\bm{x}_0$ sampled from $\mu_{eval}$, the initial state distribution for evaluation.

For structured exploration, we use Diffusion Roadmaps, which were introduced previously. The exploitation phase consists of reinforcement learning on the constructed roadmaps. Hence, it involves collecting experience while restricted to the graphs in the roadmaps. Then, this experience will be used to update the policy. While any off-policy agent suffices, we use Diffusion Q-Learning to learn a diffusion policy. Experience extracted from such graphs is naturally multi-modal, i.e., with many parallel paths between two nodes. Finally, the task policy is also learned via Diffusion Q-learning

Rapidly-exploring Reinforcement Learning with Diffusion Roadmaps and Diffusion Q-Learning is described in Alg \lref{alg:rrl+drm}.

\begin{algorithm}[t]
\caption{Rapidly-exploring Reinforcement Learning with Diffusion Roadmaps}\llabel{alg:rrl+drm}
\begin{algorithmic}[1]
\Statex Initialize task diffusion policy $\pi_{\theta}$, planner diffusion policy  $\pi_{\eta}$
\Statex Initialize value networks $Q_{\phi_1}$ and $Q_{\phi_2}$
\Statex Initialize empty graph $G$
\For{each iteration}
    \Statex \hspace{\algorithmicindent} {\color{gray} \# Planning}
    \State $G \gets$ DRM($G$, $\pi_{\eta}$, $N_{max}$)
    \Comment{Alg \lref{alg:DRM}}
    \State $\mathcal{W}_1 \gets $ Extract rollout on from roadmaps G w.r.t $Q_{\phi_1}$ \Comment{Alg \lref{alg:extract_plans}} 
    \Statex \hspace{\algorithmicindent} {\color{gray} \# Planner policy update}
    \State Update $Q_{\phi_1}$ with Equation \leqref{eq:valupdate} using $r=r_{plan}$, $\pi=\pi_\eta$, $\mathcal{W} = \mathcal{W}_1$
    \State Update planning policy $\pi_{\eta}$  to minimize loss in Equation \leqref{eq:policyloss} using $\epsilon=\epsilon_\eta$, $Q=Q_{\phi_1}$
    \Statex \hspace{\algorithmicindent} {\color{gray} \# Task policy update}
    \State Update $Q_{\phi_2}$ with Equation \leqref{eq:valupdate}
    \State $\mathcal{W}_2 \gets $ Extract rollout on from roadmaps G w.r.t $Q_{\phi_2}$ \Comment{Alg \lref{alg:extract_plans}} 
    \State Update task policy $\pi_{\theta}$ with minimize loss in Equation \leqref{eq:policyloss} using $\epsilon=\epsilon_\theta$, $Q=Q_{\phi_2}$
\EndFor
\end{algorithmic}
\end{algorithm}

\begin{algorithm}[h!]
\caption{Extract rollouts on roadmaps}\llabel{alg:extract_plans}
\begin{algorithmic}[1]
\Require Graph $G$
\State $\mathcal{W} \gets \{\}, P \gets 1$
\While{$P<P_{max}$}
\State $l \gets 0$
\State $\tau = \{\bm{x}_{node}\}$ 
\State $\bm{x}_{node} \gets$ random node in $G$
\While{$\bm{x}_{node}$ has children}
\State Select $\bm{x}_{child}$ based on its probability $\exp[{{Q(\bm{x}_{node}, A_{child})/\tau}}]$ 
\State $\tau \gets \tau \cup \bm{x}_{child}$, $\bm{x}_{node} \gets \bm{x}_{child}$
\State $l \gets l + 1$
\EndWhile
\State $\bm{x}_{goal} \gets \bm{x}_{child}$ \Comment{Update the goal as last child}
\ForAll {$\bm{s}$ in $\tau$}
\State Update goal in $s$ with $\bm{x}_{goal}$
\EndFor 
\State $\mathcal{W} \gets \mathcal{W} \cup \tau$,  $P \gets P + 1$ \Comment{Add the walk to buffer}
\EndWhile
\end{algorithmic}
\end{algorithm}





\begin{equation}
    \displaystyle
    \underset{{\substack{(\bm{x}_t, A_{t}, \bm{x}_{t+K}) \sim \tau \\ \tau \sim \mathcal{W}\\A_{t+K} \sim \pi(.| \bm{x}_{t+K})}}}{\mathbb{E}}\| r + \gamma Q(\bm{x}_{t+K}, A_{t+K}) - Q(\bm{x}_t, A_t)\|^2
    \llabel{eq:valupdate}
\end{equation}

\begin{align}
    \mathcal{L} &= \mathcal{L}^{diffusion} + \mathcal{L}^{Q-value}\nonumber \\ 
                &= \underset{\substack{k\sim 0 \ldots N-1 \\ (\bm{x}_t, A_t^0) \sim \mathcal{W}}}{\mathbb{E}}(\epsilon^k, \epsilon_\eta(\bm{x}_t, A_t^0 + \epsilon^k, k)) \nonumber\hfill ~~\text{(diffusion loss)}\\  &~~~~~~-\alpha\underset{(\bm{x}_t, A_t^0) \sim \mathcal{W}}{\mathbb{E}} \|Q_{\phi_1}(\bm{x}_t, A_t^0) \|^2 ~~~~~ \text{(value loss)}
                \llabel{eq:policyloss}
\end{align}





\section{Experiments \& Results}
Our results are preliminary. We evaluated RRL with Diffusion Roadmaps \& Diffusion Q-Learning considering difficult 2D navigation tasks with hard-to-explore mazes.

\subsection{2D Maze Navigation}
The mazes used for evaluation are shown in Fig. \lref{fig:mazes} where the task is to start from a fixed position in the center and reach a random sampled goal.  For these tasks, we used the dense reward function given by Eq (\lref{eq:maze_reward}) where $\bm{x}$, $\bm{x}_g$ are the current and goal positions and $r_b$ is the bonus reward for reaching the goal within the radius $\epsilon$. We used the IssacLab simulation for parallelization.


\begin{figure}
    \centering
    \includegraphics[width=0.8\columnwidth]{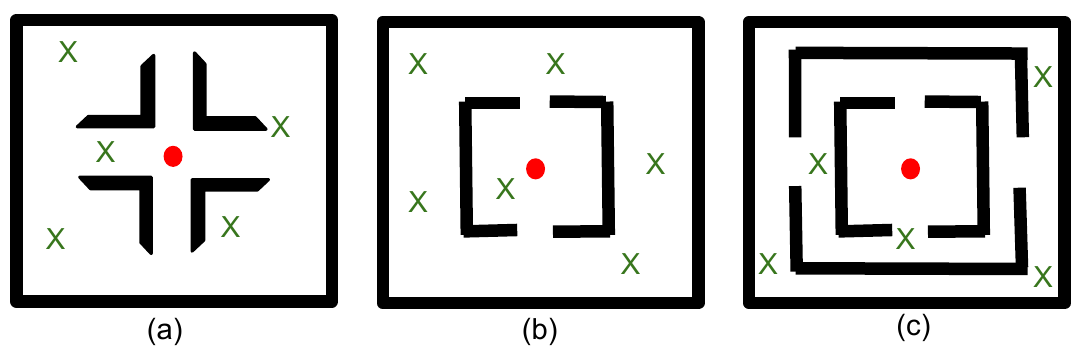}
    \caption{Mazes used for 2D navigation where the task is to reach navigate between fixed start and randomly sampled goal positions. Mazes (a) through (c) are in increasing order of difficulty.}
    \llabel{fig:mazes}
\end{figure}

\begin{equation}
    \llabel{eq:maze_reward}
    r = -\| \bm{x} - \bm{x}_g  \|^2+ r_b \bm{1}[ \| \bm{x} - \bm{x}_g \|< \epsilon]
\end{equation}

\subsection{Finger-gaiting In-hand Manipulation}

\subsection{Comparisions}

\noindent\textit{RRL(G-RRT+ DAPG)}: We compare with Rapidly-exploring Reinforcement Learning performing structured exploration with non-holonomic RRT (i.e. G-RRT) and structured exploitation with DAPG\cite{Rajeswaran2017-au}. The demonstrations for DAPG are extracted from the exploration tree, where we can find nearest states to randomly sampled goals and use the path to the root from states as optimal plans. The policy improvement step consists of minimizing the behavior cloning loss over the extracted optimal plans in addition to the RL objective as shown Eq \leqref{eq:L_v_loss}. Here, we use PPO as the RL algorithm. 

\begin{equation}
    \llabel{eq:L_v_loss}
    L(\theta) = L_{PPO}(\theta) + \lambda_0\lambda^k_1L_{BC}(\theta)
\end{equation}

\begin{algorithm}[h!]
\caption{RRL-DAPG}\llabel{alg:rrl_dapg}
\begin{algorithmic}[1]
{\small
\Require Initialize policy $\pi_{\theta}$ and value networks $V_{plan}$, $V_{task}$
\For{n = 1, \ldots}
\State Execute planner iterations G-RRT (Alg \ref{5:alg:g-rrt})
\State Extract optimal plans from RRT and store in $D$ (Alg \lref{alg:extract_plans})
\State Update reset distribution $\rho_{train}$
\Statex \texttt{\# Compute MSE loss of BC}
\State Compute $L_{BC}(\theta)$  using $D$ 
\Statex \texttt{\# Compute loss of RL}
\State Rollout task policy $\pi_{\theta}$ and store rollouts in $\mathcal{R}_{task}$
\State Compute $L_{PPO}(\theta)$ with $\mathcal{R}_{task}$
\Statex \texttt{\# Update policy}
\State Update policy $\pi_{\theta}$ with the gradient of $L_{DAPG}(\theta)$ $\left(L_{DAPG}(\theta)=L_{PPO}(\theta) + \lambda_0\lambda^k_1L_{BC}(\theta)\right)$
\State Update $V_{task}$ with $\mathcal{R}_{task}$
\EndFor
}
\end{algorithmic}
\end{algorithm}

Finally, for completeness, we also compared RL (PPO) with and without a custom hand-designed reset distribution. The custom reset distribution is straightforward to obtain by uniformly covering the state space. We denote these baselines as \textit{RL ($\rho_{train} = \rho_{custom}$)} and \textit{RL ($\rho_{train} = \rho_{start}$)}. Both baselines perform poorly, failing to achieve a success rate greater than 50\%.

\subsection{Results}

\subsubsection{2D Maze Navigation}
Our results show that RRL (DRM + DiffQL) outperforms all other baselines, achieving over a 90\% success rate across all mazes, including maze (c), the most challenging one. In comparison, the next best baseline achieves a success rate only 76\% for maze (c). 

 \begin{table*}[h]
    \centering
    \caption{The success rate for evaluation reset distribution $\rho_{eval} = \{\bm{x}_{start}\}$ at the center of the maze.} 
    \small{
    \ra{1.1}
    \begin{tabular}{cccc}
    \midrule
    \phantom{a}  & \multicolumn{3}{c}{\textbf{Task /Maze}} \\
    \midrule
    \textbf{Method} & a  & b & c \\
    \midrule
     RRL(DRM + DiffQL) & \textbf{94\%} & \textbf{99\%} & \textbf{92\%} \\
     RRL(G-RRT + DAPG) & \textbf{96\%} & \textbf{99\%} & \textbf{76\%} \\
     RL ($\rho_{train} = \rho_{custom}$) &  40\% & 51\% & 31\% \\
     RL ($\rho_{train} = \{\bm{x}_{start}\}$) & 0\% & 0\% & 0\% \\
    \midrule
    \end{tabular}
    }
    \llabel{tab:success_rate}
\end{table*}

\begin{figure}
    \centering
    \includegraphics[width=0.8\columnwidth]{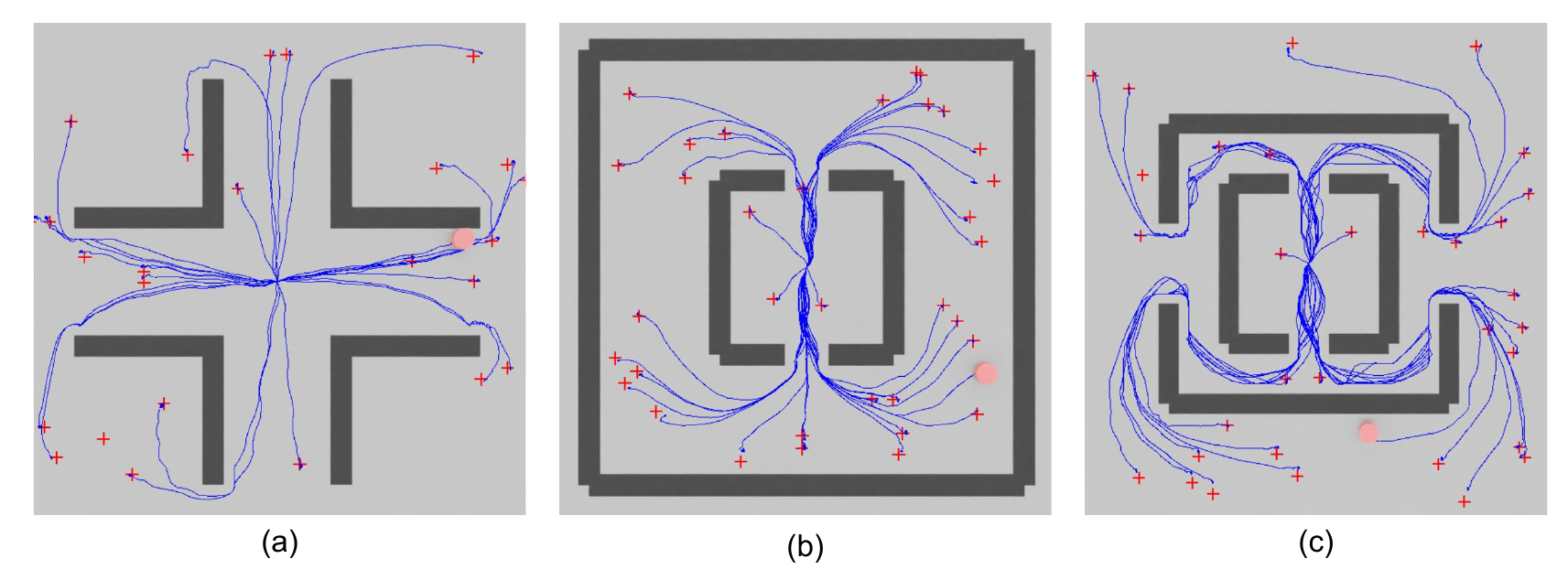}
    \caption{ The rollouts achieved by our policy in mazes (a) through (c) are in increasing order of difficulty.}
    \llabel{fig:mazes}
\end{figure}

\subsubsection{Finger-gaiting In-hand Manipulation}

Diffusion Roadmaps successfully facilitate the desired exploration for finger-gaiting in-hand manipulation, specifically reorientation about the z-axis. Fig shows the maximum rotation achieved about the z-axis, along with the average of the maximum rotations across randomly sampled nodes, as the roadmap graph expands. Not only does the Diffusion Roadmaps method achieve a full $2\pi$ rotation, but the results also indicate the existence of numerous paths within the graph that enable high degrees of reorientation. Training policies using Diffusion Roadmaps and RRL for finger-gaiting remains an exciting direction for future work.

\begin{figure}
    \centering
    \includegraphics[width=\linewidth]{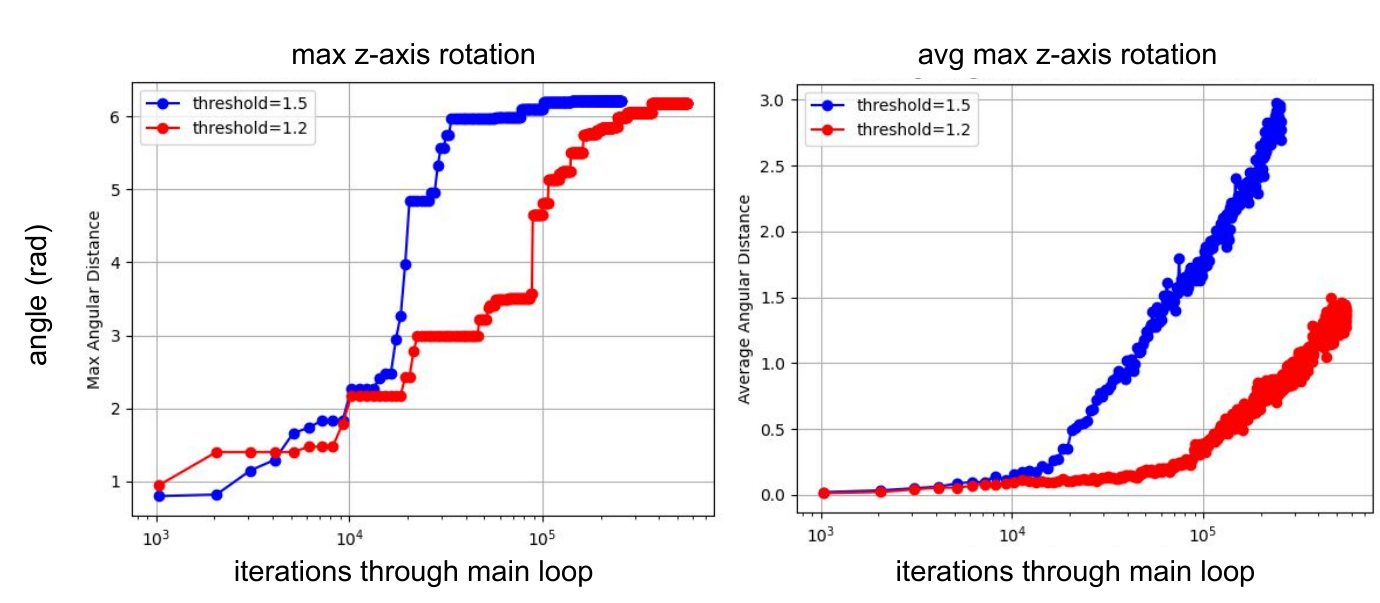}
    \caption{Angular rotation via Diffusion Roadmaps for cube for  node merging thresholds $\epsilon$.}
    \llabel{fig:ihm_fg_drm}
\end{figure}

\section{Conclusion and ongoing work}
To enable structured exploration and exploitation in reinforcement learning, we introduced Diffusion RoadMaps (DRM) for exploration, coupled with Diffusion Q-Learning (Diffusion QL) to derive effective policies from the constructed PRM graph. Preliminary experiments on challenging maze navigation tasks highlight the potential of integrating PRM-based planning with reinforcement learning. Our ongoing work extends this approach to in-hand reorientation of complex objects and long-horizon skills—tasks that underscore the need for structured exploitation strategies.

\section{Limitations}

Reinforcement learning with structured exploration in a simulated environment can be used to learn highly dexterous multi-fingered manipulation skills as we showed in previous chapters.  However, there are some key limitations to motor learning in a simulated environment. Although numerous methods are proposed to address challenges of simulation-to-real, it still remains difficult to bridge this gap. Scaling up simulations to develop diverse and generalizable skills is a time and labor intensive process. Building the required simulation infrastructure is costly and often complicated. Hence, in the following chapter we consider imitation learning.


\chapter{Dexterity from Human Demonstrations}
Imitation learning is proving to be an effective approach for learning challenging robot manipulation skills, particularly enabled by high capacity models and the tools to collect high quality demonstrations.
It is naturally capable of circumventing the sim-to-real gap, a major drawback of reinforcement learning methods for hard dexterous skills, as demonstrations are already on the real robot.

Nonetheless,  imitation learning has its own limitations. The teleoperation setups which are required in large numbers keep the infrastructure cost high. There is yet another limitation, which is arguably a fundamental limit on the dexterity of demonstrations that can be collected. 

Multi-fingered dexterity where tactile sensing is critical is indeed such a situation. Imitation by observing the hand directly such as with visual demonstration (video) sidesteps the limits on dexterity imposed by teleportation. However, visual demonstrations alone without tactile sensing are not very helpful in learning dexterity.

Taking the view human hand is merely another embodiment with different dynamics, action space etc., we propose equipping the human hand with additional sensing modalities, particularly tactile sensing. Learning with this paradigm shift in human data collection may have massive potential to achieve high levels of dexterity. Therefore, we propose a new paradigm for human demonstrations by equipping the human hand with wearable tactile sensors to collect data. Henceforth, we redefine human demonstrations to refer to these visuo-tactile demonstrations. 

We then propose a novel framework for learning dexterous skills from these demonstrations. In human-to-robot cross-embodiment transfer, it is crucial to extract action representations in-addition to state representations. While various works have successfully developed methods for learning state representations through pre-training on large quantities of human demonstrations, they often leave valuable action information untapped. Therefore, our goal is to extract both state and action representations from human demonstrations, specifically visuo-tactile, observation-only demonstrations. To accomplish this, we leverage latent dynamics learning to jointly learn state and action representations.

We consider a dataset consisting of human demonstrations alongside a smaller dataset of action labeled robot demonstrations.  The small dataset of robot demonstrations are obtained via teleoperation via teleoperation or pretrained reinforcement learning policies whichever is applicable. As a reminder, human demonstrations here refers to observations of human hand performing dexterous skills modalities beyond visual such as tactile sensing.

\begin{figure}
    \centering
    \includegraphics[width=0.8\linewidth]{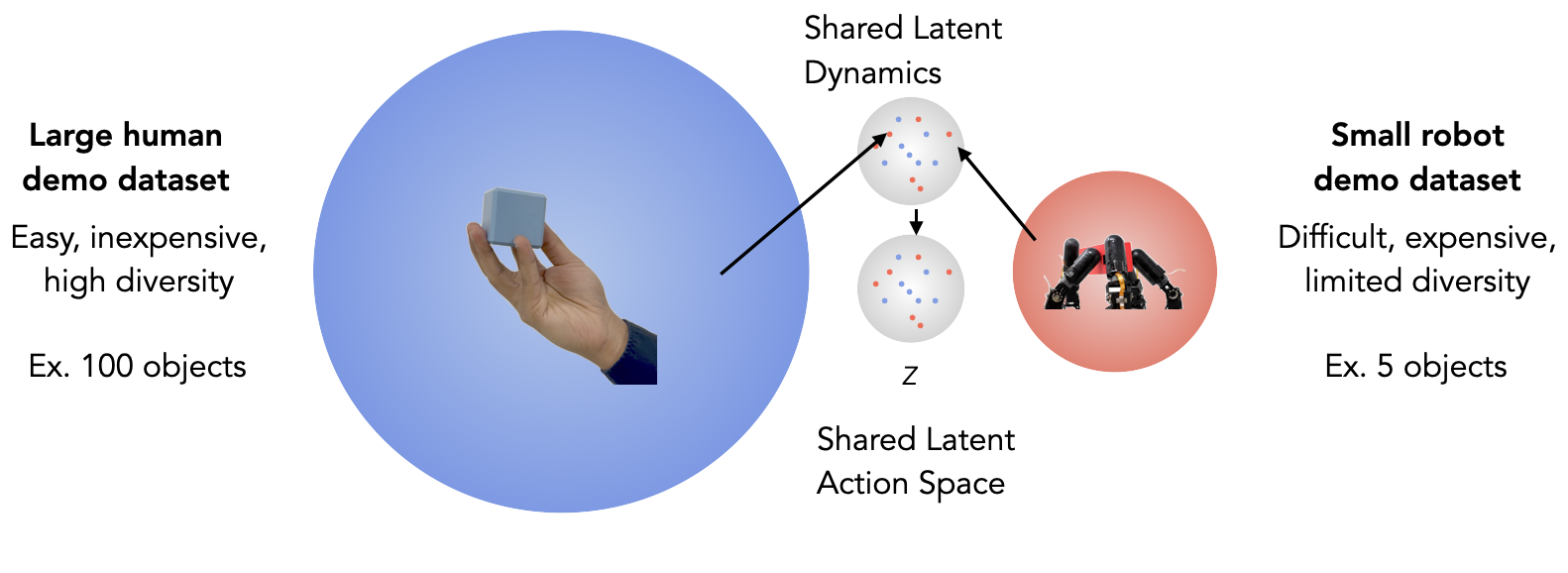}
    \caption{The key idea of our method is to learn from a large dataset consisting of human demonstrations alongside a small dataset of robot demonstrations by mapping them to a shared latent action space.}
    \llabel{fig:enter-label}
\end{figure}

We propose a pre-training framework which pre-trains on a large dataset of human demonstrations and fine-tunes on a smaller dataset of robot demonstrations to transfer dexterous skills from humans to robots. The primary challenge in learning from both human and robot demonstrations is the difference in their observation and action spaces. To address this, we align the two embodiments by enforcing a shared latent action space and developing a framework for latent imitation. As a result, our pre-training framework extracts a latent action policy from human demonstrations, facilitating the transfer of skills to robots using a limited set of robot demonstrations


\section{A new paradigm of human demonstrations for dexterity}

As discussed a fundamental challenge in collecting robot demonstrations for multi-fingered dexterous manipulation via teleoperation is the lack of tactile feedback. While numerous technologies have been developed for tactile sensing, the ability to render this back to the teleoperator is limited and is an open problem. This challenge clearly places a ceiling on quality and dexterity of the skills for which demonstrations can be acheived via teleoperation. 

\begin{figure}
    \centering
    \includegraphics[width=0.8\linewidth]{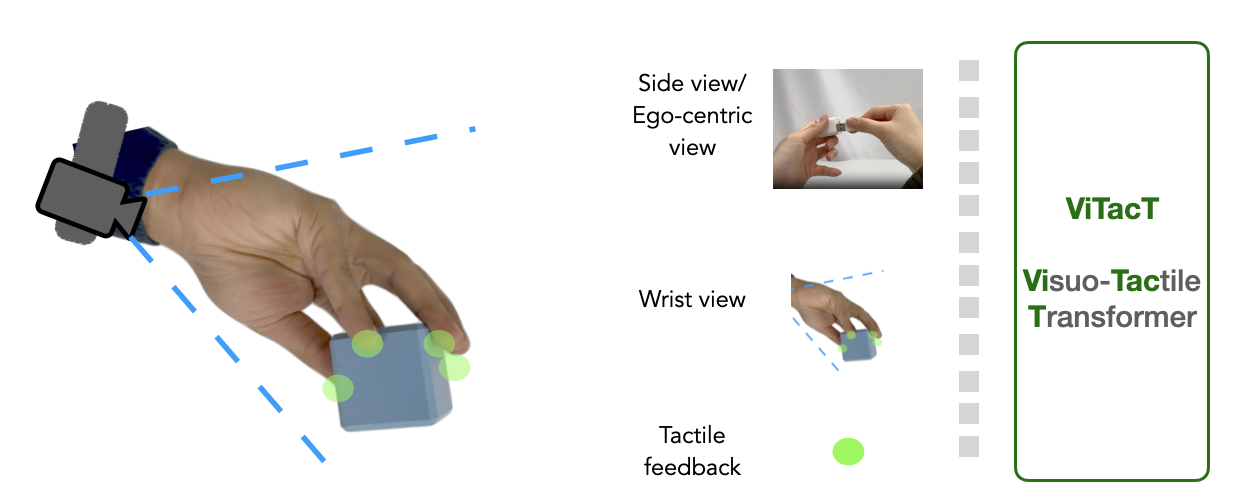}
    \caption{An illustration of the proposed method of collecting visuo-tactile human demonstrations. Multiple camera views patches and tactile sensor feedback can be encoded via visuo-tactile transformer (ViTacT).}
    \label{fig:human_demo}
\end{figure}

Therefore, we propose sensorizing the human hand and collecting demonstrations while the human operator performs dexterous skills. As shown in Fig \ref{fig:human_demo}, we can equip the fingertip tactile sensing towards collecting demonstrations for precision in-hand manipulation. In addition to tactile feedback, we consider visual feedback with a wrist view camera and ego centric camera. We encoder these multi-modal observations by extending the ViT architecture to include tactile feeback and refer to it as Visuo-tactile Transformer (ViTacT).


\section{Problem Formulation}

Let $\mathcal{D}_{H}$ be the dataset of observation-only human demonstrations as discussed above and let $\mathcal{D}_{R}$ be the data set of action-labeled demonstrations obtained by teleoperation or previously trained reinforcement learning policies. Our human demonstrations are observation-only as they do not contain action labels. Given the ease of obtaining human demonstrations and the difficulty of robot demonstrations $\mathcal{D}_H >> \mathcal{D}_R$. 

Our goal is to achieve policy for dexterous skills demonstrated in $\mathcal{D}_H$.  Let this policy be denoted by $\pi(\bm{a}_t | \bm{o}_{t}, \ldots,\bm{o}_{t-k}, u)$ where $\bm{o}_{t}, \ldots,\bm{o}_{t-k}$ is a history of observations, $u$ is some conditioning variable such as a goal, a task label, or language instruction. With the motive of learning policy in a latent-space common between human and robot embodiments, we decompose the problem of learning $\pi$ into learning a latent policy $\psi(\bm{z}_t | \bm{o}_{t}, \ldots,\bm{o}_{t-k}, u))$ and an action head $g(\bm{a}_t | \bm{z}_t)$ where $\bm{z}_t$ denotes the latent action. 

 First, we develop the observation-only imitation learning used to learn $\psi()$ with observation-only demonstrations through sections 6.1 - 6.5. In section 6.6, with our method, we describe the steps for pre-training with human demonstrations $\mathcal{D}_{R}$ and fine-tuning with robot demonstrations $\mathcal{D}_{H}$ for learning $\psi()$ and action head $g()$ to compose $\pi()$.    




\section{Learning latent action policies via forward dynamics}
\llabel{sec:lapo}

\begin{figure}[h]
    \centering
    \includegraphics[width=0.7\textwidth]{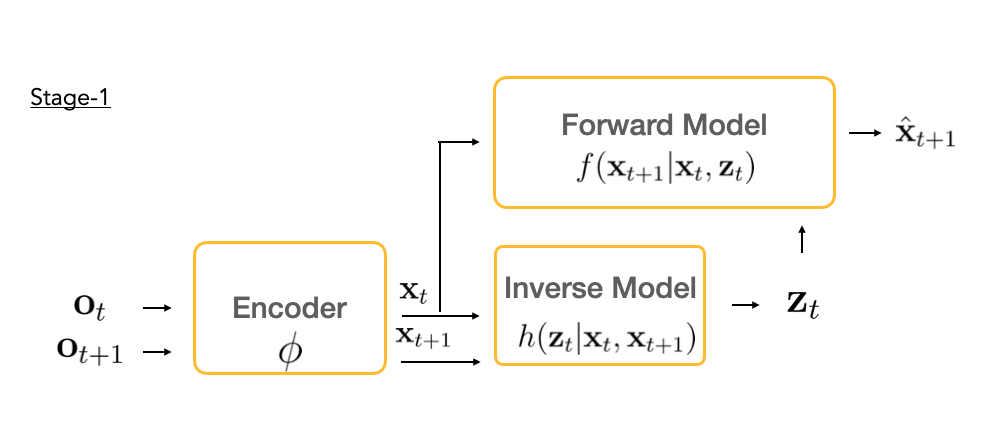}
    \caption{Latent actions achieved by learning forward dynamics in representation space achieved by minimizing the prediction error in the latent space - $d(\bm{x}_{t+1},\bm{\hat{x}}_{t+1})$ where $d$ is some similarity measure.}
    \llabel{fig:lapo}
\end{figure}

We adopt model-based observation-only imitation learning and learn to extract a latent policy. Our method to learn $\psi()$ consists of three stages.

Let encoder $\phi$ be the ViTacT encoder, and let be $\bm{x}_t$ and $\bm{x}_{t+1}$ be representations of observations $\bm{o}_t$ and $\bm{o}_{t+1}$ obtained via encoder $\phi$. 

In stage-1, we learn a latent space forward dynamics model $f(\bm{x}_{t+1} | \bm{x}_t, \bm{z}_t)$ towards learning latent policy $h(\bm{z}_t | \bm{x}_t, \bm{x}_{t+1})$, particularly extending the method proposed in LAPO. Key to this approach is the information bottleneck introduced with a low dimensional latent action. Then the prediction error in the latent state space, $d(\bm{x}_{t+1},\bm{\hat{x}}_{t+1})$ is minimized where $d$ is some similarity measure. However, optimizing latent state prediction loss can easily lead to latent collapse. Our approach to learn meaningful latent state and action space while simultaneously learning forward dynamics is discussed in \lref{sec:lapo+dino}.

\begin{figure}[h]
    \centering
    \includegraphics[width=0.8\textwidth]{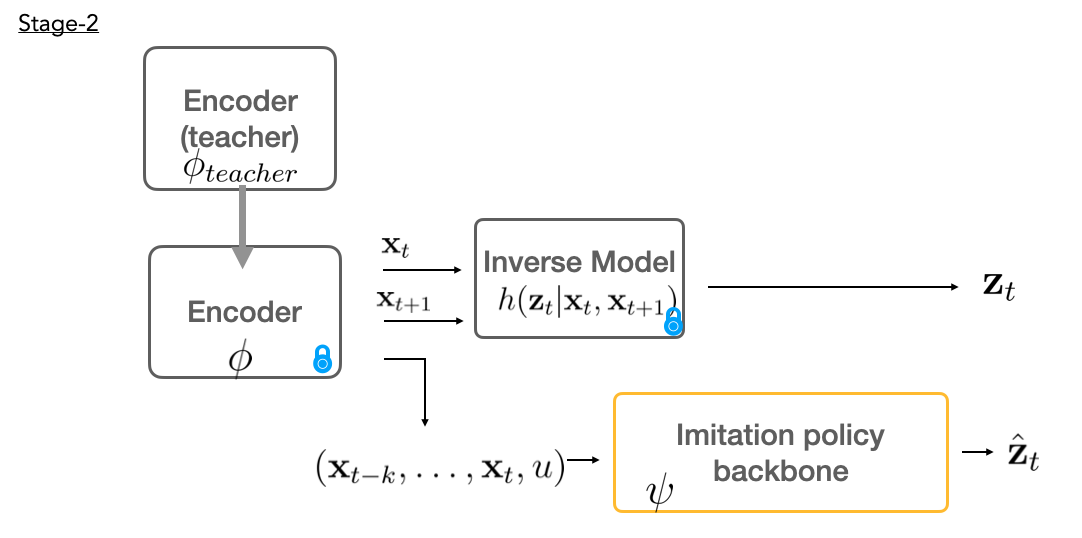}
    \caption{Learning task conditioned latent policy $\psi$ with supervision from inverse model $h()$ where $u$ denotes the task.}
    \llabel{fig:lapo-stage2}
\end{figure}

In stage-2, $\psi'(\bm{z}_t | \bm{o}_{t}, u)$ is trained to directly predict latent actions with supervision from inverse dynamics model $h()$ obtained in the stage-1. Note that $\psi'()$ can be backbone network of choice for downstream imitation learning.

\begin{figure}[h]
    \centering
    \includegraphics[width=\textwidth]{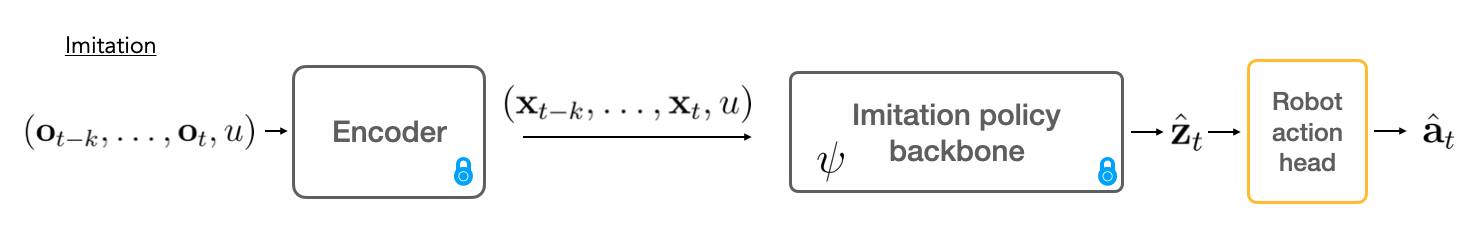}
    \caption{Learning robot action head $g()$ with robot demonstrations $\mathcal{D}_R$}
    \llabel{fig:imitation}
\end{figure}

In a final imitation stage, while freezing all other networks, the action head $g()$ is trained to predict the true actions with imitation learning loss of choice.

\section{Preliminaries: self-distillation for representation learning}

Reconstruction-based representation learning is quite computationally expensive for such high-dimensional multi-modal observations. Hence, we turn to efficient knowledge distillation for representation learning which have been demonstrated yield better representations while using lower memory and training time. 

Self-distillation consists of learning teacher and student networks. These are sometimes referred to as the target and online networks. The embeddings produced by the teacher and student are compared with a choice of similarity loss function to update the student network. While the teacher network is typically larger than the student network, recent methods use identical architecture to update teacher weights with student weights.

Distillation with no labels (DINO) is one such self-supervised learning method to train vision transformers (ViTs). It proposes simple cross-entropy loss to compare the output embeddings and exponential moving averaging to update teacher network with student network weights. Training ViTs with DINO leads to interesting emerging properties such as self-supervised learning of segmentation masks. Using global and local crops of the image is a key step in enabling DINO towards these emergent properties. The embeddings of the global crops with the teacher network are compared with embeddings of the local and global crops obtained via the student network. The cross-entropy loss is computed pairwise between these embeddings. ViTs trained with DINO on ImageNet dataset demonstrate emergence of segmentation masks as visualized by attention map. 

\begin{align}
    \mathcal{L}_{DINO} &= - \text{softmax} [g_{\theta_t}(\bm{x}_2)] \log \text{softmax} [g_{\theta_s}(\bm{x}_1)] 
\end{align}

\section{Self-distillation for learning latent action policies via forward dynamics}
\llabel{sec:lapo+dino}
We use DINO-style self-distillation representation learning to encode the modalities of demonstrations with a transformer.  In particular, we use a visuo-tactile transformer to encode multi-view visual demonstration along with tactile sensing data and use self-distillation learning steered by forward dynamics to drive representation. As in DINO, we use the cross-entropy loss but instead of encoding local and global crops of an visual input, we encode current and future visuo-tactile observations and predict the encoding of future visuo-tactile observations with forward dynamics. We minimize cross-entropy loss to compare the ground truth and future state embedding to learn the visuo-tactile transformer.

\begin{align}
    d(\bm{x}_{t+1}, \hat{\bm{x}}_{t+1})&= - \text{softmax} [\phi(\bm{o}_{t+1})] \log \text{softmax} [\phi(\hat{\bm{o}}_{t+1})]
\end{align}
\begin{figure}
    \centering
    \includegraphics[width = 0.8 \textwidth]{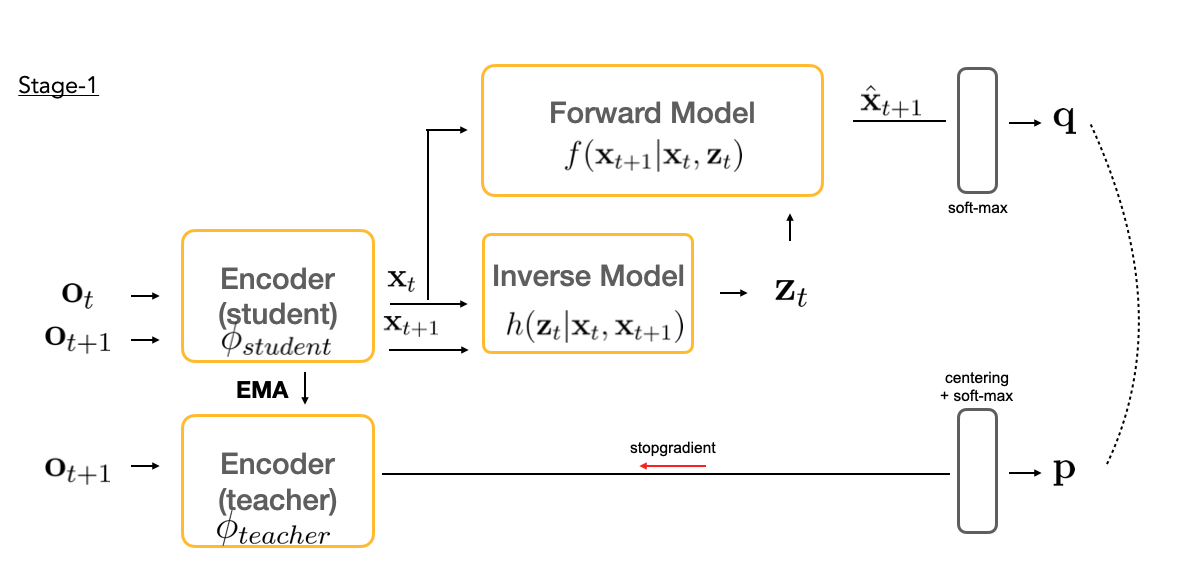}
    \caption{Learning forward dynamics via DINO style representation learning}
    \llabel{fig:lapo+dino}
\end{figure}

The total loss for learning latent policies for cross embodiment transfer includes regularizing KL-divergence loss, in addition to the cross-entropy loss above.
\begin{figure}
    \centering
    \includegraphics[width=\linewidth]{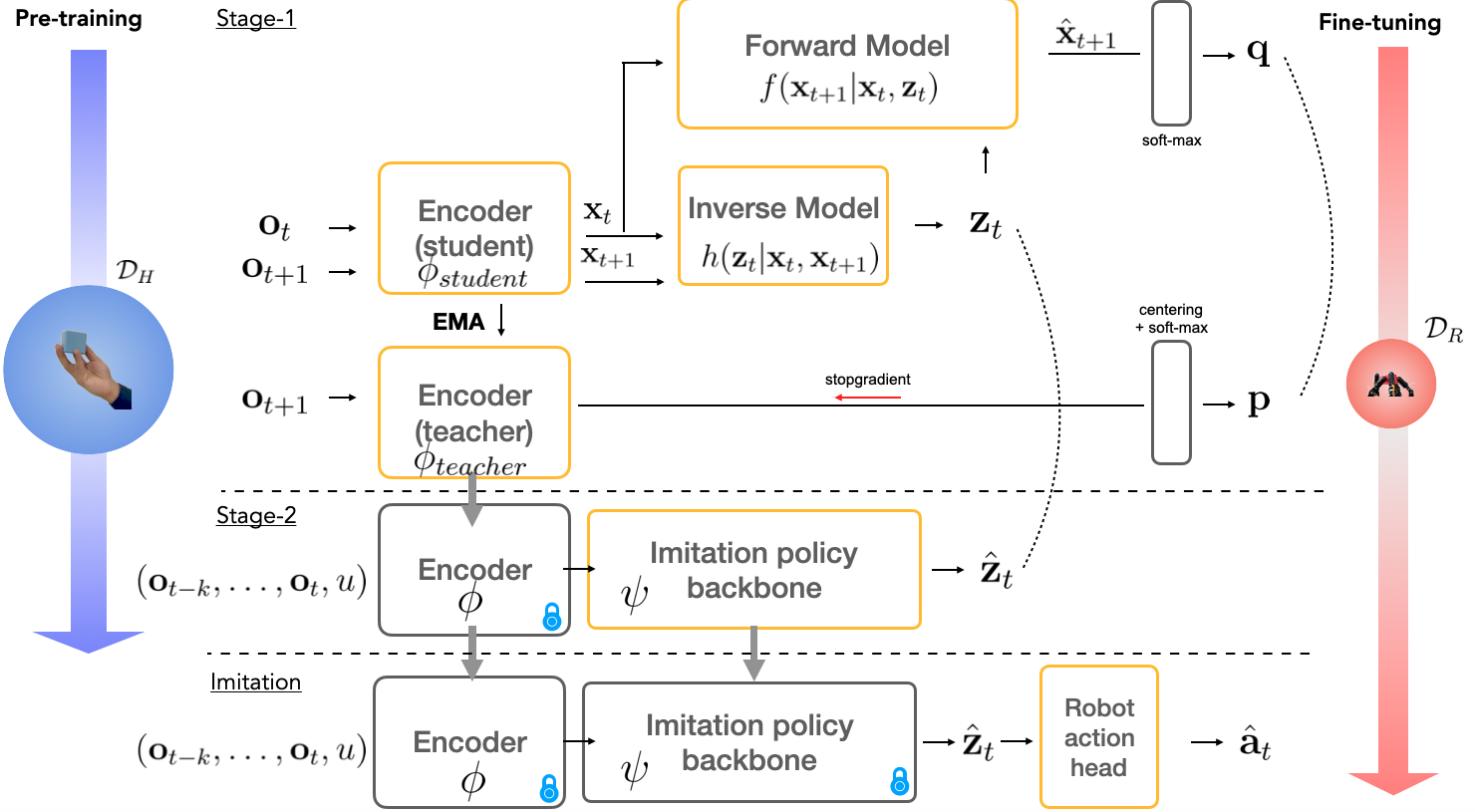}
    \caption{Pre-training with human demonstrations followed by fine-tuning and imitation with robot demonstrations}
    \llabel{fig:training}
\end{figure}

\begin{align}
    \mathcal{L} &= - \text{softmax} [\phi(\bm{o}_{t+1})] \log \text{softmax} [\phi(\hat{\bm{o}}_{t+1})] \nonumber \\ 
    &+ \beta_1\text{KL}(f(. |\bm{x}_t,  \bm{x}_{t+1}) | \mathcal{N}(0, I)) \nonumber \\ 
    &+ \beta_2\text{KL}(h(. |\bm{x}_t,  \bm{z}_t) | \mathcal{N}(0, I))
\end{align}
where $\beta_1$ and $\beta_2$ are coefficients of regularizing KL divergence loss for latent state and action distributions respectively.

\section{Pre-training and fine-tuning}
\llabel{ch5:sec:training}

We learn the encoder $\phi$, forward model $f$, inverse model $g$ and imitation policy backbone $\psi$ first by pre-training with human demonstrations $\mathcal{D}_H$ followed by fine-tuning with robot demonstrations $\mathcal{D}_R$.  Both pre-training and fine-tuning consist of two-stages first to train the encoder $\phi$, forward model $f$, inverse model $g$ followed by learning imitation policy backbone $\psi$ via supervision from latent actions provided by inverse model $g$. The complete training process is illustrated in \lref{fig:training}. The imitation policy backbone $\psi$ is used for imitation learning by learning to predict robot actions. Finally, after freezing $\phi$ and $\psi$, the separate action head $g$ is trained to predict true robot actions with $\mathcal{D}_R$.

\section{Preliminary Experiments \& Results}

The experimental results are towards validating the overall framework with a simple pick task. 

\subsection{Task and demonstrations} The task is to pickup an object placed on the table with gripper equipped with tactile sensing. To achieve this we collect demonstrations from the human and the robot and the setups are shown in Fig \lref{fig:robotsetup} and Fig \lref{fig:humansetup} respectively. We collect robot demonstrations for one object whereas we collect human demonstrations for multiple (two) objects. The robot demonstrations are collected via teleoperation and due its difficulty we collect only 20 demonstrations. However, human demonstrations are easier to achieve therefore we collect 250 demonstrations per object.
\begin{figure}
    \centering
    \includegraphics[width=0.90\linewidth]{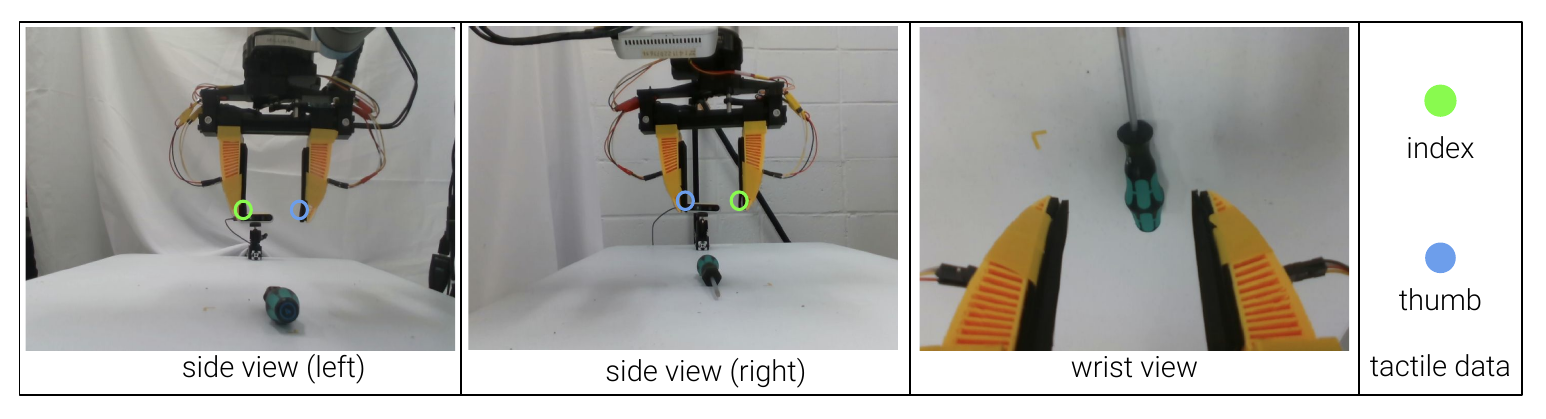}
    \caption{The robot setup shown above consists of 3 camera views - two side views and 1 wrist view as shown above is equipped with a two-fingered gripper. The tactile feedback is obtained via tactile sensor embedded in the tip of the gripper fingers}
    \llabel{fig:robotsetup}
\end{figure}

\begin{figure}
    \centering
    \includegraphics[width=\linewidth]{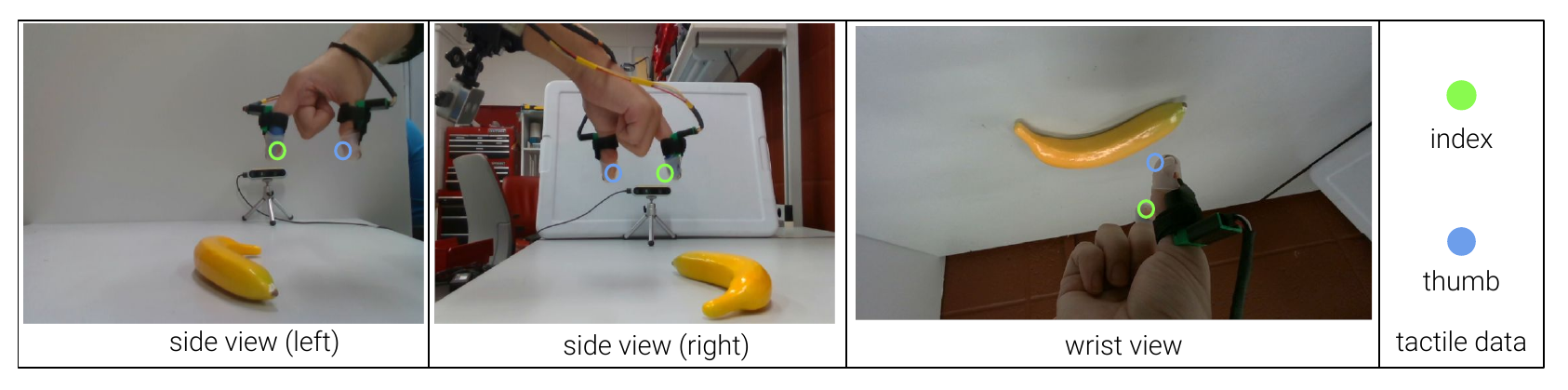}
    \caption{Human data collection setup consists camera views similar to the that of robot. Two side cameras and one wrist camera view. The tactile data is collected finger cap device that embeds the singletact capacitive tactile sensor.}
    \llabel{fig:humansetup}
\end{figure}

\subsection{Pre-training with human demonstrations}
We perform stage-1 pre-training with human demonstrations 500 human demonstrations as discussed above. We use train with ViTacT with 12 layers, 3 heads and 192 dimensions for over 100 epochs till convergence. The attention maps of ViTacT transformer. The attention heatmap shows greater attention towards the dynamic elements of the scene as visualized in Fig \lref{fig:attention}. We can conclude that latent state and action representations extracted are promising for further fine-tuning and imitation learning. We hope to achieve this in future work.   




\begin{figure}
    \centering
    \includegraphics[width=\linewidth]{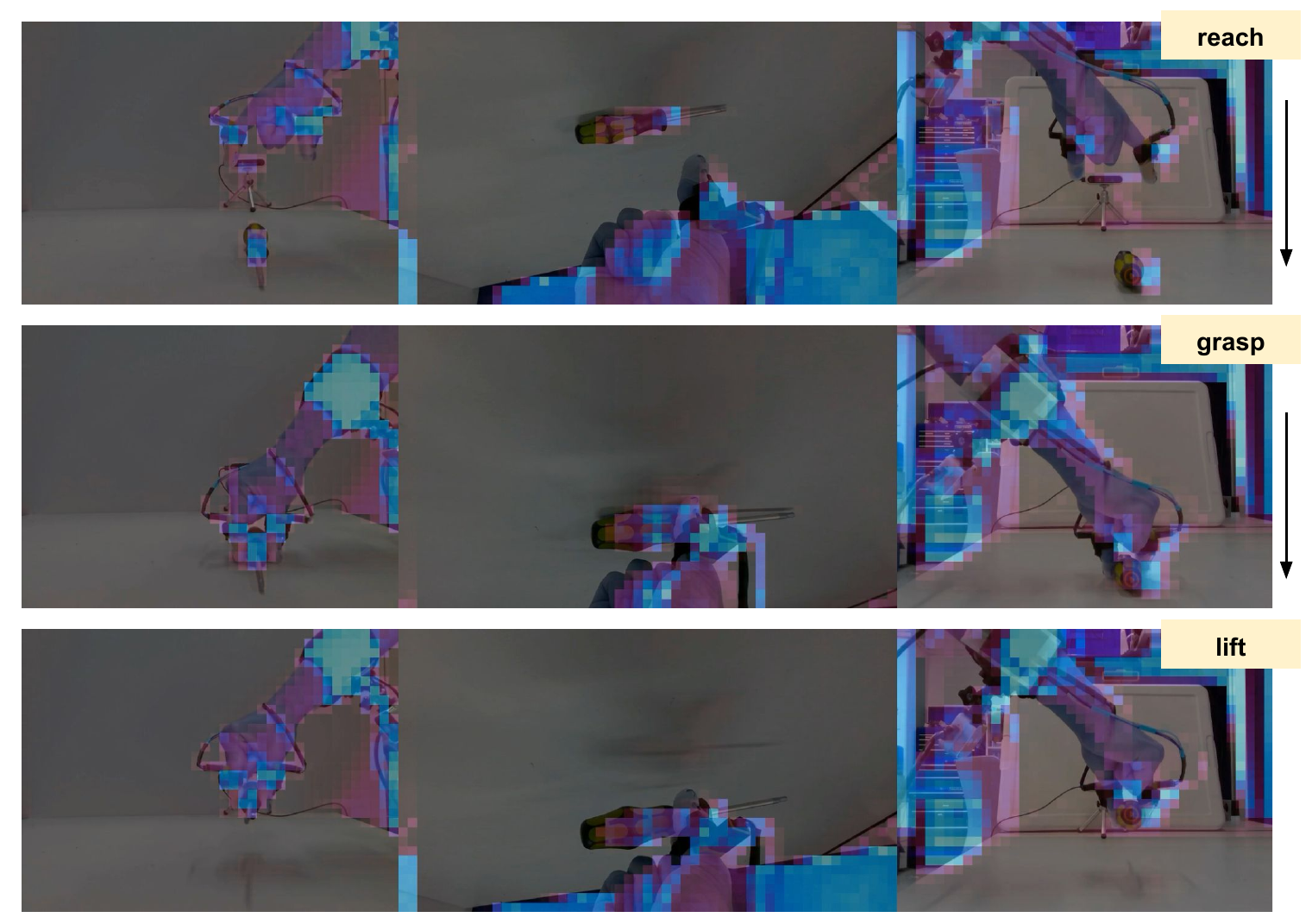}
    \caption{The attention visualized during different phases of the demonstraions i.e reach, grasp, and lift phases.}
    \llabel{fig:attention}
\end{figure}



\chapter{Conclusion}

In this thesis, I aimed to achieve human-level dexterity through data-driven robot learning. A key focus has been on in-hand manipulation, integrating the sense of touch, which is crucial for developing versatile, dexterous skills required to reach the benchmark of human-level dexterity.

This research establishes the unique challenges involved in learning multi-fingered manipulation. At its core, the difficulty lies in the problem of exploration: learning complex skills without relying on robot demonstrations, which are often impractical to generate. In-hand manipulation exemplifies these challenges, where naive random exploration proves ineffective in uncovering meaningful behaviors.

To enable reinforcement learning, this thesis emphasizes the importance of structured exploration over traditional random methods for learning complex in-hand manipulation skills. Traditional exploration often fails to adequately cover the high-dimensional state spaces typical of multi-fingered manipulation, leading to inefficiencies and limited outcomes. Structured exploration, in contrast, provides a systematic approach to guide the agent toward meaningful regions in the state space, enhancing learning efficiency and enabling the acquisition of more advanced skills.

A key strategy in this research involves leveraging reset state distributions (also known as initial state distributions) to provide the agent with informed starting points during training. By using structured, exploratory resets, the system can focus on skill spaces that would otherwise remain unexplored through unstructured exploration alone. This targeted approach accelerates the learning process and improves the chances of discovering effective strategies for in-hand manipulation, especially in challenging environments.

To construct these reset state distributions, we propose the use of sampling-based planning methods that address the data scarcity inherent in reinforcement learning for dexterous skills. These planning techniques generate valuable state-action trajectories, forming a robust foundation for exploration. Specifically, we developed modified versions of two classical sampling-based planning algorithms: Rapidly-exploring Random Tree (RRT) and Probabilistic Roadmap (PRM). Additionally, actions from these trajectories are used to bootstrap policy learning. We demonstrate that this approach, combined with effective sim-to-real transfer techniques, significantly enhances reinforcement learning outcomes.

Overall, this work presents a two-step approach for learning challenging motor skills through reinforcement learning. The first step uses sampling-based planning to generate demonstration trajectories, while the second step bootstraps reinforcement learning exploration using these trajectories. Although this method was developed specifically for complex in-hand manipulation, it holds potential benefits for a wide range of robotic manipulation and locomotion tasks achieved through reinforcement learning.

Due to several limitations of reinforcement learning, such as difficulty in scaling to a large number of skills and challenges with sim-to-real transfer, this thesis also explores imitation learning. Notably, it proposes a novel approach to collecting demonstrations by equipping human hands with tactile sensing instrumentation, allowing the use of visuo-tactile data for challenging multi-fingered manipulation tasks aimed at achieving human-level dexterity. Preliminary results are promising, demonstrating the effectiveness of ViTacT—a transformer-based encoder—for extracting latent action spaces to bootstrap imitation learning when combined with robot demonstrations.


A recurring theme throughout this work has been the interplay between imitation learning (IL) and reinforcement learning (RL). Both methods offer unique strengths and trade-offs when applied to dexterous manipulation, but the choice between them remains an open research question.

While IL with visuo-tactile demonstrations shows promising results, especially for skills that are difficult to program through reward-based exploration, the challenge lies in obtaining robot demonstrations at scale. Human demonstrations will continue to be critical, but distribution shifts between human and robot domains pose additional challenges, often necessitating RL to fine-tune models for real-world deployment. Therefore, the balance between IL and RL, along with the need for small amounts of robot demonstrations, is a critical consideration for scalable dexterous manipulation.

Planning for complex, structured skills remains another open challenge, as the computational complexity of planning scales exponentially with the planning horizon. As such, future research must consider how to scale structured exploration alongside learning methods, ensuring that both IL and RL complement each other effectively. 

It remains uncertain which paradigm—IL, RL, or some combination—will be fundamental to developing foundation models for dexterous intelligence. The answer may lie in hybrid approaches that integrate both paradigms.

Achieving dexterous skills opens new possibilities in the pursuit of embodied AI. Many of the most valuable representations of the world are rooted in interaction, and complex dexterous manipulation enables learning fine-grained representations that remain inaccessible through simpler interactions. If embodied AI—where intelligence is inherently tied to a physical form—represents the next frontier, the choice of embodiment becomes crucial. A dexterous robotic hand could play a pivotal role in this advancement, driving the development of richer, more intricate representations of the physical world and expanding the potential of embodied intelligence.

In conclusion, this thesis lays the groundwork for achieving human-level dexterity through a combination of structured exploration, imitation learning, and reinforcement learning. While the challenges of cross-embodiment learning, skill diversity, and planning remain, the potential for dexterous manipulation to advance embodied AI is undeniable. Moving forward, human demonstrations and dexterous embodiment will be pivotal in building the next generation of intelligent systems capable of interacting with the world at a human-level dexterity.


\clearpage
\phantomsection

\clearpage
\phantomsection 
\titleformat{\chapter}[display]
{\normalfont\bfseries\filcenter}{}{0pt}{\large\bfseries\filcenter{#1}}  
\titlespacing*{\chapter}
  {0pt}{0pt}{30pt}

\begin{singlespace}  
	\setlength\bibitemsep{\baselineskip}  
	\addcontentsline{toc}{chapter}{References}  
	\printbibliography[title={References}]
\end{singlespace}


\titleformat{\chapter}[display]
{\normalfont\bfseries\filcenter}{}{0pt}{\large\chaptertitlename\ \large\thechapter : \large\bfseries\filcenter{#1}}  
\titlespacing*{\chapter}
  {0pt}{0pt}{30pt}	
  
\titleformat{\section}{\normalfont\bfseries}{\thesection}{1em}{#1}

\titleformat{\subsection}{\normalfont}{\thesubsection}{0em}{\hspace{1em}#1}


\end{document}